\documentclass[10pt]{article}

\usepackage{geometry}  
\usepackage[english]{babel}
\usepackage[utf8x]{inputenc}
\usepackage[T1]{fontenc}

\usepackage[dvipsnames]{xcolor}
\usepackage{paralist}
\usepackage{graphicx}
\usepackage{subcaption}
\usepackage{longtable} 
\usepackage{multirow}
\usepackage{listings}
\usepackage{makecell}
\usepackage{array}
\usepackage{float}
\usepackage{dsfont}
\usepackage{rotating}
\usepackage{booktabs}
\usepackage{enumerate}
\usepackage{tikz}
\usepackage{pgf}
\usepackage{enumitem}

\usepackage{amsmath}
\usepackage{amssymb}
\usepackage{amsthm}
\usepackage{bm}
\usepackage{mathtools}
\usepackage{mathrsfs}

\usepackage{natbib}
\usepackage[
  colorlinks,
  citecolor=blue,
  linkcolor=red,
  anchorcolor=red,
  urlcolor=blue
]{hyperref}
\mathtoolsset{showonlyrefs}
\usepackage{authblk}
\usepackage{hyperref}
\makeatletter
\newcommand{\printfnsymbol}[1]{%
  \textsuperscript{\@fnsymbol{#1}}%
}
\makeatother

\usepackage{xcolor}

\theoremstyle{plain}

\newtheorem{definition}{Definition}
\newtheorem{theorem}{Theorem}
\newtheorem{lemma}{Lemma}

\newtheorem{corollary}{Corollary}
\newtheorem{remark}{Remark}
\newtheorem{assumption}{Assumption}

\newcommand{\var}{\mathrm{Var}}

\newcommand{\argmax}[1]{\underset{#1}{\arg\!\max}}
\newcommand{\argmin}[1]{\underset{#1}{\arg\!\min}}

\usepackage{xspace}

\def\@#1\@{\begin{align}#1\end{align}}
\def\$#1\${\begin{align*}#1\end{align*}}

\definecolor{myblue}{rgb}{.8, .8, 1}
\definecolor{mathblue}{rgb}{0.2472, 0.24, 0.6} 
\definecolor{mathred}{rgb}{0.6, 0.24, 0.442893}
\definecolor{mathyellow}{rgb}{0.6, 0.547014, 0.24}

\newcommand{\tu}{{\tilde{u}}}

\newcommand{\tQ}{{\tilde{Q}}}

\newcommand{\tZ}{{\tilde{Z}}}

\newcommand{\calA}{{\mathcal{A}}}

\newcommand{\calH}{{\mathcal{H}}}
\newcommand{\calI}{{\mathcal{I}}}

\newcommand{\calM}{{\mathcal{M}}}
\newcommand{\calN}{{\mathcal{N}}}

\newcommand{\calT}{{\mathcal{T}}}

\newcommand{\calX}{{\mathcal{X}}}

\newcommand{\EE}{\mathbb{E}}
\newcommand{\PP}{\mathbb{P}}

\newcommand{\RR}{\mathbb{R}}

\usepackage[ruled]{algorithm2e}
\usepackage{stfloats}
\newcommand{\revise}[1]{{\color{black} #1}}

\newcommand{\ts}{\tilde{s}_0}
\newcommand{\tth}{\tilde{\theta}}
\newcommand{\bth}{\bar{\theta}}
\newcommand{\hth}{\hat{\theta}}
\newcommand{\alg}{\textnormal{\textbf{Alg}}}
\newcommand{\tmj}{T^{(j)}_m}
\newcommand{\tm}{T^{(m)}}
\newcommand{\modelc}{\textsf{Model-C}}
\newcommand{\modelp}{\textsf{Model-P}}

\newcommand{\dd}{\textnormal{d}}
\newcommand{\bR}{\mathbb{R}}
\newcommand{\bE}{\mathbb{E}}
\newcommand{\Expect}{\mathbb{E}}
\newcommand{\Prob}{\mathbb{P}}
\newcommand{\Indc}{\mathbf{1}}
\newcommand{\TV}{\textsf{TV}}
\newcommand{\supp}{\textsf{supp}}
\newcommand{\Unif}{\mathsf{Unif}}

\long\def\comment#1{}

\graphicspath{{./figs/}}

\title{Dynamic Batch Learning in High-Dimensional \\
Sparse Linear Contextual Bandits
}
\author{Zhimei Ren\thanks{Department of Statistics, University of Chicago. \texttt{zmren@uchicago.edu} }
~and Zhengyuan Zhou\thanks{Stern School of Business, New York University. \texttt{zzhou@stern.nyu.edu}} }
\date{}

\begin{document}

\maketitle

\begin{abstract}
We study the problem of dynamic batch learning in high-dimensional sparse linear contextual bandits, where a decision maker, under a given maximum-number-of-batch constraint and only able to observe rewards at the end of each batch, can dynamically decide how many individuals to include in the next batch (at the end of the current batch) and what personalized action-selection scheme to adopt within each batch. Such batch constraints are ubiquitous in a variety of practical contexts, including personalized product offerings in marketing and medical treatment selection in clinical trials. We characterize the fundamental learning limit in this problem via a regret lower bound and provide a matching upper bound (up to log factors), thus prescribing an optimal scheme for this problem. To the best of our knowledge, our work provides the first inroad into a theoretical understanding of dynamic batch learning in high-dimensional sparse linear contextual bandits. Notably, even a special case of our result---when no batch constraint is present---yields that the simple exploration-free algorithm using the LASSO estimator already achieves the minimax optimal $\tilde{O}(\sqrt{s_0T})$ regret bound ($s_0$ is the sparsity parameter or an upper bound thereof and $T$ is the learning horizon) for standard online learning in high-dimensional linear contextual bandits (for the no-margin case), a result that appears unknown 
in the emerging literature of high-dimensional contextual bandits. 
\end{abstract}
 
\section{Introduction}
\label{sec:intro}
With the growing abundance of user-specific data, service personalization---tailoring service decisions based on each individual's own characteristics---has emerged to be a predominant paradigm in data-driven decision making. This is because through personalization, a decision maker can exploit the heterogeneity in a given population by selecting the best decisions on a fine-grained individual level, thereby improving the outcomes. Such heterogeneity is ubiquitous; and intelligently capturing its benefits through personalization has found immense benefits across a wide range of applications in operations management, including medical treatment selection in clinical trials, product recommendation in marketing, ads selection in online advertising and nurse staffing in hospital operating rooms~\citep{bertsimas2007learning,kim2011battle, SBF2017, mintz2017behavioral, ferreira2018online, zhou2018evaluating, bastani2018interpreting, hopp2018big, ban2019big, miao2019fast}. 

In the current era, such data-driven personalized decision making problems
often exhibit both high-dimensionality and sparsity \citep{naik2008challenges, kim2011battle, belloni2011high, bayati2014data, belloni2014inference, razavian2015population,zhou2018evaluating}. High-dimensionality refers to the fact that, as a result of modern data-collection technologies, a large number of features about individuals are collected and recorded in the datasets, hence making the covariate vector high-dimensional. At the same time, the underlying reward response model is often \textit{sparse}, where only a few of those covariates actually influence the rewards. To capture these two aspects, and to take into account the sequential decision making nature of personalization, such problems have been formalized in the framework of high-dimensional sparse linear contextual bandits, where the contexts are \textbf{iid} drawn from an underlying distribution and the context dimension $d$ is comparable or even exceeds the learning horizon $T$, while at most $s_0$ ($\ll d$) context variables influence the (random) reward, which in expectation is a linear function of the context vector.

Driven by a pressing need to achieve effective personalization in this challenging regime, an emerging line of work~\citep{wang2018minimax,kim2019doubly, bastani2020online}  has developed algorithms and established regret guarantees, where regret measures the performance difference between the cumulative reward generated by the algorithm and that achieved by an optimal policy (if the underlying model were known).  This line of work has exploited the fact that the underlying linear model is sparse in order to achieve regret bounds that scale gracefully with $s_0$ (much smaller than the ambient dimension $d$). For instance, under further margin conditions (where a gap between the optimal action and suboptimal actions can be identified with positive probability and where regret logarithmic in $T$ is thus possible), \cite{bastani2020online} has developed a forced-sampling exploration scheme that is used jointly with the LASSO estimator, and established
a $O\big(s_0^2 \cdot (\log d + \log T)^2\big)$ regret bound. Building on~\cite{bastani2020online}, \cite{wang2018minimax} then subsequently\footnote{A preprint of~\cite{bastani2020online} occurred prior to \cite{wang2018minimax}.} used the same forced-sampling exploration scheme, but with a different minimax concave penalty weighted LASSO estimator and obtained 
the $O\big(s_0^2 \cdot (\log d + s_0) \cdot \log T \big)$ regret bound, 
an improvement if $s_0$ is not much larger compared to $\log T$ and/or $\log d$.
When no margin condition exists (in which case dependence on $T$ is at best $\Omega(\sqrt{T})$),~\citet{kim2019doubly} has constructed a doubly-robust LASSO estimator based algorithm (with uniform sampling exploration) that achieves $\tilde{O}(s_0 \sqrt{T})$ regret. Additionally, several earlier works~\citep{carpentier2012bandit, abbasi2013online} also studied high-dimensional linear contextual bandits but did not use LASSO based methods: they are either restricted to specialized settings---special action set structure and nonstandard noise in \cite{carpentier2012bandit}---or obtained regret bounds that are worse\footnote{In \cite{abbasi2013online}, a $\tilde{O}(\sqrt{s_0 d T})$ regret bound is obtained; although the contexts there can be arbitrary rather than stochastic.} than $\tilde{\Theta}(\sqrt{dT})$. Taken together, these developments represent fruitful inroads into the high-dimensional regimes that intelligently exploited sparsity for practical benefits.

Despite these fruitful studies, an important aspect is missing in this line of work that limits their applicability in practice. The standard online learning model adopted in the literature---where a decision is made on the current individual, yielding an outcome that is immediately observed and incorporated to make the next decision---is simply impractical in many applications. In practice, while decision makers are able to perform active learning and incorporate feedback from the past to adapt their decisions in the future, they are often limited by the physical, cost or regulatory constraints that any adaptation is often limited to a fixed number of rounds of interaction, something we refer to as the batch constraint in this paper.  For instance, when running a personalized product marketing campaign---a prime example where high-dimensional customer data is available~\citep{bertsimas2007learning, SBF2017}---a company often needs to mail personalized product offers to its (existing and/or potential) customers. Here, the marketer will not (and cannot afford to) make a product offer to one customer, wait to receive feedback and then move on to the next customer (the standard online learning model). Instead, the marketer in practice will batch mail a set of customers, receive their feedback collectively and then design the next batch of offerings accordingly. The marketer typically has a targeted customer population at hand (selected from the entire customer base) and, working with a time and monetary budget that dictates the maximum number of batches, needs to design how to optimally partition the customer population into different batches and what product to offer to each customer in a given batch. 

\revise{Another example where such a batch constraint exists is adaptive clinical trials~\citep{robbins1952some, chow2008adaptive, pallmann2018adaptive}, where a fixed number of medical treatments (e.g. different drugs or same drug but different dosages or both) are applied to a group of patients based on the patients' medical characteristics during a phase of the trial, with the medical outcomes collected for the entire group at the end of the phase. The data collected from previous phases are then analyzed to design the next phase, including how many patients to include for the next phase and the medical treatment assignment to the patients. Here, each phase corresponds to a batch of participating patients. As pointed out in~\cite{pallmann2018adaptive}, ``adaptive designs can make clinical trials more flexible by utilising results accumulating in the trial to modify the trial’s course in accordance with pre-specified rules". Depsite being offered such flexibility in adaptive clinical trials (as compared to traditional non-adaptive trials), medical decision makers have limited adaptivity here because the medical outcome for a patient can only be observed after a sufficient amount of time has passed; as such, they must proceed in batches (phases). Note that the current FDA regulation requires four phases for a standard clinical trial. Thus, the trial patients need to be partitioned into four batches, and incorporation of new information only occurs at the end of each batch, thereby rendering the standard online learning model and hence the standard online bandits/contextual bandits algorithms inapplicable. We do point out that pharmaceutical companies that conduct clinical trials often have the sole objective of obtaining FDA approval, and hence do not have the objective of maximizing the total welfare (measured by regret) of the trial patients. In contrast, our paper's focus is on maximizing total welfare (equivalently minimizing regret), and hence would shed light for adaptive clinical trials under this criterion.}

From the above, we see that the key challenge imposed by the batch constraint is the limited adaptability: the adaptation can only occur at a batch level rather than at an individual level. Such limited adaptability forces the decision maker to carefully select the batches, based on available information from the past, so that the inability to adapt (and hence the inferior performance resulted therefrom) does not cause much degradation to the overall performance.
Motivated by these considerations, we study the problem of dynamic batch learning, where a decision maker
dynamically decides the next batch's size (at the end of the current batch) and what personalization scheme to adopt within each batch under a given maximum-number-of-batch constraint. 


\subsection{Main Results}
Our main contributions are twofold.
First, we study the fundamental limits of dynamic batch learning in 
high-dimensional sparse linear contextual bandits. By an 
information-theoretical argument that carefully selects a sequence of
 Bayesian priors, we establish an 
$\Omega\Big(\max \big\{M^{-4} 2^{-\frac{7}{2}M}\sqrt{Ts_0}({T}/{s_0})^{\frac{1}{2(2^M-1)}}, \sqrt{Ts_0}\big\}\Big)$ 
regret lower bound (Theorem~\ref{thm:dynamiclowerbnd}),
 where $M$ is the maximum number of batches allowed. This lower bound---which holds even for the simple standard Guassian contexts---indicates that regardless of how one dynamically makes partitions and/or performs action selection within each batch, the regret can never be made any smaller. For instance, if $M=4$ (as is the case for clinical trials), then no scheme can achieve better regret than $ \Omega(T^{\frac{8}{15}} {s_0}^{\frac{7}{15}})$.
The second term $\Omega(\sqrt{Ts_0})$ in the max is a lower bound\footnote{This result follows directly from the lower bound given in~\cite{chu2011contextual}, although our lower bound argument provides an alternative proof.} for the standard online learning setting, which is automatically a lower bound for dynamic batch learning since the presence of a batch constraint only makes the problem harder. 
Further, the break-even point (up to log factors) between these two terms is $M=\Theta\big(\log{\log{(T/s_0)}}\big)$, 
suggesting that---if the lower bound is tight---only $\Theta\big(\log{\log{(T/s_0)}}\big)$ (practically a constant number) batches are needed to achieve the optimal performance of standard online learning, where no batch constraint exists.

Second, we establish that this lower bound is indeed tight (up to log factors) by providing a matching upper bound. In particular, through a simple LASSO batch greedy learning algorithm (Algorithm~\ref{algo.dbgl_c}),
we establish in Theorem~\ref{thm:upperbnd_p} and Theorem~\ref{thm:upperbnd_c} that the regret is upper bounded by $\tilde{O}\big(\sqrt{Ts_0}({T}/{s_0})^{\frac{1}{2(2^M-1)}}\big)$
when the number of batches does not exceed $O\big(\log{\log{(T/s_0)}}\big)$, hence validating that only $\Theta\big(\log{\log{(T/s_0)}}\big)$ batches are needed to achieve $\tilde{O}(\sqrt{Ts_0})$ regret. Note that it suffices to look at $M$ that is $O(\log{\log{(T/s_0)}})$, because the regret will not get worse and hence will stay at
$\tilde{O}(\sqrt{Ts_0})$ when $M$ gets larger. In particular, a special case of this result (Corollary~\ref{cor:onlineupperbnd}) is that in the standard online learning setting where no-margin exists, we can achieve the minimax optimal regret $\tilde{\Theta}(\sqrt{Ts_0})$ using an exploration-free and computationally efficient algorithm, improving on the $\tilde{O}(s_0 \sqrt{T})$ regret bound given in \cite{kim2019doubly}.

Notably, the algorithm that achieves such strong guarantees is simple: it uses a static grid and is exploration-free.  By the lower bound, using a static grid is not a limitation of the algorithm, but an attestation to its strength (easy implementability in practice). That exploration-free suffices is yet another important message, both for dynamic batch learning and standard online learning.
For the latter, the existing state-of-the-art algorithms \citep{kim2019doubly,bastani2020online}
all use contrived forced-sampling exploration schemes, which is burdensome to implement in practice. However, our results show that they are not necessary, thus echoing in high dimensions a similar message advocated in \cite{bastani2017mostly} for low-dimensional linear contextual bandits.


\subsection{Managerial Insights}

Our results provide insights on how to prescribe the \textit{optimal} personalization scheme when limited adaptivity 
is present in practice and the resulting performance gap (or the absence thereof) when compared to the ideal fully online setting.  
These insights can help managerial decision makers in different ways, depending on the context.
First, when the adaptivity constraint $M$ is fixed \textit{a priori}, such as in the clinical trials setting with $M=4$, our work provides prescriptive solutions for how to design the trials to achieve optimal performance. 
Further, this optimal performance is $ \Omega(T^{\frac{8}{15}} {s_0}^{\frac{7}{15}})$, whereas the infeasible fully online
optimal performance is $\sqrt{Ts_0}$, a quantity that is quite close. 

Second, when the limited adaptivity constraint $M$ is not as rigid and can hence be thought as variable subject to certain budget limit, our results contribute meaningfully to the larger cost/benefit discussions facing the managerial decision makers. For instance, in the personalized product recommendation application, our results indicate that 
$\log\log(T/s_0)$ rounds of campaigns are needed to achieve the (practically infeasible) fully-online performance (where $T$ here corresponds to the number of customers). This is usually a very small number and the result makes it clear that the decision maker should never need to budget for more than that. On the other hand, if under a tight budget constraint (and hence unable to finance $\log\log(T/s_0)$ rounds), the decision maker would be clearly informed by  
the particular benefits under a range of feasible $M$'s and how to execute it optimally once such an $M$ is decided on.
Taken together, we believe our results provide valuable prescriptive insights in the area of adaptive personalization when
limited adaptivity is present.

\subsection{Other Related Work}
The bandits literature is extensive and much of the existing work in that space study low-dimensional contextual bandits (see~\citet{BCN2012, lattimore2018bandit,slivkins2019introduction} for three books on this research area), where the dimension $d$ of the contexts is small compared to the learning horizon $T$ and where many well-performing algorithms have been developed and strong theoretical guarantees have been established (see \citet{FCGS2010, rigollet2010nonparametric, chu2011contextual, goldenshluger2013linear, AG2013a, AG2013b, russo2016information,mintz2020nonstationary} for a highly incomplete list). 
Low-dimensional contextual bandits are not our focus here, and we simply mention in passing that applying (state-of-the-art) results from the low-dimensional contextual bandits literature to the high-dimensional setting 
often yields results that are not useful. For instance, in linear contextual bandits with no-margin, one obtains the $\tilde{\Theta}(\sqrt{dT})$ regret by applying the result in~\citep{chu2011contextual}. Even if such regret bounds continue to hold in high dimensions,\footnote{This may not be the case since the various low-dimensional regime assumptions are often required to obtain the $\tilde{\Theta}(\sqrt{dT})$ regret bounds.} such performance guarantees are not meaningful anymore, because when $d = \Omega(T)$ ($d$ could also be a lot larger than $T$), at least linear regret $\tilde{\Omega}(T)$ is incurred, thus yielding completely ineffective learning.

Additionally, we point out that batch-constrained learning in bandits has been studied before in the literature.
In 2-armed multi-armed bandits (MAB), \cite{perchet2016batched} studied static batch learning where the batch sizes must be decided \textit{a prior}, and established that 
$O(\log \log T)$ batches are needed (via a successive elimination algorithm during each batch) in order to achieve the same regret bound as in standard online learning.  \cite{gao2019batched} then generalized the result to $K$-armed bandits (using the same algorithm) and obtained a tight $\Theta(\log \log T)$ regret bound even when the batch sizes can be chosen dynamically. However, since MABs do not capture individuals' characteristics, these initial efforts \citep{perchet2016batched, gao2019batched} only operate on a population level and do not address the problem of personalized decision making, which severely limits their practical applicability. More recently, \cite{han2020sequential} has studied this problem in low-dimensional linear contextual bandits, and provides the first characterization of batch learning that incorporates personalized decision making.
In particular, a greedy ordinary least squares based algorithm is shown to achieve optimal regret (up to log factors).
Despite these strong guarantees, the results in \cite{han2020sequential} are insufficient for several reasons.
First, importantly, the setting in~\cite{han2020sequential} is limited to the low-dimensional regime where $d = O(\sqrt{T})$. Second, \cite{han2020sequential} studied static batch learning where the batch partitions must be 
chosen prior to the start of the decision-making process and cannot be changed thereafter. Consequently, this raises the critical issue of whether one can do better if \textit{dynamic batch learning} (where the decision maker can decide the next partition based on the data observed so far) is allowed, a question whose answer is not at all obvious.
Third,  \cite{han2020sequential} works exclusively with Guassian contexts, and its proofs rely on such Guassianity, which thus limits its applicability. In constrast, our goal in this paper is to delineate---in the high-dimensional sparse setting---the performance of dynamic batch learning by providing theoretical characterizations.
Additionally, when restricted to the low-dimensional setting (by taking $d= s_0$) with batch constraints, our results provide a strict generalization of~\cite{han2020sequential} on several fronts when the underlying contexts are stochastically generated (\cite{han2020sequential} also investigated adversarially generated contexts, which we do not study here): we study dynamic batch learning and we deal with general sub-Gaussian contexts (with diversity condition).
Consequently, although our goal lies in understanding dynamic batch learning under high-dimensional sparsity,  our results are also state-of-the-art in low dimensions as well.

%
\section{Problem Formulation}
\label{sec:problem_formulation}
We start with some useful notation that will be used throughout the paper. 
For a positive integer $n$, $[n]$ denotes the set $\{1, 2, \dots, n\}$; 
$\mathbb{S}^{n-1}$ denotes the $(n-1)$-dimensional unit sphere; 
$\Delta\mathbb{S}^{n-1}$ denotes the $(n-1)$-dimensional 
sphere with radius $\Delta$, for a given $\Delta > 0$;
$|S|$ denotes the cardinality of the set $S$ and $S^c$ 
denotes the complement of $S$. \revise{For a vector $v$ and 
a non-negative integer $q$, $\|v\|_q$
denotes the $\ell_q$ norm of $v$. For any positive 
semi-definite matrix $A$, $\lambda_{\min}(A)$ denotes
its smallest eigenvalue, and $\lambda_{\max}(A)$ its
largest eigenvalue.} 
We now move on to the formulation of the problem.

\subsection{High-Dimensional Sparse Linear Contextual Bandits}
Let $T$ denote the time horizon, $d$ the feature dimension 
and $K$ the number of arms. At $t\in [T]$, the decision maker 
first observes a set of $K$ $d$-dimensional feature vectors 
(i.e. contexts) $\{x_{t,a}\}_{a\in[K]}$.
If the decision maker selects action $a \in [K]$, then a reward 
$r_{t,a} \in \bR$ is incurred: $r_{t,a} =  x_{t,a}^\top \theta^\star + \xi_t,$
where $\theta^\star \in \bR^d$ is the underlying unknown parameter vector 
and $\{\xi_t\}_{t=0}^{\infty}$ is a sequence of  \textbf{iid} 
zero-mean 1-sub-Gaussian random variables: 
$\Expect[e^{\lambda \xi_t}] \le e^{\frac{\lambda^2}{2}}, \forall \lambda \in \bR$ 
(note that the constant $1$ is without loss of generality). 
Hereafter, we shall call this model \textsf{Model-C}.

We note here that in the contextual bandit literature, an 
alternative model with a set of underlying unknown $d$-dimensional parameters
$\{\theta^\star_a\}_{a \in [K]}$ is sometimes considered. In the alternative
model, at time $t\in[T]$, the decision maker observes a $d$-dimensional
context $x_t$, and if action $a\in[K]$ is chosen, the incurred
regret is $r_{t,a} = x_t^\top \theta^\star_a + \xi_t$, where
$\{\xi_t\}_{t=1}^\infty$ is similarly a sequence of \textbf{iid}
zero-mean 1-sub-Gaussin random variables. We refer to this alternative
model as \textsf{Model-P}.

Both models have been widely used in previous literature. 
For example, \modelc~is adopted 
in~\citet{han2020sequential,oh2021sparsity} and 
\modelp~in~\citet{bastani2021mostly,bastani2020online}.
The two models are in fact equivalent
in the following sense: given \modelc, one can write 
$\tilde{x}_t = (x_{t,1},\ldots,x_{t,K})$ and 
$\tilde{\theta}^*_a = (0,\ldots,\theta^*,\ldots,0)$,
and equivalently express $r_{t,a} = \tilde{x}_t^\top \tilde{\theta}^*_a + \xi_t$.
Conversely, given \modelp, 
we can let $\tilde{x}_{t,a} = (0,\ldots,x_t,\ldots,0)$ and 
$\tilde{\theta}^* = (\theta^*_1,\ldots,\theta^*_K)$. Then we have
$r_{t,a} = \tilde{x}_{t,a}^\top \tilde{\theta}^* + \xi_t$.
In this paper, we mainly focus on \modelc, while we shall 
also state parallel results under \modelp~in 
Appendix~\ref{sec:alternative_proof}.

\subsection{Assumptions}
Without loss of generality (via normalization), 
we assume $\|\theta^\star\|_2\le 1$;  the 
contexts $\{x_{t,a}\}_{a \in [K]}$ are 
random vectors \textbf{iid} drawn from 
a ($Kd$-dimensional) joint distribution
each time: the independence is across time, 
but for each $t$, $x_{t,a}$'s can be 
arbitrarily correlated across different $a$'s.
We denote by $a_t$ and $r_{t, a_t}$ the (random) action chosen and the (random) 
reward incurred at time $t$: $a_t$ is random because either it is 
randomly selected or the contexts $\{x_{t,a}\}_{a \in[K]}$ 
themselves are random, or both.
We impose the following mild conditions on the context distribution:

\begin{assumption}[Sub-Guassianity]\label{assumption:contexts}
For $\forall a \in [K]$, the marginal
distribution of $x_{t,a}$ is $1$-sub-Gaussian, 
i.e., $\bE[X] = 0$ and $\bE\big[\exp(v^\top X)\big]
\le \exp\big({\|v\|^2}/{2}\big)$, 
for $\forall v \in \mathbb{S}^{d-1}$.
\end{assumption}

\begin{remark}
Since bounded contexts are automatically sub-Gaussian, this assumption 
is more general than the bounded contexts assumption commonly adopted 
in the contextual bandits literature~\citep{bastani2017mostly, 
wang2018minimax, kim2019doubly, bastani2020online}. 	
\end{remark}	

\begin{assumption}[Diverse covariate]\label{assumption:rdi}
There are (possibly $K$-dependent) positive 
constants $\gamma(K)$ and $\rho(K)$, such that 
for any $\theta \in \mathbb{R}^d$ and any unit vector
$v\in \mathbb{R}^d$,  there is 
$\Prob\big((v^\top x_{t,a^*})^2  \ge \gamma(K)\big) \ge \rho(K)$,
where $a^*  = \argmax{a \in [K]}~x_{t,a}^\top \theta$.
\end{assumption}

\begin{remark}
The above assumption ensures there is sufficient explaration 
even with a greedy algorithm (it is also the key condition 
used in~\cite{han2020sequential} for the greedy algorithm there).
We shall provide a thorough discussion on sufficient conditions
for Assumption~\ref{assumption:rdi} in Section~\ref{sec:suff_cond}.
\end{remark}

In low dimensions~\citep[e.g.,][]{auer2002using, chu2011contextual},
regret bounds of $\tilde{\Theta}(\sqrt{dT})$---which are minimax optimal up to log factors---have been obtained under upper confidence bound based algorithms such as LinREL in~\cite{auer2002using} or LinUCB in~\cite{chu2011contextual}. However, 
these algorithms and their Thompson sampling counterpart LinTS in~\cite{AG2013b} (which performs well empirically but often exhibit slightly worse regret bounds) cease to be effective in the high-dimensional regime as mentioned in the introduction. Of course, it's important to point out that absence of any further structure, 
$\tilde{\Theta}(\sqrt{dT})$ is the optimal regret bound and hence the best one can hope for even when $d$ is very large.  In this paper, we tackle this problem in the presence of sparsity, where only a few covariates influence rewards despite a large number of ambient covariates.
In particular, we study the linear contextual bandits problem in the high-dimensional sparse regime: 
high-dimensional in the sense that $d$ is large compared to $T$ (the number of samples available in the entire learning horizon is small compared to the context dimension) and sparse in the sense that the underlying linear model is sparse: $\|\theta^\star\|_0 \ll d$.
We quantify them next.

\begin{assumption}[Sparsity in High-Dimension]\label{assumption:cov}
$d = \mathbf{Poly}(T)$ 
with sparse parameters: there exists some $\varepsilon > 0$ such that
$\|\theta^\star\|_0 \le s_0 = O(T^{1-\varepsilon})$.
\end{assumption}

\begin{remark}
In statistical learning, a regime is considered high-dimensional if the dimension of the model is larger than the number of samples \citep{wainwright2019}. In our setting, this would translate to $d > T$. Consequently, our assumption that $d$ can be any polynomial of $T$ covers very high-dimensional regimes. Further, learning becomes infeasible when $d$ becomes even larger to, say, exponential in $T$, since
a $\log d$ factor is present in the estimation accuracy even in the simple \textbf{iid} supervised learning setting \citep{hastie2015statistical}, which translates to a linear dependence on $T$.
The sparsity requirement formalizes the precise requirement of $\|\theta^\star\|_0 \ll d$.
Note that one can view  $\|\theta^\star\|_0$ (or its upper bound $s_0$) as the ``intrinsic dimension" of the linear contextual bandits; consequently $s_0$ should certainly be sublinear in $T$ in order for learning to be effective.  A typical regime of sparsity in statistical learning is $s_0 = O(\log d)$ \citep{wainwright2019}, which 
 certainly meets the $s_0 = O(T^{1-\epsilon})$ requirement since  $d = \mathbf{Poly}(T)$.
 Finally, in the above assumption, we posit that an upper bound $s_0$ on the sparsity level 
 is known to the decision maker. This assumption is standard and adopted for all the existing high-dimensional sparse linear contextual bandits~\citep{wang2018minimax, kim2019doubly, bastani2020online} in their algorithm designs.
\end{remark}	

Finally, we work in the regime where the action set size $K$ is not too large:

\begin{assumption}[Not Many Actions]\label{assumption:actions}
\revise{
The number of actions $K$ satisfy the following two upper bounds:
$\frac{\log K}{\gamma(K)\rho(K)} = O( d / s_0)$
and $\frac{\log K}{\gamma(K) \rho^3(K)} = O(\sqrt{T^{1-\beta}/s_0})$ for some $\beta>0$.
}
\end{assumption}
In our motivating applications, $K$ is small (e.g. a constant number of actions) and easily satisfies this requirement, although this assumption can tolerate a much larger number of actions since $s_0 \ll d$.  In practice, this regime typically suffices unless the number of actions is combinatorially large or when the action set is continuous, which would require a separate treatment.

\subsection{Covariate Diversity Condition}
\label{sec:suff_cond}
In this section we expand on Assumption~\ref{assumption:rdi} 
and provide a list of sufficient conditions for it.
\begin{lemma}\label{lemma:diversity_condition}
\revise{The following are sufficient conditions for Assumption~\ref{assumption:rdi}.
\begin{enumerate}
\item\label{itm:cond1} If for each $a\in[K]$, $x_{t,a} \sim \calN(0,\Sigma)$ marginally, 
    where $\lambda_{\min}(\Sigma) > 0$, then Assumption~\ref{assumption:rdi}
    holds with $\gamma(K) = \frac{\lambda_{\min}(\Sigma)}{16}$
    and $\rho(K) = \frac{1}{10}$.
  \item\label{itm:cond2} If there exists constants $\alpha,c > 0$ such that
  for each $a \in [K]$ and any unit vector $v \in \mathbb{R}^d$,
\begin{align}
\label{eq:mgf_condition}    
    \Expect\Big[\exp\big(-\delta \cdot (v^\top x_{t,a})^2\big)\Big] 
    \le c\cdot\delta^{-\alpha},
\end{align}
  for any $\delta > 0$, then $\gamma(K) = \frac{\alpha}{e}\cdot (2cK)^{-1/\alpha}$ 
  and $\rho(K) = \frac{1}{2}$. 
\item\label{itm:cond3} If there exists a constant $\Lambda >0$, such that
   for each $a\in [K]$ and any unit vector $v \in \mathbb{R}^d$, 
   $v^\top \Expect \big[ x_{t,a} x_{t,a}^\top \big] v \ge \Lambda$
   and $\textnormal{Var}\big((v^\top x_{t,a})^2) \le \frac{\Lambda^2}{8K}$,
   then $\gamma(K) = \frac{\Lambda}{2}$ and $\rho(K) = \frac{1}{2}$.
\item\label{itm:cond4} When $K=2$, if there exists a constant $\Lambda > 0$ such that for any
   $a\in[K]$,  any unit vector $v\in \mathbb{R}^d$, $v^\top \Expect[x_{t,a} x_{t,a}^\top]v \ge \Lambda$,
   and if there exists a constant $\nu>0$ such that the joint distribution of 
   $(x_{t,1},x_{t,2})$ satisfies
   $p(x_{t,1}, x_{t,2}) \ge \nu \cdot p(-x_{t,1}, -x_{t,2})$, then 
   $\gamma(K) = \frac{\Lambda}{2}$ and 
   $\rho(K) = \frac{\nu\Lambda^2}{64}$.
\item\label{itm:cond5} When $K > 2$, suppose the following three conditions hold:
  \begin{enumerate} 
  \item there exists a constant $\Lambda > 0$, such that
    for any $a\in[K]$, any unit vector $v \in \mathbb{R}^d$, 
    $v^\top \Expect[x_{t,a}x_{t,a}^\top]v \ge \Lambda$;
  \item there exists a constant $\nu_1 > 0$ such that the joint
    distribution of $(x_{t,1},\ldots,x_{t,K})$ satisfies
    $p(x_{t,1},\ldots,x_{t,K}) \ge \nu_1 \cdot p(-x_{t,1},\ldots,-x_{t,K})$;
  \item there exists a (possibly $K$-dependent) 
    constant $\nu_2(K)>0$ such that for any $\theta \in \mathbb{R}^d$,
    any permutation of $[K]$ denoted by $\{\pi_1,\ldots,\pi_K\}$ and any unit vector $v\in \RR^d$,
    we have for any $a \in [K]$
   \begin{align*}
        &\nu_2(K) \cdot \Prob\Big((v^\top x_{t,\pi_{a}})^2 \ge \frac{\Lambda}{2}, x_{t,\pi_1}^\top \theta \le \ldots \le x_{t,\pi_K}^\top \theta\Big)\\
        \le &\Prob\Big((v^\top x_{t,\pi_{1}})^2 \ge \frac{\Lambda}{2}, x_{t,\pi_1}^\top \theta \le \ldots \le x_{t,\pi_K}^\top \theta\Big) +
        \Prob\Big((v^\top x_{t,\pi_{K}})^2 \ge \frac{\Lambda}{2}, x_{t,\pi_1}^\top \theta \le \ldots \le x_{t,\pi_K}^\top \theta\Big).
      \end{align*}
\end{enumerate}
Then Assumption~\ref{assumption:rdi} holds with $\gamma(K) = \frac{\Lambda}{2}$ 
and $\rho(K) = \frac{\nu_1\nu_2(K) \Lambda^2}{128}$.
\end{enumerate}
}
\end{lemma}
\revise{
The proof of Lemma~\ref{lemma:diversity_condition}  
can be found in Appendix~\ref{appx:proof_diversity}.
Broadly speaking, the above sufficient conditions 
can be categorized into two groups: Conditions~\ref{itm:cond1}-\ref{itm:cond3} are 
assumptions on the marginal distribution of $x_{t,a}$
while Conditions~\ref{itm:cond4}-\ref{itm:cond5} are on the joint distribution 
of $\{x_{t,a}\}_{a\in[K]}$. More specifically, Condition~\ref{itm:cond1} 
is adopted from~\citet{han2020sequential} and uses the proper of 
of the Gaussian distribution; Condition~\ref{itm:cond2}
characterizes a large class of distributions:
in particular, if there exists a constant $\zeta > 0 $, 
such that for any unit vector $v$ and any $a\in[K]$, 
the distribution of $v^\top x_{t,a}$ is bounded by $\zeta$, 
then this condition holds with $\alpha = 1/2$ and 
$c = \zeta\sqrt{\pi}/2$. The bounded density condition
is similarly considered in~\citet{li2021regret} and
is quite flexible: for example, when the coordinates
$x_{t,a,j}$ are mutually independent across $j$, and the 
density of $x_{t,a,j}$ is bounded by $\zeta$, the density
of $v^\top x_{t,a}$ is bounded by $\sqrt{2}\zeta$ for any
unit vector $v$~\citep[Theorem 1.2]{rudelson2015small}.
Condition~\ref{itm:cond3} requires the population 
covariance matrix of $x_{t,a}$ to be well-conditioned, 
and the variance of $(v^\top x_{t,a})^2$
to be relatively small. Condition~\ref{itm:cond4} is inspired by the
diversity condition considered in~\citet[Lemma 1]{bastani2017mostly},
and Condition~\ref{itm:cond5} by~\citet[Assumption 6]{oh2021sparsity}.}

We note that
these assumptions are on the covariates (as opposed to those
on the underlying model), as a result of which they are always testable (the 
covariates $\{x_{t,a}\}_{a\in[K]}$ can be fully observed). This 
fact is particularly appealing to practitioners.

\subsection{Dynamic Batch Learning}
In the standard online learning setting, the decision maker
immediately observes the reward $r_{t, a_t}$ after selecting 
action $a_t$ at time $t$. After observing $r_{t,a_t}$, the 
decision maker can immediately incorporate this information
in adapting her decision for action-selection at $t+1$. In particular, 
the decision maker can utilize all the historical information---including 
contexts $\{x_{\tau,a}\}_{\tau \le t, a\in[K]}$ and rewards 
$\{r_{\tau,a_\tau}\}_{\tau\le t-1}$---in deciding what action $a_t$ to take at current time $t$ . 

In contrast, we consider a \textit{dynamic batch learning} setting, where the decision maker is only allowed to partition the $T$ units into (at most) $M$ batches, and the reward corresponding to each unit in a batch can only be observed at the end of the batch. 
Note that the decision maker can provision the partition \textit{dynamically}:
she can decide on how large the next batch is based on what she has observed 
in all the previous batches, which include all the contexts, the selected 
actions and the corresponding rewards. Note that the initial batch size 
is chosen without observing anything.

Formalizing the above, given a maximum number of batches $M$, 
a dynamic batch learning algorithm \textbf{Alg} $= (\calT, \pi)$
 has the following two components:
\begin{enumerate}
\item A \emph{dynamic grid} $\calT=\{t_1,t_2,\cdots,t_M\}$, with $0 = t_0 < t_1<\cdots<t_M=T$, where each $t_i$ is dynamically chosen at the end of $t_{i-1}$ based on all the historical information available up to and including $t_{i-1}$. More specifically, prior to starting the decision making process, the decision maker decides on $t_1$, which indicates the length of the first batch. Having selected actions for each time in the first batch, the decision maker observes all the corresponding rewards at the end of $t_1$.
  Based on such information---including $\{a_t\}_{t=1}^{t_1}, \{x_{t,1},\ldots,x_{t,K}\}_{t=1}^{t_1}$ and $\{r_{t,a_t}\}_{t=1}^{t_1}$---the decision maker then decides on what $t_2$ is. 
	This dynamic grid partitioning process continues, and the decision maker always selects where the next batch ends at the end of current batch.

	\item A sequence of policies $\pi = (\pi_1, \pi_2, \dots, \pi_T)$ such that each $\pi_t$ can only use reward information from all the prior batches and the contexts that can be observed up to $t$. 
	That is, for a given $t$, if it lies in the $i$-th batch ($t_{i-1} < t \le t_i)$, then the policy to be used
	at $t$ can utilize all the observed rewards from $\tau=1$ to $\tau = t_{i-1}$, all the selected actions from 
	$\tau=1$ to $\tau = t-1$ and all the contexts information from $\tau = 1$ to $\tau = t$.
\end{enumerate}

\begin{remark}
	Two special cases of a dynamic batch learning algorithm are worth mentioning.
	First, when the grid is fixed in advance---a static $\calT$ is chosen completely at the beginning and not adapted during the learning process---we obtain a static batch learning algorithm, which is the class of algorithms considered in~\cite{han2020sequential}.
	Second, a further special case is the fixed grid $\calT=\{1,2,\cdots,T\}$ (i.e. $M = T$). This corresponds to the standard online learning setting where the decision maker need not select a grid. 
	We also point out that $M=1$ is the other end of the spectrum, where no adaptation is allowed. 
	In this case, irrespectively of what one does, worst-case regret is always linear in $T$ and regret (as defined next in Definition~\ref{def:regret}) is a meaningless (and thus the wrong) metric. Instead, one should adopt an offline learning viewpoint and adopt generalization error as the metric. This (offline learning in contextual bandits) would be an entirely new topic, and it has been well-studied by a growing literature; see~\citet{zhao2014doubly, JMLR:v16:swaminathan15a, kitagawa2018should, kallus2018confounding, deep-learning-logged-bandit-feedback} and references therein. 
	
\end{remark}

To measure the performance of a dynamic batch learning 
algorithm \textbf{Alg}, we compare the cumulative 
reward obtained by \textbf{Alg} to that obtained 
by an \textbf{optimal} policy (an oracle that 
knows $\theta^\star$). This is formalized by regret, as defined next:

\begin{definition}\label{def:regret}
Let \textbf{Alg} $= (\calT, \pi)$ be a dynamic batch learning algorithm.
The regret of \textbf{Alg} is:
\begin{align}\label{eq.regret}
R_T(\textbf{Alg}) \triangleq \sum_{t=1}^T \Big( \max_{a\in [K]}x_{t,a}^\top \theta^\star  - x_{t,a_t}^\top \theta^\star \Big),
\end{align}
where $a_1, a_2, \dots, a_T$ are actions generated by \textbf{Alg} in the online decision making process. 	
\end{definition}	

\begin{remark}
The regret defined above is the same as used in standard online learning, but the feedback in our setting is much more restricted: batches induce delays in obtaining reward feedback, and hence the decision maker cannot immediately incorporate the feedback into his subsequent decision making process.
Consequently, all else equal, the regret will be \textit{a priori} much larger when the decision maker is constrained to work with only a small number of batches.
 	
\end{remark}

\section{Fundamental Limits: Regret Lower Bound}\label{sec:lowerbnd}
In this section, we present the minimax regret lower bound that
characterizes the fundamental learning limits of dynamic batch
learning in high-dimensional sparse linear contextual  bandits.

\begin{theorem}\label{thm:dynamiclowerbnd}
Fix any $s_0,d$ and $T$. 
Let $K = \log(T/s_0)$ and consider the
problem
$x_{t,a} \sim \calN(0,I_d), \forall a\in[K], \forall t\in[T]$, where the contexts are
independence across $t$.  Then
for any $M \le T$ and any dynamic batch learning
algorithm $\mathbf{Alg}$, we have
\begin{align}
  \sup_{\theta^\star: \|\theta^\star\|_2\le 1,\|\theta^{\star}\|_0\le s_0} \Expect_{\theta^\star}\big[R_T(\mathbf{Alg})\big]
  \ge c\cdot \max \bigg(M^{-4}2^{-7M/2} \cdot\sqrt{Ts_0}\cdot\Big(\dfrac{T}{s_0}\Big)^{\frac{1}{2(2^M-1)}}, \sqrt{Ts_0}\bigg),\label{eq:lowerbound}
\end{align}
where $\Expect_{\theta^\star}$ denotes taking expectation w.r.t.~the distribution based on the parameter $\theta^\star$,
and $c>0$ is a numerical constant independent of $(T,M,d, s_0)$.
\end{theorem}
We shall present the main steps in the proof
Theorem~\ref{thm:dynamiclowerbnd} here and
defer the details to Appendix~\ref{appendix:proof_lower_bnd}.
\begin{remark}
There are two terms on the right-hand side of Equation~\eqref{eq:lowerbound}: the first term characterizes the dependence on $M$ and the second term 
corresponds to the regret lower bound for the standard online learning setting. 
We have mentioned in the previous section that standard online learning (corresponding to $M=T$)
is a simple special case of dynamic batch learning. 
Since a larger $M$ provides better opportunities for adapting the decision-making process,
a dynamic batch learning problem will only have worse regret compared to standard online learning setting.
Consequently, a lower bound to standard online learning is immediately a lower bound
to dynamic batch learning.
Of course, the lower bound to dynamic batch learning will get worse, particularly when $M$ is small (corresponding to limited chances to adapt one's decisions), hence the first term on the right-hand side of Equation~\eqref{eq:lowerbound}. We see that the break-even point---where the two lower bound terms equalize (up to log factors)---occurs at $M=O\big(\log{\log{(T/s_0)}}\big)$. Consequently, taking into account the log terms, we see that when $M < O\big(\log{\log{(T/s_0)}}\big)$, the first term dominates the lower bound, while the second term dominates the lower bound once $M$ gets larger than $\Theta\big(\log{\log{(T/s_0)}}\big)$.
\end{remark}

\begin{remark}
\revise{In Theorem~\ref{thm:dynamiclowerbnd}, the example used to show the lower bound satisfies 
Assumption~\ref{assumption:rdi} with $\gamma(K) = \rho(K) = O(1)$ and also satisfies Assumption~\ref{assumption:contexts}}.
Further, since the lower bound is established for any $(s_0, d, T)$, it obviously holds for the
regime given in Assumption~\ref{assumption:cov} (since taking the supremum in a bigger set only results in a no-smaller lower bound).
If in addition $s_0 \log \log (\frac{T}{s_0}) = O(d)$ (a regime where $d$ is slightly larger than $s_0$),
then Assumption~\ref{assumption:actions} is also satisfied for the problem construction in Theorem~\ref{thm:dynamiclowerbnd}. 
Consequently, the lower bound holds under all four assumptions, under which the upper bound is subsequently established to match the lower bound (up to log factors).
Additionally, when $s_0 = d$ (the standard low-dimensional regime),
our lower bound still holds, hence providing a fundamental limit that is not known even in that important
special case. We do point out that in the low-dimensional regime where $s_0 = d$, unless $K = O(1)$, Assumption~\ref{assumption:actions} does not hold, in which case the subsequent upper bound
does not apply. Of course, this is not an issue at all since~\cite{han2020sequential} already provided
an upper bound for the low-dimensional setting under static batch design and matches the dynamic batch lower bound here, thereby completing the picture that even in the low-dimensional case and even when dynamic batch is allowed, one cannot do better than the static batch learning characterized in~\cite{han2020sequential} . 
\end{remark}
 

\textbf{Main Proof Outline of Theorem~\ref{thm:dynamiclowerbnd}} 
A key difficulty of the proof is that the grid is determined 
adaptively based on the observations from the previous batches.
We briefly highlight the main proof steps here, each of which 
will be elaborated and rigorously formalized  in a subsequent subsection. 

We start from the regime of small $M$.
Suppose $M = O\big(\log{\log{(T/s_0)}}\big)$. Define
for any $m\in[M]$
\begin{align*}
T_m = \bigg\lfloor s_0\cdot \Big(\dfrac{T}{s_0}\Big)^{\frac{1-2^{-m}}{1-2^{-M}}} \bigg\rfloor,
\qquad 
\Delta_m = \dfrac{1}{24 \cdot M^2 \cdot 2^{3M}} 
\cdot \Big(\dfrac{T}{s_0}\Big)^{-\frac{1-2^{1-m}}{2(1-2^{-M})}}.
\end{align*}
Considering $K = 2^M$ arms, we shall construct a prior $Q$ 
for $\theta^\star$ and examine the 
regret under $Q$. Here, the prior is carefully designed 
such that for any $m\in[M]$, we can divide the $2^M$ arms into $2^{M-1}$
pairs such that the diffrence between each pair of arms 
is approximately the scale of $\Delta_m$; the values of $T_m$ and $\Delta_m$
are chosen such that the number of observations up to $T_{m-1}$ 
is simply too few for the decision maker to distinguish the
two arms in a pair (and learn a effective policy).
Consequently, when the decision maker
deploys this (ineffective) policy to this batch 
(from $T_{m-1} +1$ to $T_m$), even when restricted 
to the portion from $T_{m-1} +1$ to $T_m$,
the total expected regret incurred---$(T_m - 
T_{m-1})\cdot \Delta_m$---is still large.
Section~\ref{sec:hard_inst_modelc} details the
construction of the prior, and 
Section~\ref{sec:regret_decomposition_modelc} connects the worst-case
regret to that under $Q$.

Given an $\textbf{Alg}$, of course its grid design $\{t_1,\ldots,t_M\}$
can be different from our ``ideal'' design $\{T_1,\ldots,T_M\}$.
However, we now define for each $m\in[M]$ the ``bad'' event 
$B_m = \{t_{m-1}\le T_{m-1}<T_m\le t_m\}$:
$B_m$ is a ``bad" event because, when $B_m$ occurs, 
the number of observations up to $t_{m-1}$ is too few (since $t_{m-1}\le T_{m-1}$) 
to distinguish pairs of arms that are $\Delta_m$ apart and
learn an effective policy; when this (ineffective) policy is applied to
this batch (from $t_{m-1} +1$ to $t_m$), the total expected regret incurred is still large (since $t_m\ge T_m$).
In fact, we don't need a bad event to happen surely to guarantee 
that the total expected regret incurred is large: a bad event need only happen with a large enough probability to meet this purpose 
(with the probability taken over the randomness of the observations and the that 
of the parameters $\theta^*$).  
Section~\ref{sec:regret_bad_event_modelc} formalizes and establishes this step:
if at least one $B_m$ occurs with a large enough probability,
then we obtain the desired final regret lower bound.

Section~\ref{sec:bad_event_high_prob_modelc} is devoted to establishing that ``if'' is true.  
Note that by a simple combinatorial argument, at least one of the $B_m$ events will happen (under the convention that $t_0 = 0$, and since $t_M = T$, we are throwing $M-1$ points $t_1, t_2, \dots, t_{M-1}$ into the $M$ intervals partitioned by $0, T_1, T_2, \dots, T_{M-1}, T$, hence the conclusion). 
In other words, $\{B_m\}_{m\in[M]}$ constitute a (non-disjoint) partition of the whole probability space.
Hence, at least one bad event will happen with probability greater than $1/M$.

Finally, Section~\ref{subsec:lower3} establishes the
lower bound for standard (fully) online learning ($M=T$):
since $M \le T$ in dynamic batch learning, this is 
clearly always a lower bound to the regret, which
corresponds to the second term of the right-hand 
side of Equation~\eqref{eq:lowerbound}. Taken 
together, these three steps complete the picture.
We next dive into more details and begin with some useful notation.

\subsection{Construction of the prior}
\label{sec:hard_inst_modelc}
Let $\tilde{s}_0 = \lfloor s_0 \cdot 2^{-M} \rfloor \cdot 2^M$,
and we have $s_0 - \ts \le 2^M = O\big(\log(T/s_0)\big)$.
Next, we divide $[\ts]$ into consecutive subgroups at different
levels of ``resolution''.
At the first level of resolution, we divide $[\ts]$ into two consecutive 
group of equal sizes, denoted by $I_0$ and $I_1$ respectively, where
\begin{align*}
  I_0 = \big \{1,\cdots,\tfrac{1}{2}\ts \big\},\quad 
  I_1 = \big\{\tfrac{1}{2}\ts+1,\cdots,\ts \big\};
\end{align*}
at the second level of resolution, we further divide
$I_0$ into two equal subgroups $I_{00}$ and $I_{01}$,
and $I_1$ into $I_{10}$ and $I_{11}$, where
\begin{align*}
  &I_{00} = \big \{1,\cdots,\tfrac{1}{4}\ts \big\},\quad 
   I_{01} = \big\{\tfrac{1}{4}\ts+1,\cdots,\tfrac{1}{2}\ts \big\},\quad
   I_{10} = \big \{\tfrac{1}{2}\ts+1,\cdots,\tfrac{3}{4}\ts \big\},\quad 
   I_{11} = \big\{\tfrac{3}{4}\ts+1,\cdots,\ts \big\}.
\end{align*}
Repeating the above steps, at the $M$-th level of resolution
we obtain $2^M$ subgroups of equal sizes:
\begin{align*}
I_{0\cdots 00} = \big\{1,\cdots,\tfrac{1}{2^M}\ts\big\},
~I_{0\cdots 01} = \big\{\tfrac{1}{2^M}\ts+1,\cdots,\tfrac{1}{2^{M-1}}\ts\big\},
~\cdots,~
I_{1\cdots 11} = \big\{\ts - \tfrac{1}{2^M}\ts+1,\cdots,\ts\big\}.
\end{align*}
To summarise, for any $m\in[M]$, a subgroup at the $m$-th 
level of resolution is represented by a $m$-dimensional 
vector $\sigma$ in $\Pi(m)\,:=\,\{0,1\}^m$.

Next, we construct the prior $Q$ on the true parameter 
$\theta$. Generate $\theta_1,\ldots,\theta_M$ independently 
from $\Unif(\mathbb{S}^{\frac{\ts}{2^M}})$.
For each $m\in[M]$, we define $\tth_m \in \RR^{\ts}$ in the 
following way: 
for each $\sigma \in \Pi(M)$,
\begin{align*}
  \tth_m(I_{\sigma}) = 
  \begin{cases}
     \theta_m & \sigma_m = 0,\\
    - \theta_m & \sigma_m = 1,
  \end{cases}
\end{align*}
where $\sigma_m$ refers to the $m$-th 
coordinate of $\sigma$.
As a concrete example, for $m=1$ and $2$,
\begin{align*}
  \tth_1 = (\underbrace{\theta_1,\cdots,\theta_1,}_{2^{M-1}\mbox{ items }}
  \underbrace{-\theta_1,\cdots,-\theta_1}_{2^{M-1} \mbox{ items }}) \quad
  \tth_2 = (\underbrace{\theta_2,\cdots,\theta_2}_{2^{M-2}\mbox{ items }},
  \underbrace{-\theta_2,\cdots,-\theta_2}_{2^{M-2} \mbox{ items }},
  \underbrace{\theta_2,\cdots,\theta_2,}_{2^{M-2}\mbox{ items }}
  \underbrace{-\theta_2,\cdots,-\theta_2}_{2^{M-2} \mbox{ items }}).
\end{align*}
Setting $\tth = \sum^M_{m=1}\Delta_m\theta_m$, we construct 
$\theta := f(\theta_1,\theta_2,\ldots,\theta_M)\in \bR^d$ by letting its first $\ts$ coordinates be $\tth$
and the others zero. It can checked that 
$\|\theta\|_2 = \|\tth\|_2 \le \sum_{m=1}^M \Delta_m  \|\tth_m\|_2 \le 1$
and $\|\theta\|_1 = \ts$.

We now proceed to specify the joint distribution of the $ K =2^M$ arms. 
For each $t\in [T]$, we first draw $x_t \sim \calN(0,I_d)$. 
To simplify the notation, we let $S = \{1,2,\ldots,\ts\}$
and $S^c = \{\ts+1,\ldots,d\}$. 
For each $a\in[K]$, we first let $x_{t,a}(S^c) = x_t(S^c)$.
It remains to specify the first $\ts$ coordinates of $x_{t,a}$.
To do so, we again divide $S$ into $2^M$ consecutive groups,
each represented by $\sigma \in \Pi(M)$,
and will specify the value of each group. 
Given an arm $a$, we can uniquely write 
$a = 1 + \sum^M_{m=1}a_{m} \cdot 2^{m-1}$
where $a_m \in \{0,1\}$ for
each $m\in[M]$; define a mapping
$\calM_a: \Pi(M)\mapsto \Pi(M)$,
where for any $m\in[M]$, 
$\calM_a(\sigma)_m = (1-a_m) \cdot \sigma_m + a_m \cdot (1-\sigma_m)$.
We then let $x_{t,a}(\sigma) = x_t(\calM_a(\sigma))$,
for any $\sigma \in \Pi(M)$.
For example, when $M = 2$, we have 
four arms, where
\begin{align*}
  &x_{t,1} = x_t,~x_{t,2} = (x_t(I_1),x_t(I_0),x_t(S^c)),\\
  &x_{t,3} = (x_t(I_{01}),x_t(I_{00}), x_t(I_{11}), x_t(I_{10}),x_t(S^c)),\\
  &x_{t,4} = (x_t(I_{11}),x_t(I_{10}), x_t(I_{01}), x_t(I_{00}),x_t(S^c)).
\end{align*}
By construction, for any $a \in [K]$, $x_{t,a} \sim \calN(0,I_d)$
marginally, thus satisfying Assumption~\ref{assumption:rdi}.

\subsection{Notation for Regret Decomposition}
\label{sec:regret_decomposition_modelc}
We streamline the notation for a regret decomposition that will be used throughout:
\begin{align*}
	\sup_{\theta^\star:\|\theta^\star\|_2\le 1,~\|\theta^\star\|_0\le s_0}\Expect_{\theta^\star}\big[R_T(\textbf{Alg})\big]
  \ge \Expect_Q\Expect_{\theta}\big[R_T(\textbf{Alg})\big] 
	= \sum^T_{t=1}\Expect_{Q}
  \Big(\Expect_x\Expect_{P^t_{\theta,x}}\Big[\max_{a\in [K]}
  ~x_{t,a}^\top\theta -x_{t,a_t}^\top \theta  \Big] \Big),
\end{align*}
where $\Expect_{Q}$ denotes taking expectation with respect to the prior $Q$ of $\theta$, 
$\Expect_x$ denotes taking expectation with respect to all the random contexts at all times
(note that it is both equivalent and conceptually simpler to imagine all the contexts
$x = \{x_{t,a}\}_{t\in[T], a \in [K]}$ have been drawn once for all ahead of time
before the decision-making process starts), and $P^t_{\theta,x}$ denotes the distribution 
of all observed rewards before time $t$ (and hence before the start of the current
batch that contains $t$) conditional on the parameter $\theta$ and the contexts $x$.
Note that per its definition, the distributions $P^t_{\theta,x}$ and $P^{t+1}_{\theta,x}$ are the same
if $t$ and $t+1$ belong to the same batch.

Recall that for each $j\in[2^M]$, we write $j = 1 + \sum^M_{m=1}j_m\cdot 2^{m-1}$.
Then for each $t\in[T]$ and any $m\in[M]$,
\begin{align}
\label{eq:regret_decomp}
& \max_{a\in [K]} \big( x_{t,a}^\top\theta - x_{t,a_t}^\top\theta\big)
= \sum_{j \in [K]} \Indc \{a_t = j\} \cdot  
\max_{a\in [K]}\big(x_{t,a}^\top \theta - x_{t,j}^\top \theta\big)\nonumber\\
\stackrel{\rm (a)}{=} &\sum_{j\in[K]:j_m = 0} 
\Indc \{a_t = j\} \cdot \max_{a\in [K]}
\big(x_{t,a}^\top \theta - x_{t,j}^\top \theta\big)
+ \Indc \{a_t = j + 2^{m-1}\} \cdot \max_{a\in [K]}
\big(x_{t,a}^\top \theta - x_{t,j+2^{m-1}}^\top \theta\big)\nonumber\\
\ge &\sum_{j\in[K]:j_m = 0} 
\Indc \{a_t = j\} \cdot \max_{a\in \{j,j+2^{m-1}\}}
\big(x_{t,a}^\top \theta - x_{t,j}^\top \theta\big)
+ \Indc \{a_t = j + 2^{m-1}\} \cdot \max_{a\in \{j,j+2^{m-1}\}}
\big(x_{t,a}^\top \theta - x_{t,j+2^{m-1}}^\top \theta\big)\nonumber\\
= & \sum_{j\in[K]:j_m = 0}
\Indc\{a_t = j\} \cdot \big(x_{t,j+2^{m-1}}^\top\theta - x_{t,j}^\top\theta\big)_+
+ \Indc\{a_t = j + 2^{m-1}\} \cdot 
\big(x_{t,j+2^{m-1}}^\top\theta - x_{t,j}^\top\theta\big)_-,
\end{align}
where in step (a) we categorize the arms into two 
groups by the value of $j_m$. For a $j$ such that
$j_m = 0$, we can write
\begin{align*}
  &x_{t,j+2^{m-1}}^\top\theta - x_{t,j}^\top\theta 
  = \sum_{\sigma\in\Pi(M)} x_{t,j+2^{m-1}}(I_{\sigma})^\top
  \theta(I_{\sigma}) - x_{t,j}(I_{\sigma})^\top \theta(I_\sigma)\\
  = &\sum_{\sigma\in\Pi(M):\sigma_m =0} x_{t,j+2^{m-1}}(I_{\sigma})^\top
  \theta(I_{\sigma}) - x_{t,j}(I_{\sigma})^\top \theta(I_\sigma)
    +\sum_{\sigma\in\Pi(M):\sigma_m=1} x_{t,j+2^{m-1}}(I_{\sigma})^\top
  \theta(I_{\sigma}) - x_{t,j}(I_{\sigma})^\top \theta(I_\sigma)\\
  = & 2 \Delta_m\cdot \theta_m^\top
  \Big(\sum_{\sigma \in \Pi(M):\sigma_m = 1} x_{t}(I_\sigma) 
  - \sum_{\sigma \in \Pi(M): \sigma_m = 0} x_t(I_{\sigma})\Big).
\end{align*}
To simplify the notation, we define
\begin{align*}
d_{m,t} = \sum_{\sigma \in \Pi(M):\sigma_m = 2} x_{t}(I_\sigma) 
- \sum_{\sigma \in \Pi(M): \sigma_m = 1} x_t(I_{\sigma}),\quad
u_{m,t} = \frac{d_{m,t}}{\|d_{m,t}\|_2},
\end{align*}
and $\calA_{m} = \big\{j\in[K]:j_m = 0 \big\}$. With the above,
we continue decomposing the regret
\begin{align*}
  \eqref{eq:regret_decomp} = &2\Delta_m \sum_{j\in \calA_m}
  \Indc\{a_t = j\} \cdot \big(d_{m,t}^\top\theta_m\big)_+ 
  + \Indc\{a_t = j+2^{m-1}\} \cdot
  \big(d_{m,t}^\top\theta_m\big)_-\\
  = &2\Delta_m \cdot \Indc\{j\in\calA_m\}
  \cdot \big(d_{m,t}^\top\theta_m\big)_+
  + \Indc\{j\in \calA_m^c\} \cdot \big(d_{m,t}^\top \theta_m\big)_-.
\end{align*}
As a result, we have
\begin{align}
  \label{eq:regret_decomp_2}
  &\Expect_{Q}\Expect_{P^t_{\theta,x}}
  \Big[\max_{a\in[K]}(x_{t,a}^\top\theta - x_{t,a_t}^\top\theta )\Big]\nonumber\\
  \ge &2\Delta_m \cdot \Expect_{Q}\Big[\big(d_{m,t}^\top \theta_m\big)_+ 
    \cdot \Expect_{P^t_{\theta,x}} \big[\Indc\{a_t\in\calA_m\}\big]
  + \big(d_{m,t}^\top\theta_m\big)_-\cdot
\Expect_{P^t_{\theta,x}}\big[\Indc\{a_t\in\calA_m^c\}\big]\Big],
\end{align}
where we note that conditioned on $\theta$ and $x$, $a_t$ depends on the distribution of observed rewards
$P^t_{\theta,x}$ (hence the inner expectation is taken with respect to this distribution).
Through a change of measure, we define two new probability measures via
\begin{align*}
  \dfrac{dQ^+_{m,t}}{dQ}(\theta) = \dfrac{(d_{m,t}^\top\theta_m)_+}{Z_m(d_{m,t})},
~ \dfrac{dQ^-_{m,t}}{dQ}(\theta) = \dfrac{(d_{m,t}^\top\theta_m)_-}{Z_m(d_{m,t})},
\end{align*}
where $Z_m(d_{m,t}) = \EE_Q[(d_{m,t}^\top \theta_m)_+] = \EE_Q[(d_{m,t}^\top \theta_m)_-]$ 
is a common normalizing factor. Then,
\begin{align*}
	&\Expect_{Q}\Expect_{P^t_{\theta,x}}
  \Big[\max_{a\in[K]} \big(x_{t,a}^\top\theta - x_{t,a_t}^\top\theta \big)\Big] 
  \ge &2\Delta_m Z_m(d_{m,t}) 
  \cdot\Big(\Expect_{P^t_{\theta,x} \circ Q^+_{m,t}} \big[\Indc\{a_t\in \calA_m\}\big]
  +  \Expect_{P^t_{\theta,x} \circ Q^-_{m,t} }\big[\Indc \{a_t=\calA_m^c\} \big]\Big),
\end{align*}
where $P^t_{\theta,x} \circ Q^+_{m,t}$ (resp. $P^t_{\theta,x} \circ Q^-_{m,t}$)
is a mixed distribution:  $\theta$ is drawn from $Q^+_{m,t}$ (resp. $Q^-_{m,t}$)
and observed rewards are then drawn from $P^t_{\theta,x}$. Note that 
$Z_m(\cdot)$ is a function and $Z_m(d_{m,t})$ emphasizes that the 
common normalizing factor depends on $d_{m,t}$. 

Reparametrizing of the regret in terms of the two newly-defined priors
allows us to connect the regret with the distance between measures, from which
lower bounds can be established with information-theoretic tools.

\subsection{Regret lower bound when a ``bad'' event happens with large probability}
\label{sec:regret_bad_event_modelc}
When a ``bad'' event $B_m$ is likely to
happen under prior $Q$, large regret follows: 
\begin{lemma}
If there exists $m \in [M]$, such that
\begin{align}
  \sum^{T_m}_{t=T_{m-1}+1} \Expect_x \Big[Z_m(d_{m,t}) \cdot 
  \Expect_{P_{\theta,x}\circ Q^+_{m,t}}\big[\Indc\{B_m\}\big] \Big]
  \ge \dfrac{T_m-T_{m-1}}{8\cdot 2^{\frac{M}{2}} M^2},
  \label{assumption:asst_on_m_c}
\end{align} 
then there eixsts a numerical constant $c>0$, independent of $(T,M,d, s_0)$, such that, 
\begin{align*}
  \sup_{\theta^\star:\|\theta^\star\|_2\le 1,~\|\theta^\star\|_0\le s_0} \Expect_{\theta^\star}\big[R_T(\mathbf{Alg})\big]
  \ge \frac{c}{M^4 2^{3M}}\sqrt{Ts_0}\left(\dfrac{T}{s_0}\right)^{\frac{1}{2(2^M-1)}}.
\end{align*}
\end{lemma}

Using the decomposition of the reget, we have for any $m\in[M]$,
\begin{align*}
& \sup_{\theta^\star:\|\theta^\star\|_2\le 1,~\|\theta^\star\|_0\le s_0}
\Expect_{\theta^\star}\big[R_T(\textbf{Alg})\big] \\
\ge & 2\Delta_m \sum^{T_m}_{t=T_{m-1}+1} \Expect_x\bigg[Z_m(d_{m,t}) \cdot 
\Big(\Expect_{P^t_{\theta,x}\circ Q^+_{m,t} }\big[\Indc \{a_t \in\cal A_m \}\big]
+ \Expect_{P^t_{\theta,x}\circ Q^-_{m,t}} 
\big[\Indc\{a_t\in \calA_m^c\}\big]\Big)\bigg]\\
\stackrel{(a)}{\ge} & 2\Delta_m\sum^{T_m}_{t=T_{m-1}+1} 
\Expect_x\bigg[Z_m(d_{m,t})\cdot\Big(1-\TV\big( P^t_{\theta,x}\circ Q^+_{m,t}, 
P^t_{\theta,x}\circ Q^-_{m,t} \big) \Big) \bigg]\\
\stackrel{(b)}{\ge} & 2\Delta_m \sum^{T_m}_{t = T_{m-1}+1} \Expect_x
\bigg[Z_m(d_{m,t})\cdot\Big(1-\TV \big(P^{T_{m}}_{\theta,x}\circ 
Q^+_{m,t},P^{T_{m}}_{\theta,x}\circ Q^-_{m,t} \big) \Big) \bigg],
\end{align*}
where step (a) is because $P(A)+Q(A^c) \ge 1-\TV(P,Q)$, 
and step (b) follows from the data processing inequality
of the total variation distance (Lemma~\ref{lemma:data_process}). 
For the total variation, 
\begin{align}
& 1 - \TV\big( P^{T_{m}}_{\theta,x}\circ Q^+_{m,t},  
  P^{T_{m}}_{\theta,x}\circ Q^-_{m,t}\big)
= \int\min\big(dP^{T_{m}}_{\theta,x}\circ Q^+_{m,t},
  dP^{T_{m}}_{\theta,x}\circ Q^-_{m,t}\big)\nonumber\\
\ge & \int_{B_m}  \min\left(dP^{T_{m}}_{\theta,x}\circ Q^+_{m,t},
  dP^{T_{m}}_{\theta,x}\circ Q^-_{m,t}\right)\nonumber\\
= & \frac{1}{2} \int_{B_m}  \Big(dP^{T_{m}}_{\theta,x}\circ Q^+_{m,t}
  + dP^{T_{m}}_{\theta,x}\circ Q^-_{m,t} 
  -\big|dP^{T_m}_{\theta,x}\circ Q^+_{m,t}
  -dP^{T_m}_{\theta,x}\circ Q^-_{m,t}\big|\Big) \nonumber\\
= & \frac{1}{2} \int_{B_m} \Big( dP^{T_{m-1}}_{\theta,x}\circ Q^+_{m,t}
  + dP^{T_{m-1}}_{\theta,x}\circ Q^-_{m,t} 
  -\big|dP^{T_{m-1}}_{\theta,x}\circ Q^+_{m,t}
  -dP^{T_{m-1}}_{\theta,x}\circ Q^-_{m,t}\big|\Big),
  \label{eq:boundtv}
\end{align}
where the last equality uses the fact that on the event
$B_m$, $dP^{T_{m-1}}_{\theta,x} = dP^{T_{m}}_{\theta,x}$.
Using the property that $\TV(P,Q) = \frac{1}{2}\int |dP-dQ|$ 
and $|P(A)-Q(A)|\le \TV(P,Q)$, we have
\begin{align*}
\eqref{eq:boundtv} = & \dfrac{1}{2}\bigg(\Expect_{P^{T_{m-1}}_{\theta,x}\circ Q^+_{m,t}}\big[\Indc \{B_m\} \big]
  +\Expect_{P^{T_{m-1}}_{\theta,x}\circ Q^-_{m,t}}\big[ \Indc \{A_m\} \big]\bigg)
  - \TV\Big(dP^{T_{m-1}}_{\theta,x}\circ Q^+_{m,t}, dP^{T_{m-1}}_{\theta,x}\circ Q^-_{m,t}\Big)\\
    \ge & \Expect_{P_{\theta,x}\circ Q^+_{m,t}}\big[\Indc \{B_m\}\big]
    -\dfrac{3}{2}\TV\Big( P^{T_{m-1}}_{\theta,x}\circ Q^+_{m,t},P^{T_{m-1}}_{\theta,x}\circ Q^-_{m,t} \Big).
\end{align*}
Applying Pinsker's inequality (Lemma \ref{lemma.TV_KL}),
we have 
\begin{align}
\TV\Big(P^{T_{m-1}}_{\theta,x}\circ Q^+_{m,t},
P^{T_{m-1}}_{\theta,x}\circ Q^-_{m,t}\Big) \le 
\sqrt{\frac{1}{2}D_{\rm KL}\Big(P^{T_{m-1}}_{\theta,x}\circ Q^+_{m,t}
~\big\|~P^{T_{m-1}}_{\theta,x}\circ Q^-_{m,t} \Big)}.
\label{eq:measurechange}
\end{align}

To simplify the right-hand side of Equation~\eqref{eq:measurechange}, 
we utilize the rotational invariance of the uniform distribution. 
First, let $v_1,v_2,\ldots,v_{2^{-M}\ts}$ be an orthonormal
basis of $\bR^{2^{-M}\ts}$, where $v_1 = u_{m,t}$; define two rotational
matrices $R_1 = [v_1,v_2,\cdots,v_{2^{-M}\ts}]$ and
$R_2 = [-v_1,v_2,\cdots,v_{2^{-M}\ts}]$; letting
$\theta'_m = \theta_m - 2(v_1^\top\theta_m)v_1 = R_1 R_2^\top \theta_m$,
we have $\theta'_m \stackrel{\mathrm{d}}{=} \theta_m$
and $\theta_m'^\top d_{m,t}  = -\theta_m^\top d_{m,t}$.
Further, let $\theta'$ denote the parameter induced by 
$\theta_1,\ldots,\theta_m',\ldots,\theta_M$---i.e.,
$\theta' := f(\theta_1,\ldots,\theta'_m,\ldots,\theta_M)$---we
then have $\theta' \stackrel{\rm d}{=} \theta$ and
\begin{align*}
  \eqref{eq:measurechange} = 
  & \sqrt{\frac{1}{2}D_{\rm KL}\big(P^{T_{m-1}}_{\theta,x}\circ Q^+_{m,t} ~\big\|~
  P^{T_{m-1}}_{\theta',x}\circ Q^+_{m,t}\big)}
  \le \sqrt{\frac{1}{2}\Expect_{Q^+_{m,t}}
  \Big[D_{\rm KL}\big(P^{T_{m-1}}_{\theta,x}~\big\|~
  P^{T_{m-1}}_{\theta',x} \big)\Big]},
\end{align*}
where the inequality is due to the joint convexity 
of the KL-divergence (Lemma~\ref{lemma:kl_conv}). 
The KL-divergence can then be explicitly computed:
\begin{align*}
  \Expect_{Q^+_{m,t}}\Big[D_{\rm KL}\big(P^{T_{m-1}}_{\theta,x} \big\| 
  P^{T_{m-1}}_{\theta',x}\big)\Big] = &
  \dfrac{1}{2}\sum^{T_{m-1}}_{\tau=1}\Expect_{Q^+_{m,t}}
  \Big[\big(f(\theta_1,\ldots,\theta_m,\ldots,\theta_M) 
  - f(\theta_1,\ldots,\theta_m',\ldots,\theta_M)\big)^\top x_{\tau,a_{\tau}}\Big]^2\\
  = &2 \Delta_m^2 \cdot \Expect_{Q^+_{m,t}}\big[|u_{m,t}^\top\theta_m|^2\big]
  \cdot u_{m,t}^\top\Big(\sum^{T_{m-1}}_{\tau=1} h_{\tau,a_{\tau}}h_{\tau,a_{\tau}}^\top\Big)u_{m,t}\\
  \le &2 \Delta_m^2 \cdot \Expect_{Q^+_{m,t}}\big[|u_{m,t}^\top\theta_m|^2\big]
  \cdot u_{m,t}^\top\Big(\sum^{T_{m-1}}_{\tau=1}\sum_{j\in[K]} 
  h_{\tau,j}h_{\tau,j}^\top\Big)u_{m,t},
\end{align*}
where $h_{\tau,j} = \sum_{\sigma\in\Pi(M):\sigma_m = 0} x_{\tau,j}(I_\sigma)
- \sum_{\sigma \in \Pi(M): \sigma_m = 1} x_{\tau,j} (I_{\sigma})$. Note that
$$\Expect_{Q^+_{m,t}}\big[|u_{m,t}^\top \theta_m|^2\big] =
\dfrac{\|d_{m,t}\|_2}{2Z_m(d_{m,t})}\cdot \Expect_{Q}\big[|u_{m,t}^\top\theta_m|^3\big] 
= \dfrac{\Expect_{Q}\big[|\theta_{m,1}|^3\big]}{2Z_m(u_{m,t})}
=  \dfrac{\Expect_{Q} \big[|\theta_{m,1}|^3\big]}{\Expect_{Q}\big[|\theta_{m,1}|\big]} 
\stackrel{\rm (a)}{=}  \dfrac{2}{2^{-M}\cdot \ts+1}
\le \frac{2^{M+1}}{\ts},
$$
where step (a) follows from Lemma~\ref{lemma:uniformmoment}.
We then can lower bound the regret as:
\begin{align}\label{eq:regret_final}
& \sup_{\theta^\star:\|\theta^\star\|_2\le 1,\|\theta^\star\|_0\le s_0}
\Expect_{\theta^\star}\big[R_T(\textbf{Alg})\big] \nonumber \nonumber\\ 
\ge & 2\Delta_m\sum^{T_m}_{t=T_{m-1}+1} \Expect_x\Bigg[ Z_m(d_{m,t})
  \cdot\bigg(\Expect_{P_{\theta,x}\circ Q^+_{m,t}}
  \big[\Indc \{B_m\}\big] 
  -\dfrac{3}{2}\sqrt{\dfrac{2^{M+1}\Delta_m^2}{\ts} 
  \cdot u_{m,t}^\top \Big(\sum^{T_{m-1}}_{\tau=1}\sum_{j\in[K]}
  h_{\tau,j}h_{\tau,j}^\top\Big)u_{m,t}}\bigg)\Bigg]\nonumber\\
\overset{\rm (a)}{\ge} & 2\Delta_m \sum^{T_m}_{t=T_{m-1}+1} \Expect_x\Bigg[ Z_m(d_{m,t})
  \cdot\bigg(\Expect_{P_{\theta,x}\circ Q^+_{m,t}}
  \big[\Indc \{B_m\}\big]-\dfrac{3}{2}\sqrt{\dfrac{2^{3M+1}\Delta_m^2T_{m-1}}{\ts}}\bigg)\Bigg]\nonumber\\
\stackrel{\rm (b)}{\ge} & 2\Delta_m\sum^{T_m}_{t=T_{m-1}+1} \Expect_x \bigg[ Z_m(d_{m,t})
  \cdot\Big(\Expect_{P_{\theta,x} \circ 
  Q^+_{m,t}}\big[\Indc\{B_m\}\big]- \dfrac{1}{2\cdot  2^{M+2}M^2} \Big)\bigg],
\end{align}
where step (a) uses the independence between $(x_{t,1},x_{t,2})$ and 
$\{(x_{\tau,1},x_{\tau,2})\}_{\tau\le T_{m-1}}$ and the concavity 
of $x\mapsto \sqrt{x}$; step (b) is due to the choice of $\Delta_m$
and $T_{m-1}$.
Note also that 
\begin{align*}
  \Expect_x \big[Z_m(d_{m,t})\big] = 
  \Expect_x \Big[\dfrac{\|d_{m,t}\|_2}{2}\Big]
  \cdot \Expect_{Q}\big[|u^\top_t \theta|\big] 
  = \Expect_x\Big[\dfrac{\|d_{m,t}\|_2}{2}\Big]
  \cdot\Expect_{Q}\big[|\theta_1|\big] 
  \overset{\rm (a)}{\le} 
  \dfrac{2^{\frac{M}{2}}}{\sqrt{\ts}}\Expect_x \big[\|d_{m,t}\|_2\big]
  \le 2^{\frac{M}{2}},
\end{align*}
where step (a) follows from Lemma \ref{lemma:uniformmoment}.
Consequently, 
\begin{align*}
  \eqref{eq:regret_final} \ge 2\Delta_m\bigg(\sum^{T_m}_{t = T_{m-1}+1}
    \Expect_x \Big[Z_m(d_{m,t})\cdot \Expect_{P_{\theta,x}\circ Q^+_{m,t}}
    \big[\Indc\{B_m\}\big]\Big]
    -\frac{1}{2\cdot 2^{\frac{M}{2}+2}M^2}.
  \bigg)
\end{align*}
Finally, letting $m$ be the batch that satisfies
Equation~\eqref{assumption:asst_on_m_c}, we have
\begin{align*}
  \sup_{\theta^\star:\|\theta^\star\|_2\le 1,~\|\theta^\star\|_0\le s_0} 
  \Expect_{\theta^\star}\cdot\big[R_T(\textbf{Alg})\big] \ge & 
  \dfrac{(T_m-T_{m-1}) \Delta_m}{2 \cdot 2^{\frac{M}{2}+2} M^2} 
  \ge \frac{c}{M^4 2^{7M/2}}\cdot\sqrt{s_0 T} \left( \dfrac{T}{s_0}\right)^{\frac{1}{2(2^M-1)}}.
\end{align*}

\subsection{A ``bad'' event happens with large enough probability}
\label{sec:bad_event_high_prob_modelc}
Our main result here is that a bad event occurs with sufficiently high probability that 
\eqref{assumption:asst_on_m_c} holds: 
\begin{lemma}
There exists some $m\in[M]$, such that: 
\begin{align*}
\sum^{T_m}_{t=T_{m-1}+1}\Expect_x \Big[ Z_m(d_{m,t}) 
\cdot \Expect_{P_{\theta,x}\circ Q^+_{m,t}} 
\big[\Indc \{B_m\} \big]\Big]
\ge \dfrac{T_m-T_{m-1}}{2^{\frac{M}{2}+2}M^2}.
\end{align*}
\end{lemma}
Since the union of $\{B_m\}_{m\in[M]}$ is the whole space, by a union bound, we have
$\sum^M_{m=1}P(B_m) \ge P(\cup_{m=1}^M B_m) = 1$, where $P$ is any probability measure.
Hence $P(B_m)\ge 1/M$ for at least one $m$. For any $m\in[M]$,
\begin{align*}
& \sum^{T_m}_{t=T_{m-1}+1}\Expect_x \Big[Z_m(d_{m,t}) \cdot 
\Expect_{P_{\theta,x}\circ Q^+_{m,t}}\big[\Indc\{B_m\}\big]\Big] 
= \sum^{T_m}_{t=T_{m-1}+1} \Expect_x\Expect_{Q}
\big[(d_{m,t}^\top\theta_m)_+ \cdot P_{\theta,x}(B_m)\big]
\end{align*}
Note that conditional on $\{x_{t,a}\}_{t \le T_{m-1}, a\in[K]}$, 
$\Indc \{B_m\}$ is independent of $\{x_{t,a}\}_{t>T_{m-1},a\in[K]}$.
Hence,
\begin{align*}
  P_{\theta,x}(B_m) = &\Prob_{\theta}\Big(t_{m-1} \le T_{m-1} <T_m \le t_m 
  \mid \{x_{1,a}\}_{a\in[K]},\ldots,\{x_{T,a}\}_{a\in [K]}\Big)\\
= &\Prob_{\theta}\Big(t_{m-1} \le T_{m-1} <T_m \le t_m
\mid \{x_{1,a}\}_{a\in[K]}, \ldots,\{x_{T_{m-1},a}\}_{a\in[K]}\Big)
\end{align*}
Consequently, using the independence between contextx across $t$,
we have
\begin{align*}
  & \Expect_x\Big[ \Expect_{Q}\big[(d_{m,t}^\top \theta_m)_+ P_{\theta,x}(B_m)\big]\Big]
  =  \Expect_{Q}\Big[ \Expect_x \big[(d_{m,t}^\top \theta_m)_+ P_{\theta,x}(B_m)\big]\Big]\\
  = & \Expect_{Q}\Big[ \Expect_x \big[(d_{m,t}^\top \theta_m)_+\big]\cdot  \Expect_x \big[P_{\theta,x}(B_m)\big]\Big]
  =  \Expect_{Q}\Big[ \Expect_x \big[(d_{M,T}^\top \theta_m)_+\big] \cdot \Expect_x \big[P_{\theta,x}(B_m)\big]\Big]\\
  = & \Expect_{Q}\Big[ \Expect_x \big[(d_{M,T}^\top \theta_m)_+ P_{\theta,x}(B_m)\big]\Big]
  =  \Expect_x\Big[\Expect_{Q}\big[(d_{M,T}^\top \theta_m)_+ P_{\theta,x}(B_m)\big]\Big]
\end{align*}
Using the above result, we obtain that,
\begin{align*}
  &\sum^{T_m}_{t=T_{m-1}+1}\Expect_x \Big[ Z_m(d_{m,t}) \cdot 
  \Expect_{P_{\theta,x}\circ Q^t_{1,m}} \big[ \Indc \{B_m\}\big] \Big]
  = \sum^{T_m}_{t=T_{m-1}+1}\Expect_x \Expect_{Q} 
  \big[(d_{M,T}^\top\theta_m)_+ \cdot P_{\theta,x}(B_m)\big] \\
  \ge & 
  \sum^{T_m}_{t=T_{m-1}+1}\Expect_x \Expect_{Q} \Big[\min_{m'\in[M]}
    \big\{(d_{M,T}^\top\theta_{m'})_+\big\}
  \cdot P_{\theta,x}(B_m)\Big]\\
  \stackrel{(a)}{=} &\sum^{T_m}_{t = T_{m-1}+1}\tZ \cdot \Expect_{P_{\theta,x}\circ \tQ}\big[\Indc\{B_m\}\big]
  = (T_m - T_{m-1}) \tZ \cdot \Expect_{P_{\theta,x}\circ \tQ}\big[\Indc\{B_m\}\big]
\end{align*}
where in step (a) we define the measure $\tQ$ via
$\frac{d\tQ}{dQ\times dP_x}(x,\theta) = \frac{\min_{m'\in[M]} (d_{M,T}^\top\theta_{m'})_+}{\tZ},$
and $\tZ = \Expect_x\Expect_{Q}\big[\min_{m'\in[M]} (d_{M,T}^\top\theta_{m'})_+\big]$ is a normalizing constant.
$$
\tZ = \Expect_x\Big[\|d_{M,T}\|_2 \cdot \Expect_{Q}\big[\min_{m\in[M]}~(u_{M,T}^\top\theta_m)_+ 
\big]\Big]= 
\Expect_x\Big[\|d_{M,T}\|_2 \cdot \Expect_{Q}\big[\min_{m\in[M]}~(\theta_{m,1})_+ 
\big]\Big],
$$
where the last equality is due the fact that $\theta_1,\ldots,\theta_M$
are independent of each other and the rotational invariance of the 
uniform distribution. Note also
\begin{align*}
\Expect_Q\big[\min_m (\theta_{m,1})_+\big]
= & \int^{\infty}_0 \Prob\big(\min_m (\theta_{m,1})_+ > t\big)\dd t
= \int^{\infty}_0 \Prob\big( (\theta_{1,1})_+ > t\big)^M \dd t
= \frac{1}{2^M}\int^{\infty}_0 \Prob\big( |\theta_{1,1}| > t\big)^M \dd t\\
= & \frac{1}{2^M}\int^{\infty}_0 \Prob\big( |\theta_{1,1}|^2 > t^2\big)^M \dd t
\ge \frac{1}{2^M} \int^{\textsf{B}(\frac{1}{2},\frac{\ts2^{-M}-1}{2})/2}_0
\Big(1 - \frac{2t}{\textsf{B}(\frac{1}{2},\frac{\ts2^{-M}-1}{2})}\Big)^M \dd t\\
= & \frac{\textsf{B}(\frac{1}{2},\frac{\ts 2^{-M}-1}{2})}{2^{M+1}(M+1)}
\ge \frac{1}{2^{\frac{M+1}{2}}(M+1)\sqrt{\ts}},
\end{align*}
where $\textsf{B}(\alpha,\beta)$ denotes the beta function with 
parameters $\alpha$ and $\beta$. With the above result, 
$$\tZ \ge \frac{1}{2^{\frac{M+1}{2}}\cdot(M+1)}$$
Since $\sum^M_{m=1}\Expect_{P_{\theta,x}\circ \tQ}\big[\Indc\{B_m\}\big]\ge 1$,
there exists $m\in[M]$, such that $\Expect_{P_{\theta,x}\circ \tQ}\big[\Indc \{B_m\}\big]\ge \frac{1}{M}$, and hence,
\begin{align*}
  \tZ\cdot  \Expect_{P_{\theta,x}\circ\tQ}\big[\Indc\{B_m\}\big] \ge \dfrac{\tZ}{M}
  \ge \dfrac{1}{2^{\frac{M}{2}+2}M^2}.
\end{align*}
Finally for this $m$, $\sum^{T_m}_{t=T_{m-1}+1}\Expect_x \Big[ Z_m(d_{m,t}) 
\cdot \Expect_{P_{\theta,x}\circ Q^+_{m,t}}\big[\Indc\{B_m\}\big]\Big]
\ge \dfrac{(T_m-T_{m-1}) }{2^{\frac{M}{2}+2}M^2}.$

\subsection{Lower Bound for Fully Online Learning Setting}\label{subsec:lower3}
So far we have establishesd the left-hand side of \eqref{eq:lowerbound} 
is greater or equal to the first term on the right-hand side when $M=O\big(\log{\log{(T/s_0)}}\big)$.
When $M=\Omega\big(\log{\log{(T/s_0)}}\big)$, the first term is dominated by the second term, so it suffices 
to show that the regret is lower bounded by the second term.
Lemma~\ref{lemma:online_modelc} completes the picture by showing the second part of the inequality.

\begin{lemma}
\label{lemma:online_modelc}
When $M = T$, there exists a two-arm setting with 
independent Guassian contexts, for which we have (for 
some numerical constant $c$ independent of $T,M,d, s_0$):
  \[
    \sup_{\theta^\star:\|\theta^\star\|_2\le 1, \|\theta^\star\|_0\le s_0}
    \Expect_{\theta^\star}\big[R_T(\mathbf{Alg})\big] \ge c\cdot \sqrt{Ts_0}.
  \]
\end{lemma}
The proof is a simple variant of the first part and the proof is in 
Appendix~\ref{sec:proofonline}. Note that our online regret lower bound recovers 
the lower bound obtained in \citet{chu2011contextual}---their lower bound is stated in the dense
and low-dimensional setting, but the adaptation is straightforward. 

\section{Achievable Guarantees: Regret Upper Bound}
\label{sec:upperbnd}

In this section we propose the {\em LASSO Batch Greedy Learning}
(LBGL) algorithm, similar in spirit to the (low-dimensional) 
Greedy Bandit algorithm \citep{bastani2017mostly}, to tackle 
the high-dimensional dynamic batch learning problem.  
This simple algorithm is minimax optimal (up to log factors).

\subsection{LASSO Batch Greedy Learning}
LBGL has two important features: 1) at each time $t$, 
it exploits the current estimate of the true parameter
$\theta^\star$ without further exploration; 2) it uses a static 
grid that is not adaptive (of course, a static grid
is a particular type of dynamic grid).
As it turns out, this already achieves the optimal regret bound.
Concretely, given a grid choice $\calT = \{t_1,\ldots,t_M\}$, 
at the beginning of batch $m$, the algorithm constructs a Lasso
estimate $\hat{\theta}_{m-1}$ 
of the true parameter using the data in the previous batches; 
then it selects the action $a\in[K]$ which maximizes the estimated 
reward $x_{t,a}^\top\hat{\theta}_{m-1}$ for any $t\in\{t_{m-1}+1,\ldots,t_m\}$; 
at the end of the $m$-th batch, the algorithm updates the estimate of 
the underlying parameters with the new observations in the current batch.
Finally, regarding the grid choice: inspired by the grid choice 
in \citet{han2020sequential}, we adopt a similar but somewhat 
different static grid for our setting:
\[
  t_1 = b\sqrt{s_0},\qquad t_m = \big\lfloor b\sqrt{t_{m-1}} \big\rfloor, \qquad m \in \{2,3,\ldots,M\},
\]
where $b = \Theta\big(\sqrt{T}\cdot ({T}/{s_0} )^{\frac{1}{2(2^M-1)}} \big)$ is chosen such that
$t_M=T$. The complete algorithm is described in Algorithm~\ref{algo.dbgl_c}.
We emphasize again this static grid choice--rather 
than a dynamic one--is not a limitation of our 
algorithm: as we discuss next, it is sufficient
to achieve the optimal regret bound (up to log 
factors) for the class of dynamic batch learning algorithms.

\begin{algorithm}[ht]
  \DontPrintSemicolon  
  \SetAlgoLined 
  \BlankLine
  \caption{LASSO Batch Greedy Learning (LBGL) under \modelc}
  \label{algo.dbgl_c}
  \textbf{Input} Time horizon $T$; context dimension $d$; number of batches $M$; sparsity bound $s_0$.\;
  \textbf{Initialize} $b = \Theta\Big( \sqrt{T} \cdot ({T}/{s_0})^{\frac{1}{2(2^M-1)}}\Big)$; $\hat{\theta}_0 = \mathbf{0}\in \mathbb{R}^d$;\;
  \textbf{Static grid} $\calT = \{t_1,\ldots,t_M\}$, with $t_1 = b\sqrt{s_0}$ and $t_m = b\sqrt{t_{m-1}}$ for $t \in\{2,\ldots,M\}$;\;
  \textbf{Partition} each batch into $M$ intervals evenly, i.e., $(t_{m-1},t_{m}] = \cup^M_{j=1}T_m^{(j)}$, for $m\in[M]$.\;
  \For{$m \gets 1$ \KwTo $M$}{
    \For{$t \gets t_{m-1}$ \KwTo $t_m$}{
      (a) Choose $a_t = \argmax{a\in[K]} x_{t,a}^\top \hat{\theta}_{m-1}$ (break ties with lower action index).\;
    (b) Incur reward $r_{t,a_t}$.
  }
  $T^{(m)}\leftarrow \cup^m_{m'=1}T^{(m)}_{m'}$;  $\lambda_m \leftarrow 10\sqrt{\dfrac{2\log{K}(\log{d} + 2\log{T})}{|T^{(m)}|}}$;\;
  Update $\hat{\theta}_{m} \leftarrow \argmin{\theta\in\mathbb{R}^d}~\dfrac{1}{2|T^{(m)}|} \sum_{t\in T^{(m)}}(r_{t,a_t} - x_{t,a_t}^\top\theta)^2 + \lambda_m \|\theta\|_1$.
}
\end{algorithm}

Theorem~\ref{thm:upperbnd_c} characterizes the performance of the LBGL algorithm.
In this section, we present the main steps in proving Theorem~\ref{thm:upperbnd_c}, leaving the
details to Appendix~\ref{appendix:proof_upperbound_modelc}.
\begin{theorem}
\label{thm:upperbnd_c}
Under \modelc, Assumptions~\ref{assumption:contexts}-\ref{assumption:actions}
and $M = O\big(\log{\log{(T/s_0)}}\big)$, we have
\begin{align}
\sup_{\theta^\star:\|\theta^\star\|_2\le 1, \|\theta^\star\|_0\le s_0} 
\Expect_{\theta^\star}\big[R_T(\mathbf{Alg})\big]
\le \frac{C \cdot M^{3/2} \sqrt{\log{K}\log{(KT)}\log{(dT)}} }{\gamma(K)\rho(K)} 
\cdot \sqrt{Ts_0}\left(\dfrac{T}{s_0}\right)^{\frac{1}{2(2^M-1)}},\label{eq:main_upperbnd_c}
\end{align}
where \textbf{Alg} is LBGL and $C>0$ is a numerical constant independent of $(T, d, M, K, s_0)$.
\end{theorem}

\begin{remark}
This regret upper bound matches the lower bound proved 
in Theorem~\ref{thm:dynamiclowerbnd} 
(up to logarithmic factors). That we only stated the theorem
for $M = O\big(\log{\log{(T/s_0)}}\big)$ is not a restriction, 
but instead a merit of our result:  with the number of batches
$M = O\big(\log{\log{(T/s_0)}}\big)$, we are already able 
to achieve the fully online optimal regret (up to log factors) 
$\tilde{O}(\sqrt{Ts_0})$: note that Lemma~\ref{lemma:online_modelc} 
has established the $\Omega(\sqrt{Ts_0})$ lower bound for 
fully online learning (under $K=2$) and hence a matching 
$\tilde{O}(\sqrt{Ts_0})$ regret bound indicates that it is minimax optimal.
Consequently, for any larger $M$, the achievable regret--which 
a priori will not get worse--cannot get better.

Note that the regret of any dynamic batch learning algorithm can be achieved by a fully online learning algorithm---in the online setting you can always divide the observations into batches and run the corresponding batch algorithm---and this observation immediately yields Corollary \ref{cor:onlineupperbnd}.
\end{remark}

\begin{corollary}
\label{cor:onlineupperbnd}
In the fully online learning setting ($M=T$) and under 
Assumptions~\ref{assumption:contexts}-\ref{assumption:actions}:
\begin{align}
\sup_{\theta^\star:\|\theta^\star\|_2\le 1, \|\theta^\star\|_0\le s_0}
\Expect_{\theta^\star}\big[R_T(\mathbf{Alg})\big]
\le \frac{C\sqrt{\big(\log{\log{(T/s_0)}}\big)^3\log{K}\log{(KT)}\log{(dT)}}
}{\gamma(K)\rho(K)} 
\cdot\sqrt{Ts_0},
\label{eq:online_upperbnd_c}
\end{align}
where $C>0$ is a numerical constant independent of $(T, d, M, K, s_0)$.
\end{corollary}

\subsection{Regret Analysis}
In this section, we present the main steps of proving 
Theorem~\ref{thm:upperbnd_c}. We start by showing that
the empirical covariance matrices are well-conditioned
even when the arms are adaptively chosen: in particular,
even though unlike in the low-dimensional settings the 
empirical covariance matrices are rank-deficient (as a 
result of high-dimensional features), the restricted 
eigenvalues are well-behaved. Then we leverage standard 
Lasso results to show that with ``well-behaved'' empirical 
covariance matrices, the Lasso estimates of $\theta^\star$
is reasonably close to the true parameters. Finally we 
translate the above results into the regret analysis, and
establish the desired regret upper bound.

\subsubsection{Establishing the restricted eigenvalue condition}
Given a sparsity parameter $s$ and a matrix $A$, we define the key quantity {\em restricted eigenvalues}:
\begin{align*}
  \phi_{\min}(s,A) \stackrel{\Delta}{=} \min_{v\in\mathbb{R}^d: \|v\|_0\le s}\bigg\{ \dfrac{v^\top A v}{\|v\|_2^2} \bigg\},
  \qquad  \phi_{\max}(s,A) \stackrel{\Delta}{=} \max_{v\in\mathbb{R}^d: \|v\|_0\le s}\bigg\{ \dfrac{v^\top A v}{\|v\|_2^2} \bigg\}.
\end{align*}
Following the notation in Algorithm~\ref{algo.dbgl_c}, $T_m^{(j)}$ denotes the $j$-th interval of the $m$-th batch (where the $m$-th batch has been divided evenly into $M$ intervals).
We then define for any $j,m\in[M]$ the empirical covariance matrix: $D_{m,j} = \sum_{t\in T^{(j)}_m} x_{t,a_t}x_{t,a_t}^\top$  and $A_m = \sum^m_{j=1} D_{j,m}$.
Lemma \ref{lemma:eigenvalue_condition} shows that the restricted eigenvalues are bounded from both above and below with high probabilities.
\revise{
\begin{lemma}
\label{lemma:eigenvalue_condition}
Suppose Assumptions \ref{assumption:contexts}-\ref{assumption:actions} hold. 
Given a sparsity parameter $s$, with probability at least 
$1-2M^2\exp (O(s\log d) -\Omega(\rho^2(K) \cdot \sqrt{Ts_0}/M))$, 
for any $j, m\in[M]$,
\begin{align*}
  \phi_{\max}\bigg(s,\dfrac{D_{m,j}}{|T^{(j)}_m|}\bigg) \le 16\log{K} ,
  \qquad \phi_{\min}\bigg(s,\dfrac{D_{m,j}}{|T^{(j)}_m|}  \bigg) \ge \dfrac{\gamma(K)\rho(K)}{4}.
\end{align*}
\end{lemma}
}
The detailed proof of Lemma~\ref{lemma:eigenvalue_condition} is 
deferred to Appendix~\ref{appendix:proof_eigenvalue}, 
and we provide the high-level steps here.
For a given $v\in\mathbb{R}^d$ such that $\|v\|_0\le s$ and $\|v\|_2=1$, 
we prove the upper bound of $v^\top D_{m,j} v$ using standard concentration inequalities.
We then generalize the upper bound to an $\varepsilon$-net of the set of all 
$s$-sparse $v$ by taking a union bound. Finally we extend the result to any 
$s$-sparse $v$ by utilizing the property of the $\varepsilon$-net. The proof of the lower bound is similar to that of the upper bound, except that we apply Assumption \ref{assumption:rdi} when proving the lower bound for a single vector.

\subsubsection{Bounding Lasso estimation error}
With well-behaved restricted eigenvalues, Lemma \ref{lemma:lasso_err} leverages standard Lasso results to prove an
estimation error bound for $\|\hat{\theta}_m-\theta^\star\|_2$.
\begin{lemma}
\label{lemma:lasso_err}
Under Assumptions \ref{assumption:contexts}-\ref{assumption:actions}, 
with probability at least $1-M\exp (\log{d}-\log{K}\cdot 
\Omega(\sqrt{Ts_0}/M)) - 2M^2 \cdot \exp 
(O(s_0 \cdot \frac{\log K \log d}{\gamma(K)\rho(K)})
-\Omega(\rho^2(K)\sqrt{Ts_0}))-M\cdot T^{-2}$,
for any $m\in [M]$,
\begin{align*}
  \revise{\|\hat{\theta}_m - \theta^\star\|_2\le \frac{800\sqrt{2}}{\gamma(K)\rho(K)} 
  \cdot \sqrt{s_0 M} \cdot \sqrt{\dfrac{\log{K}\cdot(2\log{T} + \log{d})}{t_m}}.}
\end{align*}
\end{lemma}
The proof utilizes classical Lasso theory \citep{bickel2009simultaneous} and is given in Appendix \ref{assumption:proof_of_lasso}.

\subsubsection{Analyzing regret upper bound}
With Lemmas \ref{lemma:eigenvalue_condition} and \ref{lemma:lasso_err}, we are now ready to bound the regret of  Algorithm \ref{algo.dbgl_c}.
Given $m\in[M]$, consider $t\in\{t_{m-1}+1,\ldots,t_m\}$, the instantaneous regret can be bounded as: 
$
\max_{a\in[K]}~\left(x_{t,a} - x_{t,a_t}\right)^\top\theta^\star \le  \max_{a\in[K]}~(x_{t,a}-x_{t,a_t})^\top(\theta^\star-\hat{\theta}_{m-1})\le 2\max_{a\in[K]}~|x_{t,a}^\top(\theta^\star - \hat{\theta}_{m-1})|,$
where the first inequality is from the definition of $a_t$. 

For a fixed $a\in[K]$, $x_{t,a}^\top(\theta^\star -\hat{\theta}_{m-1}) $ is $\|\theta^\star - \hat{\theta}_{m-1}\|^2_2$-sub-Gaussian. Thus, applying a sub-Gaussian maximal inequality, we get that given a $t\in[T]$, with probability at least $1-T^{-3}$, 
$$2\max_{a\in[K]}~\big|x_{t,a}^\top(\theta^\star - \hat{\theta}_{m-1})\big|\le 6\sqrt{\log{(TK)}}\cdot \|\theta^\star - \hat{\theta}_{m-1}\|_2.$$
Applying a union bound over the batch $m$ with $m \ge 2$ and invoking 
Lemma \ref{lemma:lasso_err}, we have with probability at least $1-(1+M)\cdot T^{-2}
-M\cdot\exp\big(\log{d}-\log{K}\cdot\Omega(\sqrt{Ts_0}/M)\big)-2M^2 \cdot \exp\Big(O\big(s_0\frac{\log K \log d}{\gamma(K)\rho(K)}\big)
-\Omega(\rho^2(K)\cdot\sqrt{Ts_0}/M)\Big)$, 
\begin{align*}
  \max_{a\in[K]}~(x_{t,a} -x_{t,a_t})^\top\theta^\star \le \frac{C}{\gamma(K)\rho(K)} 
  \cdot \sqrt{s_0M\log{(T K)}}
  \cdot \sqrt{\dfrac{\log{K}(2\log{T}+\log{d})}{t_{m-1}}},~\forall~t \in [t_{m-1}+1,t_m],
\end{align*}
where $C>0$ is a numerical constant.
Summing over the regret incurred in the $m\ge 2$ batches yields:
\begin{align*}
  \sum^M_{m=2}\sum^{t_m}_{t=t_{m-1}+1}\max_{a\in[K]}~(x_{t,a} - x_{t,a_t})^\top\theta^\star 
  \le&  \frac{C}{\gamma(K)\rho(K)} \cdot 
  bM^{3/2} \cdot \sqrt{s_0 \log K \log{(TK)} (\log d + 2\log{T})}\\
  \le & \dfrac{C'}{\gamma(K)\rho(K)} \cdot M^{3/2}
  \cdot \sqrt{\log{K}\log{(TK)} (\log{d}+2\log{T})}\sqrt{Ts_0}\left(\dfrac{T}{s_0}\right)^{\frac{1}{2(2^M-1)}},
\end{align*}
where $b = \Theta\big(\sqrt{T}\cdot (T/s_0)^{\frac{1}{2(2^M-1)}}\big)$ is from the choice of grids.
Finally for the first batch, since no rewards are observed, it suffices for us to adopt a crude bound:
\begin{align*}
  \sum^{t_1}_{t=1} \max_{a\in[K]}~(x_{t,a} - x_{t,a_t})^\top\theta^\star \le 2\sum^{t_1}_{t=1}\big(\max_{a\in[K]}~x_{t,a}^\top\theta^\star\big).
\end{align*}
Applying a sub-Gaussian maximal inequality and a union bound over all $t\in[t_1]$, we have with probability at least $1-T^{-2}$, 
\begin{align*}
  \sum^{t_1}_{t=1}\max_{a\in[K]}~(x_{t,a} - x_{t,a_t})^\top\theta^\star \le 6\sqrt{\log{(KT)}}\cdot t_1 
  = \Theta\bigg(\sqrt{\log{(KT)}}\sqrt{Ts_0}\cdot \Big(\dfrac{T}{s_0}\Big)^{\frac{1}{2(2^M-1)}} \bigg).
\end{align*}
Putting everything together, we then have that with probability at least $1-(2+M)\cdot T^{-2}-2M^2\cdot
\exp(O(s_0 \frac{\log K \log d}{\gamma(K)\rho(K)} )
-\Omega(\rho^2(K) \cdot \sqrt{Ts_0}/M))-M \cdot \exp(\log{d}-\log{K}\cdot\Omega(\sqrt{Ts_0}/M))$,
\begin{align*}
  R_T(\mathbf{Alg}) \le \frac{C''}{\gamma(K)\rho(K)} 
  \cdot M^{3/2} \cdot \sqrt{\log{K}\log{(KT)}\log{(dT)}}
  \cdot \sqrt{Ts_0} 
  \cdot \bigg( \dfrac{T}{s_0}\bigg)^{\frac{1}{2(2^M-1)}},\label{eq:high_prob_event}
\end{align*}
where $C''>0$ is a numerical constant resulting from merging the constant corresponding to the first batch and the constant $C^{\prime}$ (corresponding to all subsequent batches).
Since $M=O(\log{\log{(T/s_0)}})$, the above high-probability regret bound immediately implies the expected regret bound:
\begin{align*}
  \Expect_{\theta^\star}\big[ R_T(\textbf{Alg})\big] 
  \le  \frac{C'''}{\gamma(K)\rho(K)} \cdot
  M^{3/2} \cdot \sqrt{\log{K}\log{(KT)}\log{(dT)}}
  \cdot \sqrt{Ts_0} \cdot \bigg(\dfrac{T}{s_0}\bigg)^{\frac{1}{2(2^M-1)}},
\end{align*}
where $C'''>0$ is a numerical constant.

\section{Discussion}

Through matching lower and upper regret bounds, our work completes (up to certain log factors) the picture of dynamic batch learning in high-dimensional sparse linear contextual bandits. Further, the algorithm provided is very simple to implement in practice, an important merit from a practical standpoint. We close the paper by discussing possible extensions of our work.

\paragraph{Extension to sub-expoential reward distribution}
In this paper, we have focused on sub-Gaussian reward distribution.
It would be interesting to consider the high-dimensional 
dynamic batch learning problem with sub-exponential 
reward distribution (although this is a hard task even in
the fully onine setting).

\paragraph{Extension to sparsity-agnostic algorithm} 
It would be desirable to have a dynamic batch learning algorithm
that would not require any knowledge of a sparsity upper bound.
In the fully online decision making setting,~\cite{oh2021sparsity}
propose such an algorithm. Adapting it the the batched 
setting, however, is challengibg since the grid design
critically depends on $s_0$.

\paragraph{Other extensions} 
It would be interesting to explore the continuous action set case and understand whether learning guarantees in this regime are materially worse. 
Finally, going beyond to the non-parametric contextual bandits setting would also be useful.

\bibliographystyle{apalike}
\bibliography{di}

\appendix
\numberwithin{assumption}{section}

\section{Definitions and Auxiliary Results}\label{appendix.auxiliary}

We collect in this section all the known results in the existing literature that will be useful for us.

\begin{definition}
Let $(\mathcal{X}, \mathcal{F})$ be a measurable space and $P$, $Q$
be two probability measures on $(\mathcal{X}, \mathcal{F})$. 
\begin{enumerate}[label = (\alph*)]
	\item The total-variation distance between $P$ and $Q$ is defined as:
	$$ \mathsf{TV}(P,Q) = \sup_{A \in \mathcal{A}} |P(A) - Q(A)|.$$
	\item The KL-divergence between $P$ and $Q$ is:
	\begin{equation*}
	D_{\text{\rm KL}}(P\|Q) = \begin{cases}
	\int \log \frac{dP}{dQ} dP \text{\quad if $P << Q$} \\
	+\infty \text{\quad otherwise}
	\end{cases}
	\end{equation*}
\end{enumerate}
\end{definition}

\begin{lemma}[Paley-Zygmund inquality]
\label{lemma:pz}
If $X \ge 0$ is a random variable whose variance is finite,
Then for any $\theta \in (0,1)$,
\begin{align*}
& (1)~\PP\big(X > \theta \EE[X]\big)
\ge (1-\theta)^2 
\frac{\EE[Z]^2}{\EE[Z^2]};\\
& (2)~\PP\big(X > \theta \EE[X]\big) \ge 
\frac{(1-\theta)^2 \EE[X]^2}{\var(Z) + (1-\theta)^2 \EE[X]^2}.
\end{align*}
\end{lemma}

\begin{lemma}[Data-processing inequality \citep{Cover--Thomas2006}]
\label{lemma:data_process}
Let $X,Y,Z$ denote random variables drawn from a Markov chain in the order (denoted by $X\rightarrow Y\rightarrow Z$) that the conditional distribution of
Z depends only on Y and is conditionally independent of X. 
Then if $X\rightarrow Y \rightarrow Z$, we have $I(X;Y)\ge I(X;Z)$, where $I(X;Y)$ is the mutual information between $X$ and Y.
\end{lemma}

\begin{lemma}[Pinsker's inequality]\label{lemma.TV_KL}
Let $P$ and $Q$ be any two probability measures on the same measurable space. Then
$\mathsf{TV}(P,Q)\le \sqrt{\frac{1}{2} \cdot D_{\rm KL}(P\|Q)}$.
\end{lemma}

\begin{lemma}[Joint convexity of the KL-divergence \citep{Cover--Thomas2006}]\label{lem:kl_conv}
\label{lemma:kl_conv}
$D_{\rm KL}(P\| Q)$ is jointly convex in its arguments $P$ and $Q$: let $P_1,P_2,Q_1,Q_2$ be distributions on $\calX$, then for any $\lambda\in[0,1]$,
  \[
    D_{\rm KL}(\lambda P_1 + (1-\lambda)P_2 \| \lambda Q_1 + (1-\lambda)Q_2) 
    \le \lambda D_{\rm KL}(P_1 \| Q_1) + (1-\lambda)D_{\rm KL}(P_2\| Q_2).
  \]
\end{lemma}
\begin{lemma}[Sub-Gaussian maximal inequality~\citep{rigollet201518}]
Let $X_1,\ldots,X_K$ be $K$ centered  $\sigma^2$-sub-Gaussian random variables, then for any $t>0$,
$\Prob\left( \max_{k\in[K]} X_k\ge t \right)\le K e^{-\frac{t^2}{2\sigma^2}}.$
\end{lemma}

\section{Proof of Lemma~\ref{lemma:diversity_condition}}
\label{appx:proof_diversity}
\begin{enumerate}
\item The proof follows directly from~\citet[Lemma 4]{han2020sequential}.
\item Given a unit vector $v \in \mathbb{R}^d$, for any $\delta > 0$,
\begin{align*}
  &\Prob\Big((v^{\top} x_{t,a_t})^2 \le 
  \frac{\alpha}{e} \cdot (2cK)^{-\frac{1}{\alpha}}\Big)
  =    \Prob\Big(-(v^{\top} x_{t,a_t})^2 \ge 
  -\frac{\alpha}{e} \cdot (2cK)^{-\frac{1}{\alpha}}\Big)\\ 
  = & \Prob\Big(\exp\big(-(v^\top x_{t,a_t})^2 \cdot \delta \big) \ge 
  \exp\big(-\frac{\alpha}{e} \cdot (2cK)^{-\frac{1}{\alpha}} \cdot \delta\big)\Big) 
  \stackrel{\textnormal{(a)}}{\le}  \exp\Big(\frac{\alpha}{e} \cdot \delta 
  \cdot (2cK)^{-\frac{1}{\alpha}}\Big) \cdot
  \Expect\Big[\exp\big(-(v^\top x_{t,a_t})^2\cdot\delta \big)\Big]\\
  \le & \exp\Big(\frac{\alpha}{e} \cdot \delta 
  \cdot (2cK)^{-\frac{1}{\alpha}} \Big) \cdot
  \sum_{a\in [K]} \Expect\Big[\exp\big(-(v^\top x_{t,a})^2 \cdot\delta \big)\Big]
  \stackrel{\textnormal{(b)}}{\le} cK \cdot \exp\big(\frac{\alpha}{e} \cdot \delta
  \cdot (2cK)^{-\frac{1}{\alpha}}\big) \cdot \delta^{-\alpha},
\end{align*}
where step (a) is due to Markov's inequality and step (b) follows
from Equation~\eqref{eq:mgf_condition}. Taking $\delta = e \cdot (2cK)^{1/\alpha}$ 
(the minimizer of the upper bound), we arrive at 
$\Prob \big( (v^\top x_{t,a_t})^2 \ge \frac{\alpha}{e}\cdot (2cK)^{-1/\alpha} \big) \ge \frac{1}{2}$.

\item Let $v \in \mathbb{R}^d$ be an arbitrary unit vector. For each $a\in [K]$, 
\begin{align*}
    \Prob\Big((v^\top x_{t,a})^2 \ge \frac{1}{2} \cdot \Expect\big[(v^\top x_{t,a})^2\big] \Big) 
    \stackrel{\textnormal{(a)}}{\ge} & \frac{\frac{1}{4}\cdot \Expect \big[(v^\top x_{t,a})^2 \big]^2}
    {\textnormal{Var}\big((v^\top x_{t,a})^2\big) 
    + \frac{1}{4} \cdot  \Expect\big[ (v^\top x_{t,a})^2 \big]^2}\\ 
    \stackrel{\textnormal{(b)}}{\ge} &\frac{\frac{1}{4} \cdot 
      \Expect\big[(v^\top x_{t,a})^2 \big]^2}{\frac{\Lambda^2}{8K} + 
    +\frac{1}{4} \cdot  \Expect\big[(v^\top x_{t,a})^2\big]^2}
    \stackrel{\textnormal{(c)}}{\ge}  \frac{2K}{2K + 1}.
  \end{align*}
  Above, the step (a) is due to the Paley-Zygmund inequality (Lemma~\ref{lemma:pz});
  step (b) and (c) follow from the assumption.
  As a consequence, we have $\Prob\big((v^\top x_{t,a})^2 < \frac{\Lambda}{2} \big)
  \le \frac{1}{2K+1}$.
  Finally,
  \begin{align*}
    \Prob\Big((v^\top x_{t,a_t})^2 \ge \frac{\Lambda}{2}\Big)
    \ge & \Prob\Big(\min_{a\in[K]} (v^\top x_{t,a})^2 \ge \frac{\Lambda}{2} \Big)
    = 1 - \Prob\Big(\min_{a\in[K]} (v^\top x_{t,a})^2 < \frac{\Lambda}{2} \Big)\\
    \ge & 1- \sum_{a \in [K]}\Prob\Big((v^\top x_{t,a})^2 < \frac{\Lambda}{2} \Big)
    \ge 1- \frac{K}{2K+1} \ge \frac{1}{2}, 
  \end{align*}
  completing the proof.

  \item Without loss of generality, we assume $\nu\le 1$.
    For an arbitrary unit vector $v \in \mathbb{R}^d$, 
    \begin{align*}
      \Prob\Big((v^\top x_{t,a_t})^2 \ge \frac{\Lambda}{2} \Big) 
    = \sum_{a = 1}^2 \Prob\Big(a_t = a, (v^\top x_{t,a})^2 
      \ge \frac{\Lambda}{2} \Big). 
    \end{align*}
    By symmetry, we only need to focus on $a = 1$, for which we have 
    \begin{align*}
      &\Prob\Big(a_t = 1, (v^\top x_{t,1})^2 \ge \frac{\Lambda}{2} \Big)
      = \int \Indc \Big\{x_{t,a}^\top \theta \ge x_{t,2}^\top \theta, 
      (v^\top x_{t,1})^2 \ge \frac{\Lambda}{2}\Big\}
      \cdot p(x_{t,1},x_{t,2})dx_{t,1}dx_{t,2}\\
      \stackrel{\text{(a)}}{\ge} & \frac{1}{2} \cdot \Prob \Big(a_t = 1, (v^\top x_{t,a}) \ge\frac{\Lambda}{2} \Big) 
      + \frac{\nu}{2}\int \Indc\Big\{x_{t,1}^\top \theta \ge x_{t,2}^\top \theta, 
      (v^\top x_{t,1})^2 \ge \frac{\Lambda}{2}\Big\} \cdot p(-x_{t,1},-x_{t,2}) dx_{t,1}dx_{t,2}\\ 
      = & \frac{1}{2} \cdot \Prob \Big(a_t = 1, (v^\top x_{t,1})^2 \ge \frac{\Lambda}{2} \Big)
      + \frac{\nu}{2} \cdot \Prob \Big(a_t = 2, (v^\top x_{t,1})^2 \ge \frac{\Lambda}{2} \Big)\\
      \ge & \frac{\nu}{2} \cdot \Prob \Big((v^\top x_{t,1})^2 \ge \frac{\Lambda}{2} \Big)
      \ge \frac{\nu}{2} \cdot \Prob\Big( (v^\top x_{t,a})^2 \ge \frac{\Expect\big[(v^\top x_{t,1})^2 \big]}{2}\Big)
      \stackrel{\textnormal{(b)}}{\ge}  \frac{\nu}{8} \cdot \frac{\Expect\big[(v^\top x_{t,a})^2\big]^2}
      {\Expect\big[(v^\top x_{t,a})^4\big]}
      \stackrel{\textnormal{(c)}}{\ge} \frac{\nu \Lambda^2}{128},
    \end{align*}
    where step (a) is by the assumption; step (a) is due to the Paley-Zygmund inequality; 
    step (c) is because of the assumption and $v^\top x_{t,a}$ is 1-sub-Gaussian. Combining 
    the case of $a = 1$ and $a=2$, we have
    \begin{align*}
      \Prob\Big((v^\top x_{t,a_t})^2 \ge \frac{\Lambda}{2} \Big) 
      \ge \frac{\nu \Lambda^2}{64}.
    \end{align*}

  \newcommand{\tmid}{\textnormal{mid}}
\item Without loss of generality, we assume $\nu_1,\nu_2(K)\le 1$. 
    Fix an arbitrary unit vector $v$, and $\theta \in \RR^d$. 
    To start, we focus on $a=1$. We decompose all the permutations of $[K]$
    into three subsets: $\calI_{\min}$, $\calI_{\max}$ and $\calI_{\tmid}$, where 
    $\calI_{\min}\,:=\,\{\pi: \pi_1 = 1\}$, $\calI_{\max}\,:=\,\{\pi:\pi_K=1\}$ and 
    $\calI_{\tmid} \,\:=\, \{\pi: \pi_1 \neq 1, \pi_K \neq 1\}$. We then have
    \begin{align*}
      \Prob\Big((v^\top x_{t,1})^2 \ge \frac{\Lambda}{2} \Big)
      = & \sum_{\pi:\pi \in \calI_{\min}} \Prob\Big((v^\top x_{t,1})^2 
      \ge \frac{\Lambda}{2}, x_{t,\pi_1}^\top \theta \le \ldots 
    \le x_{t,\pi_K}^\top \theta \Big)\\
      + & \sum_{\pi:\pi \in \calI_{\max}} \Prob\Big((v^\top x_{t,1})^2 
      \ge \frac{\Lambda}{2}, x_{t,\pi_1}^\top \theta \le \ldots 
    \le x_{t,\pi_K}^\top \theta \Big)\\
      + & \sum_{\pi:\pi \in \calI_{\tmid}} \Prob\Big((v^\top x_{t,1})^2 
      \ge \frac{\Lambda}{2}, x_{t,\pi_1}^\top \theta \le \ldots 
    \le x_{t,\pi_K}^\top \theta \Big)
    \end{align*}
    By the assumption, for any permutation $\pi$,
    \begin{align*}
    &\nu_2(K) \cdot \Prob\Big((v^\top x_{t,1})^2 
    \ge \frac{\Lambda}{2}, x_{t,\pi_1}^\top \theta \le \ldots 
    \le x_{t,\pi_K}^\top \theta \Big)\\ 
    & \le 
    \Prob\Big((v^\top x_{t,\pi_1})^2 
    \ge \frac{\Lambda}{2}, x_{t,\pi_1}^\top \theta \le \ldots 
    \le x_{t,\pi_K}^\top \theta \Big)  + \Prob\Big((v^\top x_{t,\pi_K})^2 
    \ge \frac{\Lambda}{2}, x_{t,\pi_1}^\top \theta \le \ldots 
    \le x_{t,\pi_K}^\top \theta \Big).
    \end{align*}
    As a consequence, 
    \begin{align*}
    \Prob\Big((v^\top x_{t,1})^2 \ge \frac{\Lambda}{2} \Big)
    \le \frac{1}{\nu_2(K)}\cdot\sum_{\pi}
    & \Prob\Big((v^\top x_{t,\pi_1})^2 
    \ge \frac{\Lambda}{2}, x_{t,\pi_1}^\top \theta \le \ldots 
    \le x_{t,\pi_K}^\top \theta \Big) + \\ 
    &\qquad \Prob\Big((v^\top x_{t,\pi_K})^2 
    \ge \frac{\Lambda}{2}, x_{t,\pi_1}^\top \theta \le \ldots 
    \le x_{t,\pi_K}^\top \theta \Big). 
    \end{align*}
    Above, by the relaxed symmetry condition, 
    \begin{align*}
      &\Prob \Big((v^\top x_{t,\pi_1})^2 \ge \frac{\Lambda}{2}, x_{t,\pi_1}^\top \theta
      \le \ldots \le x_{t,\pi_K}^\top \theta \Big) \\
      \le & \frac{1}{\nu_1} \cdot \int \Indc\Big\{(v^\top x_{t,\pi_1})^2 \ge \frac{\Lambda}{2}, x_{t,\pi_1}^\top \theta
      \le \ldots \le x_{t,\pi_K}^\top \theta \Big\} \cdot p(-x_{t,1},\ldots,-x_{t,K}) 
      dx_{t,1}\ldots dx_{t,K}\\
        = & \frac{1}{\nu_1} \cdot \Prob\Big((v^\top x_{t,\pi_1})^2 \ge \frac{\Lambda}{2}, x_{t,\pi_1}^\top \theta
        \ge \ldots \ge x_{t,\pi_K}^\top \theta \Big).
    \end{align*}
    We then have 
    \begin{align*}
    &\Prob\Big((v^\top x_{t,1})^2 \ge \frac{\Lambda}{2}\Big) \\
      \le &
    \frac{1}{\nu_1 \nu_2(K)} \cdot\sum_{\pi}
    \bigg(\Prob\Big((v^\top x_{t,a_t})^2 
    \ge \frac{\Lambda}{2}, x_{t,\pi_1}^\top \theta \ge \ldots 
    \ge x_{t,\pi_K}^\top \theta \Big)  + \Prob\Big((v^\top x_{t,a_t})^2 
    \ge \frac{\Lambda}{2}, x_{t,\pi_1}^\top \theta \le \ldots 
    \le x_{t,\pi_K}^\top \theta \Big) \bigg)\\
    = & \frac{2}{\nu_1\nu2(K)}\cdot 
    \Prob\Big((v^\top x_{t,a_t})^2 \ge \frac{\Lambda}{2}\Big).
    \end{align*}
    Finally, we arrive at 
    \begin{align*}
      \Prob\Big((v^\top x_{t,a_t})^2 \ge \frac{\Lambda}{2} \Big) 
      \ge \frac{\nu_1 \nu_2(K)}{2} \cdot \Prob\Big((v^\top x_{t,1})^2 \ge \frac{\Lambda}{2}\Big)
      \ge & \frac{\nu_1\nu_2(K)}{2} \cdot \Prob\Big((v^\top x_{t,1})^2 \ge
      \frac{v^\top \Expect[x_{t,1}x_{t,1}^\top]v}{2}\Big)\\
        \stackrel{\textnormal{(a)}}{\ge} & \frac{\nu_1\nu_2(K)}{8} \frac{\Lambda^2}{16} = \frac{\nu_1\nu_2(K)\Lambda^2}{128},
    \end{align*}
    where step (a) is due to the Paley-Zygmund (Lemma~\ref{lemma:pz}) 
    inequality and the assumption.
\end{enumerate}


\section{Proof of Main Lemmas in Section~\ref{sec:lowerbnd}}
\label{appendix:proof_lower_bnd}
\subsection{Proof of Lemma \ref{lemma:online_modelc}}
\label{sec:proofonline}
As in the batched case (and with the same notation), 
we construct a prior $Q$ for $\theta$, where 
$\theta(S)\sim \Unif(\Delta \mathbb{S}^{s_0-1})$ 
and $\theta^\star(S^c)=0$, and $\Delta = \sqrt{{s_0}/{32T}}$.
Then:
\begin{align}
\label{eq:online_regret}
\sup_{\theta^\star:\|\theta^\star\|_2\le 1,\|\theta^\star\|_0\le s_0} 
\Expect_{\theta^\star}\big[R_T(\textbf{Alg})\big]
\ge &\Expect_Q\Expect_\theta \big[R_T(\textbf{Alg})\big]
= \sum^T_{t=1}\Expect_{Q} 
\Expect_x\Expect_{P^t_{\theta,x}}\Big[
\max_{a\in \{1,2\}}(x_{t,a}^\top\theta-x_{\tau,a_\tau}^\top\theta)
\Big]\nonumber\\
= & \sum^T_{t=1} \EE_x\EE_Q \EE_{P_{\theta,x}^t}
\big[\Indc\{a_t = 1\} \cdot  (d_t^\top \theta)_+ 
+ \Indc\{a_t=2\} \cdot (d_t^\top\theta)_-\big],
\end{align}
where $d_t = x_{t,2} - x_{t,1}$. Define two new measures via:
$\frac{dQ^+_t}{dQ}(\theta) = \frac{(d_t^\top\theta)_+}{Z(d_t)}$
and $\frac{dQ^-_t}{dQ} (\theta)= \frac{(d_t^\top\theta)_-}{Z(d_t)}$;
$Z(d_t) = \EE_Q[(d_t^\top\theta)_+] = \EE_Q[(d_t^\top \theta)_-]$ 
is a common normalizing constant. Using this representation, we have
\begin{align}
\label{eq:online_regret_2}
\eqref{eq:online_regret} = 
&\sum^T_{t=1}\Expect_x\bigg[ Z(d_t)\cdot \Big( \Expect_{P_{\theta,x}\circ Q^+_t}\big[\Indc\{a_t=1\}] + 
\Expect_{P_{\theta,x}\circ Q^-_t}\big[\Indc\{a_t=2\}] \Big)\bigg]\nonumber\\
\stackrel{(a)}{\ge}& \sum^T_{t=1}\Expect_x
\Big[Z(d_t) \cdot \big(1-\TV(P^{t-1}_{\theta,x}\circ Q^+_t,
P_{\theta,x}^{t-1}\circ Q^-_t ) \big) \Big]\nonumber\\
\stackrel{(b)}{\ge} & \sum^T_{t=1}\Expect_x\Bigg[Z(d_t) \cdot \bigg(
1-\sqrt{\frac{1}{2} D_{\rm KL}\big(P^{t-1}_{\theta,x}\circ Q^+_t~\|~
P^{t-1}_{\theta-2(\tu_t^\top\theta)\tu_t,x}\circ Q^+_t \big)}
\bigg)\bigg]\nonumber\\
\stackrel{(c)}{\ge} & \sum^T_{t=1}\Expect_x\Bigg[ Z(d_t) \cdot \bigg(1-\sqrt{ \frac{1}{2}\Expect_{Q^+_t} 
\Big[ D_{\rm KL}\big(P^{t-1}_{\theta,x}~\|~ P^{t-1}_{\theta-2(\tu_t^\top\theta)\tu_t,x} \big)\Big]}\bigg) \Bigg] 
\end{align}
where step (a) follows from $P(A)+Q(A^c)\le 1-\TV(P,Q)$; step (b) is due to 
a change of measure and Lemma~\ref{lemma.TV_KL}; step (c) is because 
of the joint convexity of the KL-divergence. Above,
\begin{align*}
	D_{\rm KL}\big(P^{t-1}_{\theta,x}~\|~P^{t-1}_{\theta-2(\tu_t^\top\theta)\tu_t,x} \big) 
  = \dfrac{1}{2}\sum^{t-1}_{\tau=1}\big(2(\tu_t^\top\theta)\cdot(\tu_t^\top x_{\tau,a_{\tau}})\big)^2
  = 2(\tu_t^\top\theta)^2 \cdot \tu_t^\top \Big(\sum^{t-1}_{\tau=1}x_{\tau,a_{\tau}}x_{\tau,a_{\tau}}^\top \Big)\tu_t,
\end{align*}
where $\tu_t \in \mathbb{R}^d$ satisfies $\tu_t(S) = \frac{d_t(S)}{\|d_t(S)\|_2}$
and $\tu_t(S^c) = 0$.
Plugging in the expression of the KL-divergence, we have
\begin{align*}
\eqref{eq:online_regret_2} 
\ge &\sum^T_{t=1}\Expect_x\Bigg[ Z(d_t)\cdot \bigg(
1-\sqrt{  \Expect_{Q^t_1}\big[(\tu_t^\top\theta)^2\big] \cdot
 \tu_t^\top\Big(\sum^{t-1}_{\tau=1} x_{\tau,a_\tau}^\top x_{\tau,a_{\tau}}\Big)\tu_t }
\bigg)\Bigg]\\
\ge &\sum^T_{t=1}\Expect_x\Bigg[ Z(d_t)\cdot \bigg(
1-\sqrt{  \Expect_{Q^t_1}\big[(\tu_t^\top\theta)^2\big]\cdot 
 \tu_t^\top\Big(\sum^{t-1}_{\tau=1} x_{\tau,1}x_{\tau,1}^\top+x_{\tau,2}x_{\tau,2}^\top  \Big)\tu_t }
\bigg)\Bigg]\\
\stackrel{(a)}{\ge} & \sum^T_{t=1}\Expect_x\bigg[ Z(d_t)\cdot\Big(
1-\sqrt{\dfrac{4t\Delta^2}{s_0}}
\Big)
\bigg]
\stackrel{(b)}{\ge}  \dfrac{1}{2}\sum^T_{t=1}\Expect_x\big[Z(d_t)\big]
\ge  \dfrac{T \Delta}{10}
= \dfrac{\sqrt{T s_0}}{40\sqrt{2}},
\end{align*}
where step (a) is by taking expectation w.r.t. 
$\big\{(x_{\tau,1},x_{\tau,2})\big\}_{\tau\le t-1}$; 
step (b) follows from the choice of $\Delta$. The proof is completed.

\section{Proof of Main Lemmas in Section \ref{sec:upperbnd}}
\label{appendix:proof_upperbound_modelc}

\subsection{Proof of Lemma~\ref{lemma:eigenvalue_condition}}
\label{appendix:proof_eigenvalue}
Consider the $m$-th batch, for any $j\in[M]$,
by definition $a_t  = \argmax{a\in[K]} x_{t,a}^\top \hat{\theta}_{m-1}$
for any $t\in T^{(j)}_m$, where $\hat{\theta}_{m-1}$ depends {\em only}
on the observations from batch $1$ to $m-1$. Hence $\{x_{t,a_t}\}_{t \in T^{(j)}_m}$
are mutually independent and follow the same distribution conditional on the previous batches.
Consider now a fixed sparsity upper bound $s$.

\paragraph{Upper bound} 
Given a vector $v\in\mathbb{R}^d$, such that $\|v\|_0\le s$ and $\|v\|_2=1$. 
Let $\supp(v)$ denote the support of $v$, where without loss of generality we 
assume $|\supp(v)|=s$ (otherwise we can include extra zero coordinates in 
$\supp(v)$), and let $\calN(\varepsilon)$ denote the  $\varepsilon$-net 
of $\mathbb{S}^{s-1}$.
For notational simplicity, denote $Y_{t,a} = (v^\top x_{t,a})^2$. 
For any $\delta,\mu>0$, one has 
\begin{align*}
& \Prob\bigg(\sum_{t\in T^{(j)}_m}Y_{t,a_t}  \ge (4+\delta)\cdot |T^{(j)}_m| \,\Big|\, \hat{\theta}_{m-1} \bigg) 
\stackrel{(a)}{\le}  \exp\Big(-\mu (4+\delta) \cdot |T^{(j)}_m| \Big) \cdot
\Expect\bigg[\exp \Big(\mu \cdot \sum_{t\in T^{(j)}_m}Y_{t,a_t}\Big) \,\Big|\, \hat{\theta}_{m-1}\bigg]\nonumber \\
%
%
\stackrel{(b)}{=} & \exp \Big(-\mu (4+\delta) \cdot |T^{(j)}_m| \Big)\cdot
\prod_{t\in T^{(j)}_m}\Expect\big[\exp (\mu \cdot Y_{t,a_t}) \,|\, \hat{\theta}_{m-1}\big] 
%
\le  \exp\Big(- \mu (4+\delta) \cdot |T^{(j)}_m| \Big)
\prod_{t\in T^{(j)}_m} \Big( \sum_{a\in[K]}\Expect\big[ \exp(\mu Y_{t,a}) \big]\Big)  
\end{align*}
where step (a) follows from the Markov's inequality and step (b) is due to the (conditional) independence
across $t$. Since $x_{t,a}$ is 1-sub-Gaussian, $v^\top x_{x,a}$ is as well 1-sub-Gaussian.
As a result, $Y_{t,a} - \Expect[Y_{t,a}]$ is sub-exponential with parameter $(4\sqrt{2},4)$, and $\Expect[Y_{t,a}]\le 4$.
With this, we obtain a Bernstein-type bound, 
\begin{align*}
\Prob\bigg(\sum_{t\in T^{(j)}_m}Y_{t,a_t}  \ge (4+\delta)\cdot |T^{(j)}_m| \bigg) 
\le & \exp{\bigg(\Big(-\min \Big(\frac{\delta^2}{64},\frac{\delta}{8}\Big) +\log{K}
\Big)\cdot|T^{(j)}_m| \bigg)}.
\end{align*}
Combining everything above and letting $\delta = 9\log{K}$ we arrive at,
\begin{align*}
\Prob\bigg(\dfrac{1}{|T^{(j)}_m|} \sum_{t\in T^{(j)}_m}(v^\top x_{t,a_t})^2
\ge  4+9\log{K} \bigg) \le \exp\Big(-\dfrac{\log{K}}{8}\cdot |T^{(j)}_m| \Big).
\end{align*}
Taking a union bound, 
we get that with probability at least $1-\exp(s\log{d}+s\log{(1+1/\varepsilon)}-|T_m^{(j)}| \log{K}/8)$, 
for any $v$ such that $\|v\|_0\le s$, $\|v\|_2=1$ and $v(\supp(v))\in \calN(\varepsilon)$, 
\begin{align*}
\dfrac{1}{|T_m^{(j)}|} \sum_{t \in  T^{(j)}_{m}}(v^\top x_{t,a_t})^2 
\le 4+9\log{K} \le 15\log{K}.
\end{align*}
Let $u\in\mathbb{R}^d$ be an {\em arbitrary} vector such that $\|u\|_0\le s$ and $\|u\|_2=1$.
By the definition of the $\varepsilon$-net, there exists $v_0\in\calN(\varepsilon)$, such that $\|u(\supp(u)) - v_0\|_2\le \varepsilon$. Let $v\in\mathbb{R}^d$ be a vector such that $v(\supp(u)) = v_0$ and $v(\supp(u)^c) = 0$. 
By construction $\|u-v\|_2\le \varepsilon.$
Consequently,
\begin{align*}
\dfrac{u^\top D_{m,j} u}{|T_m^{(j)}|}  - \dfrac{v^\top D_{m,j} v}{|T_m^{(j)}|} 
=  \dfrac{u^\top D_{m,j} (u-v)}{|T^{(j)}_m|}+\dfrac{(u-v)^\top D_{m,j} v}{|T^{(j)}_m|}
\le   2\varepsilon \phi_{\max}\bigg(s,\dfrac{D_{m,j}}{|T^{(j)}_m|} \bigg). 
\end{align*}
Note that $|T^{(j)}_m| = \Omega(\sqrt{Ts_0}/M)$, for any $j,m\in[M]$. 
Taking the supreme over $u$ and rearranging yields that 
with probability at least $1-\exp(s\log{d}+s\log{(1+1/\varepsilon)}-\Omega(\sqrt{Ts_0}/M))$, 
\begin{align}
  \phi_{\max}\bigg(s,\frac{D_m}{|T^{(j)}_m|} \bigg)\le 
  \dfrac{15\log{K}}{1-2\varepsilon}.\label{eq:lambdamax}
\end{align}

\paragraph{Lower bound}
We now proceed to prove a lower bound for the restricted eigenvalues.
By Assumption~\ref{assumption:rdi},
$\Prob(Y_{t,a_t} \ge \gamma(K) \mid \hat{\theta}_{m-1}) \ge \rho(K)$.
We then have that 
\begin{align*}
  &\Prob\bigg(\frac{1}{|T_m^{(j)}|} \sum_{t \in T^{(j)}_m}Y_{t,a_t} \le \frac{\gamma(K)\rho(K)}{2} \,\big|\, \hat{\theta}_{m-1}\bigg)
  \le \Prob\bigg(\frac{1}{|\tmj|}\sum_{t\in\tmj}\Indc\big\{Y_{t,a_t} \ge \gamma(K) \big\} \le \frac{\rho(K)}{2} \,\big| \, \hat{\theta}_{m-1}\bigg)\\
  \le &\Prob\bigg(\frac{1}{|\tmj|}\sum_{t\in\tmj}\Indc\big\{Y_{t,a_t} \ge \gamma(K) \big\}
  - \Prob\big(Y_{t,a_t} \ge \gamma(K) \mid \hat{\theta}_{m-1}\big) \le -\frac{\rho(K)}{2}\,\big| \, \hat{\theta}_{m-1}\bigg)
  \le \exp\Big(-\frac{\rho^2(K)}{2} \cdot |\tmj|\Big),
\end{align*}
where the last inequality is due to the Chernoff bound.

Taking a union bound over all $s$-sparse unit vector $v$ 
whose support is in $\calN(\varepsilon)$, 
we conclude that with probability at least 
$1-\exp{\left(s\log{d}+s\log{(1+1/\varepsilon)}- \rho^2(K)|T^{(j)}_m|/2  \right)}$, for any $v$ whose support is in $\calN(\varepsilon)$,  
\begin{align}
  \dfrac{1}{|T^{(j)}_m|} \sum_{t \in T^{(j)}_m} (v^\top x_{t,a_t})^2 >\dfrac{\gamma(K)\rho(K)}{2}.\label{eq:event}
\end{align}
We now condition on the events~\eqref{eq:lambdamax} and~\eqref{eq:event},
and turn our attention to an arbitrary vector $u\in\mathbb{R}^d$ such that $\|u\|_0\le s$ and $\|u\|_2=1$. 
By the definition of $\varepsilon$-nets, there exists $v_0\in\mathcal{N}(\varepsilon)$
such that $\|u(\supp(u))-v_0\|_2\le \varepsilon$. Let $v\in \mathbb{R}^d$ be the vector 
such that $v(\supp(u)) = v_0$ and $v(\supp(u)^c) = 0$. 
Then 
\begin{align*}
  \dfrac{1}{|T^{(j)}_{m}|}\sum_{t\in T^{(j)}_m}(u^\top x_{t,a_t})^2  
  \ge & \dfrac{1}{|T^{(j)}_m|}\bigg(\sum_{t\in T^{(j)}_{m}}(v^\top x_{t,a_t})^2 +2(u-v)x_{t,a_t}x_{t,a_t}^\top v \bigg) \\
  \ge &\dfrac{\gamma(K)\rho(K)}{2} - 2\varepsilon\phi_{\max}\bigg(s,\dfrac{D_{m,j}}{|T^{(j)}_m|}\bigg) 
  \ge \dfrac{\gamma(K) \rho(K)}{2} - \dfrac{30\varepsilon \log{K}}{1-2\varepsilon}
\end{align*}
Finally letting $\varepsilon = \min(\frac{1}{32},\frac{\gamma(K)\rho(K)}{128 \log{K} })$
and taking a union bound over $j,m\in[M]$, we conclude that with probability at least
$1-2M^2\exp{(s\log{d} + s\log(1+\frac{128\log{K}}{\gamma(K) \rho(K)}) - \Omega(\rho^2(K)\cdot \sqrt{Ts_0}/M ))}$, for any $j,m\in[M]$,
\begin{align}
  \phi_{\min}\bigg(s,\dfrac{D_m}{|T^{(j)}_m|} \bigg) \ge 
  \dfrac{\gamma(K) \rho(K)}{4},
  \qquad \phi_{\max} \bigg(s,\dfrac{D_m}{|T^{(j)}_m|}\bigg)\le 16\log{K}.
  \label{eq:eigevent}
\end{align}

\subsection{Proof of Lemma~\ref{lemma:lasso_err}}
\label{assumption:proof_of_lasso}
To start, we work on an upper bound on the magnitude of $x_{t,a_t,l}$ (the $l$-th coodinate of $x_{t,a_t}$). 
Given any $m\in[M]$ and $l\in[d]$, define $M_{m,l} = \sqrt{\frac{1}{T^{(m)}}\sum_{t\in T^{(m)}}x_{t,a_t,l}^2}$. For any $\delta >0$ and $0<\mu<1/4$,
\begin{align}
\Prob(M_{m,l}^2\ge 16 \log{K}+\delta)\stackrel{(a)}{\le} & 
\Prob\bigg(\sum_{t\in T^{(m)}}\Big(\max_{a\in[K]} x^2_{t,a,l}-\Expect[\max_{a\in [K]}x^2_{t,a,l}] 
\Big)\ge |T^{(m)}| \cdot \delta \bigg)\nonumber \\
\stackrel{(b)}{\le}& \Expect\left[\exp{\left(\mu\sum_{t\in T^{(m)}}\left(\max_{a\in [K]} x^2_{t,a,l}-\Expect[\max_{a\in [K]}x^2_{t,a,l}] \right)\right) } \right]\exp\left(-\mu\delta |T^{(m)}|\right)\nonumber \\
  \stackrel{(c)}{\le} & \prod_{t\in T^{(m)}}\left\{\sum_{a\in[K]}\Expect\left[\exp{(\mu (x_{t,a,l}^2-\Expect[x^2_{t,a,l}] ))} \right] \right\}\exp\left(-\mu\delta |T^{(m)}|\right)\nonumber \\
\stackrel{(d)}{\le} &
\exp \big((\log{K}+16\mu^2-\mu\delta ) \cdot |T^{(m)}|   \big),\label{eq:bernsteinbnd}
\end{align}
where step (a) is because $\Expect[\max_{a\in[K]} x^2_{t,a,l}] \le 16\log{K}$; step (b) follows from the Markov's inequality; step (c) is due to the independence of $x_{t,a,l}$ across $t$ and step (d) is because $x^2_{t,a,l}-\Expect[x_{t,a,l}^2]$ is $(4\sqrt{2},4)$-sub-exponential.
Optimizing the right-hand side of \eqref{eq:bernsteinbnd} over $0<\mu\le 1/4$ and taking a union bound over $l\in[d]$, we obtain that 
\begin{align*}
  \Prob\Big(\max_{l\in[d]} M_{m,l}^2 \ge 16\log{K}+ \delta \Big)
  \le d\exp\bigg(\Big(\log{K}-\min\Big(\dfrac{\delta^2}{64},\dfrac{\delta}{8}\Big)\Big)\cdot |T^{(m)}| \bigg).
\end{align*}
Taking $\delta = 9\log{K}$, one has that with probability at least $1-\exp{(\log{d}-\log{K}\cdot|T^{(m)}|/8)}$, for all $l\in[d]$.
\begin{align}
  M^2_{m,l}\le 25\log{K}.\label{eq:magnitudeevent}
\end{align}
For any $m\in [M]$, any $s \le d$ and any $v\in \mathbb{R}^d$ such that $\|v\|_0\le s$ and $\|v\|_2 = 1$, 
\begin{align*}
  \dfrac{1}{|T^{(m)}|}v^\top A_m v = \sum^m_{j=1} \dfrac{|T_j^{(m)}|}{|T^{(m)}|}\left(\dfrac{v^\top D_{j,m}v}{|T_j^{(m)}| } \right),
\end{align*}
and consequently,
\begin{align*}
  \phi_{\max}\left( s, \dfrac{A_m}{|T^{(m)}|}\right)  \le \max_{j\in[m]}
  ~ \phi_{\max}\left(s, \dfrac{D_{j,m}}{T^{(m)}_j}\right), 
  \qquad \phi_{\min}\left(s,\dfrac{A_m}{|T^{(m)}|}\right)\ge \min_{j\in [m]} 
  ~\phi_{\min}\left(s,\dfrac{D_{j,m}}{|T^{(m)}_j|}\right).
\end{align*}
By Lemma \ref{lemma:eigenvalue_condition}, with probability
at least $1-2M^2 \cdot \exp(O(s\log(\frac{d\log K}{\gamma(K)\rho(K)}))-\Omega(\rho^2(K) \cdot \sqrt{Ts_0}/M)) $, 
for any $j, m\in[M]$,
\begin{align}
  \phi_{\max}\left(s,\dfrac{A_m}{|T^{(m)}|}\right)\le 16\log{K},\qquad 
  \phi_{\min}\left(s,\dfrac{A_m}{|T^{(m)}|} \right)\ge \dfrac{\gamma(K)\rho(K)}{4}. \label{eq:nice_event}
\end{align}
By the definition of $\hat{\theta}_m$,
\begin{align*}
  \dfrac{1}{2|T^{(m)}|}\sum_{t\in T^{(m)}}(r_{t,a_t} - x_{t,a_t}^\top \hat{\theta}_m)^2 + \lambda_m\|\hat{\theta}_m\|_1\le &\dfrac{1}{2|T^{(m)}|}\sum_{t\in T^{(m)}}(r_{t,a_t} - x_{t,a_t}^\top \theta^\star)^2 + \lambda_m\|\theta^\star\|_1.
\end{align*}
Rearranging yields
\begin{align*}
  \dfrac{1}{2|T^{(m)}|}\sum_{t\in T^{(m)}}(x_{t,a_t}^\top \theta^\star -x_{t,a_t}^\top\hat{\theta}_m)^2 +\lambda_m\|\hat{\theta}_m\|_1\le \lambda_m \|\theta^\star\|_1 + \dfrac{1}{|T^{m}|}\sum_{t\in T^{(m)}}(x_{t,a_t}^\top\hat{\theta}_m - x_{t,a_t}^\top\theta^\star)\varepsilon_t.
\end{align*}
By the construction of $T^{(m)}$, $\{\varepsilon_t\}_{t\in T^{(m)}}$ 
are mutually independent conditional on the selected contexts, 
we obtain that with probability at least  $1-T^{-2} - \exp (\log d - \log K \cdot |T^{(m)}|/2)$,
\begin{align*}
  \dfrac{1}{|T^{(m)}|}\sum_{t\in T^{(m)}}(x_{t,a_t}^\top \hat{\theta}_m - x_{t,a_t}^\top\theta^\star)\varepsilon_t \le \sum^d_{l=1} M_{n,l}\sqrt{\dfrac{2(\log{d}+2\log{T})}{|T^{(m)}|}}|\hat{\theta}_{m,l} -\theta^\star_l| \le \dfrac{\lambda_m}{2} \|\hat{\theta}_m  - \theta^\star\|_1.
\end{align*}
With the above two inequalities together, we obtain that
\begin{align}
  \dfrac{1}{2|T^{(m)}|}\sum_{t\in T^{(m)}}(x_{t,a_t}^\top \theta^\star -x_{t,a_t}^\top\hat{\theta}_m)^2 +\dfrac{\lambda_m}{2}\|\hat{\theta}_m-\theta^\star\|_1\le \lambda_m\left(  \|\theta^\star\|_1-\|\hat{\theta}_m\|_1 + \|\hat{\theta}_m-\theta^\star\|_1\right).\label{eq:basicineq}
\end{align}
Define $S_0 = \supp(\theta^\star)$. An immediate result of \eqref{eq:basicineq} is that 
\begin{align*}
  &\dfrac{1}{2}\|\hat{\theta}_m-\theta^\star\|_1\le \|\theta^\star(S_0)\|_1-\|\hat{\theta}_m(S_0)\|_1+\|\hat{\theta}_m(S_0)-\theta^\star(S_0)\|_1\\
  \Rightarrow & \|\hat{\theta}_m(S_0^c) - \theta^\star(S_0^c)\|_1\le 3\|\hat{\theta}_m(S_0)-\theta^\star(S_0)\|_1
\end{align*}
Before proving the final result, we state the following lemma from~\citet{bickel2009simultaneous} 
that links the restricted eigenvalues to the condition for recovering sparse signals, 
where we slightly modify the notation in our presentation.
\begin{lemma}[\citet{bickel2009simultaneous}]
\label{lemma:reconditions}
Fix a matrix $A$.  Assume that there exists an integer $r$, such that $r\ge s_0$ and $s_0+r\le d$, such that
  \begin{align*}
    \kappa \stackrel{\Delta}{=} \sqrt{\phi_{\min}(s_0+r,A)}\left(1-3\sqrt{\dfrac{s_0\phi_{\max}(r,A)}{r\phi_{\min}(s_0+r,A) }}\right)>0.
  \end{align*}
Then
\begin{align}
  &\min\left\{\dfrac{v^\top A v}{\|v(S)\|_2^2}: S\subset[d],|S|\le s_0 ,v\neq0,\|v(S^c)\|_1\le 3\|v(S)\|_1 \right\}>0,\label{eq:re1}\\
  &\min\left\{\dfrac{v^\top A v}{\|v(\widetilde{S})\|_2^2}: S\subset[d],|S|\le s_0 ,v\neq0,\|v(S^c)\|_1\le 3\|v(S)\|_1 \right\} = \kappa^2>0,\label{eq:re2}
\end{align}
where $\widetilde{S}$ is the union of $S$ and the set of $r$ largest in absolute value coordinates of $v$ outside $S$.
\end{lemma}
Now take $r = \frac{1152 s_0\log{K}}{\gamma(K) \rho(K)}$. By construction $r \ge s_0$ and
Assumption~\ref{assumption:actions} ensures $s_0 + r \le d$. 
Using Lemma~\ref{lemma:eigenvalue_condition} with $s = s_0 + r$, 
we have with probability at least $1 - 2M^2 \cdot \exp(O(s_0\frac{\log K \log d}{\gamma(K)\rho(K)}
- \Omega(\rho^2(K) \cdot \sqrt{Ts_0}/M))$,
\begin{align*}
  \phi_{\max}\bigg(s_0 + r, \frac{A_m}{|T^{(m)}|} \bigg) \le 16 \log (K), \qquad
  \phi_{\min}\bigg(s_0 + r, \frac{A_m}{|T^{(m)}|} \bigg) \ge \frac{\gamma(K)\rho(K)}{4}.
\end{align*}
On the above event, we have
\begin{align*}
\dfrac{9s_0\phi_{\max}(r,A_m/|T^{(m)}|)}{r\phi_{\min}(s_0+r,A_m/|T^{(m)}|)} 
\le \dfrac{576  s_0\log{K}}{r\gamma(K)\rho(K)} = \dfrac{1}{2},
\end{align*}
and 
\begin{align*}
  \kappa = \sqrt{\phi_{\min}\left(s_0+r,\dfrac{A_m}{|T^{(m)}|} \right)}
  \left(1-3\sqrt{\dfrac{s_0\phi_{\max}(r,A_m/|T^{(m)}|) }{ r\phi_{\min}(s_0+r, A_m/|T^{(m)}|)}}\right)\ge 
  \frac{\sqrt{\gamma(K)\rho(K)}}{2} \cdot \Big(1-\dfrac{\sqrt{2}}{2} \Big)>0.
\end{align*}
By Lemma \ref{lemma:reconditions}, both \eqref{eq:re1} and \eqref{eq:re2} hold, and consequently
\begin{align}
  \dfrac{1}{|T^{(m)}|}\sum_{t\in T^{(m)}}(x_{t,a_t}^\top \theta^\star -x_{t,a_t}^\top \hat{\theta}_m)^2\ge 
  \kappa^2 \|\theta^\star(\widetilde{S}_0)-\hat{\theta}_m(\widetilde{S}_0)\|_2^2.\label{eq:inter1}
\end{align}
Additionally by \eqref{eq:basicineq},
\begin{align}
  \dfrac{1}{2|T^{(m)}|}\sum_{t\in T^{(m)}}(x_{t,a_t}^\top \theta^\star -x_{t,a_t}^\top\hat{\theta}_m)^2 
  {\le} &2\lambda_m \|\hat{\theta}_m(S_0)-\theta^\star(S_0)\|_1 
  \stackrel{(a)}{\le} 2\lambda_m \sqrt{s_0}\|\hat{\theta}_m(S_0)-\theta^\star(S_0)\|_2\nonumber \\
  \le & 2\lambda_m \sqrt{s_0}\|\hat{\theta}_m(\widetilde{S}_0)-\theta^\star(\widetilde{S}_0)\|_2
  \label{eq:inter2}
\end{align}
where step (a) is due to the Cauchy-Schwarz inequality. Combining \eqref{eq:inter1} and \eqref{eq:inter2} yields
\begin{align*}
  \|\theta^\star(\widetilde{S}_0) -\hat{\theta}_m(\widetilde{S}_0)\|_2\le \dfrac{4\lambda_m \sqrt{s_0}}{\kappa^2}.
\end{align*}
Observe that the $k$th largest coordinates of $|\theta^\star(S_0^c)-\hat{\theta}_m(S_0^c)|$ is bounded by $\|\theta^\star(S_0^c)-\hat{\theta}_m(S_0^c)\|_1/k$, and consequently,
\begin{align*}
  \|\theta^\star(\widetilde{S}_0^c)-\hat{\theta}_m(\widetilde{S}_0^c)\|_2^2 \le& \|\theta^\star(S^c_0)-\hat{\theta}_m(S^c_0)\|_1^2\sum^d_{k=r+1}\dfrac{1}{k^2}\le \dfrac{1}{r}\|\theta^\star(S^c_0)-\hat{\theta}_m(S^c_0)\|_1^2\\
  \le & \dfrac{9}{r}\|\theta^\star(S_0) - \hat{\theta}_m(S_0)\|_1^2 \le \dfrac{9s_0}{r}\|\theta^\star(\widetilde{S}_0) - \hat{\theta}_m(\widetilde{S}_0)\|_2^2,
\end{align*}
where the last inequality follows from the Cauchy-Schwarz inequality.
A result of the above inequality is that,
\begin{align*}
  \|\theta^\star - \hat{\theta}_m\|_2
  \stackrel{(a)}{\le}& \left(1+3\sqrt{\dfrac{s_0}{r}}\right)\|\theta^\star(\widetilde{S}_0) - \hat{\theta}_m(\widetilde{S}_0)\|_2\le \left(1+3\sqrt{\dfrac{s_0}{r}}\right)\dfrac{4\lambda_m \sqrt{s_0}}{\kappa^2}.
\end{align*}
Finally taking a union bound, we conclude that with probability 
at least $1-MT^{-2} - M\exp{(\log{d} - \log{K}\cdot 
\Omega(\sqrt{Ts_0}/M))}-2M^2\exp{(O(s_0\frac{\log K \log d}{\gamma(K)\rho(K)}-\Omega(\rho^2(K) \cdot \sqrt{Ts_0}/M))}$,
for any $m\in[M]$,
\begin{align*}
  \|\hat{\theta}_m-\theta^\star\|_2\le \frac{800\sqrt{2}}{\gamma(K) \rho(K)}\cdot
  \sqrt{s_0 M} \cdot \sqrt{\dfrac{\log{K}(\log{d}+2\log{T})}{t_m}}.
\end{align*}

\section{Auxiliary Lemmas}
\begin{lemma}\label{lemma:uniformmoment}
Suppose that $\theta \sim \Unif(\mathbb{S}^{s_0-1})$, 
then the moment of $|\theta_1|$ can be computed:
\begin{align*}
  \Expect |\theta_1|^p = \begin{cases}
    \dfrac{2 \Gamma(\frac{s_0}{2}+1)}{\sqrt{\pi}s_0 \Gamma(\frac{s_0+1}{2})} & p=1,\\
    \dfrac{1}{s_0} & p=2,\\
    \dfrac{4 \Gamma(\frac{s_0}{2}+1)}{\sqrt{\pi}s_0(s_0+1) \Gamma(\frac{s_0+1}{2})} & p=3,\\
    \dfrac{3}{s_0(s_0+2)} & p=4.
  \end{cases}
\end{align*}
Moreover, we have that
 $\dfrac{2}{5\sqrt{s_0}} \le \Expect |\theta_1| \le \dfrac{2}{\sqrt{s_0}}.$
\end{lemma}
\proof{Proof of Lemma \ref{lemma:uniformmoment}}
The density of $\theta$ is 
$
f(\theta) = f(\theta_2,\ldots,\theta_{s_0}) = 
\left(\dfrac{s_0\pi^{s_0/2} }{\Gamma(\frac{s_0}{2}+1)} \right)^{-1}
\dfrac{2}{\sqrt{1 - \theta_2^2-\ldots - \theta_{s_0}^2}}\cdot
\Indc\left({\sum^{s_0}_{l=2}\theta_l^2\le 1}\right),
$
where $\Gamma(x) = \int^{\infty}_0 s^{x-1}e^{-s}ds$ is the Gamma function. To compute the integrals, we leverage the spherical coordinates
\begin{align*}
  \begin{cases}
    \theta_2 = r\cos{\phi_1},\\
    \theta_3 = r\sin{\phi_1}\cos{\phi_2},\\
    \vdots\\
    \theta_{s_0-1} = r\sin{\phi_1}\sin{\phi_2}\cdots\sin{\phi_{s_0-3}}\cos{\phi_{s_0-2}},\\
    \theta_{s_0} = r\sin{\phi_1}\sin{\phi_2}\cdots\sin{\phi_{s_0-3}}\sin{\phi_{s_0-2}}.
  \end{cases}
\end{align*}
Then by direct calculation,
\begin{align*}
  \Expect |\theta_1| = & \left(\dfrac{s_0\pi^{s_0/2} }{\Gamma(\frac{s_0}{2}+1)} \right)^{-1}\int_{\sum^{s_0}_{l=2}\theta_l^2\le 1} 
  2 d\theta_2\ldots d\theta_{s_0}\\
  = &2 \left(\dfrac{s_0\pi^{s_0/2} }{\Gamma(\frac{s_0}{2}+1)} \right)^{-1}\int^{1}_0 \int^{\pi}_0\cdots \int^{2\pi}_0 r^{s_0-2}\sin^{s_0-3}{\phi_1}\sin^{s_0-4}{\phi_2}\cdots\sin{\phi_{s_0-3}}dr d\phi_1 d\phi_2\cdots d\phi_{s_0-2}\\
  =&2\left(\dfrac{s_0\pi^{s_0/2}}{\Gamma(\frac{s_0}{2}+1)} \right)^{-1} \cdot\dfrac{1}{s_0-1} \cdot\dfrac{\Gamma(\frac{s_0-2}{2})\Gamma(\frac{1}{2})}{\Gamma(\frac{s_0-1}{2})}\cdot \dfrac{\Gamma(\frac{s_0-3}{2})\Gamma(\frac{1}{2})}{\Gamma(\frac{s_0-2}{2})}\cdots\dfrac{\Gamma(1)\Gamma(\frac{1}{2})}{\Gamma(\frac{3}{2})} \cdot  2\pi\\
  =& \dfrac{2}{\sqrt{\pi}s_0}\dfrac{\Gamma(\frac{s_0+2}{2})}{\Gamma(\frac{s_0+1}{2})}
  =\begin{cases}
    \frac{1}{2^{s_0-1}}{s_0-1\choose{(s_0-1)/2}}, & \text{if $s_0$ is odd},\\
    \frac{2^{s_0+1}}{\pi s_0} {s_0 \choose s_0/2 }^{-1},& \text{if $s_0$ is even}.
  \end{cases}
\end{align*}
From the Sterling's formula, we arrive at
$
\dfrac{2}{5\sqrt{s_0}}\le \Expect|\theta_1|
\le \dfrac{2}{\sqrt{s_0}}.
$
Similarly, we can compute the higher moments of $|\theta_1|$. As for the second moment,
\begin{align*}
  \Expect|\theta_1|^2 
  = 4\left(\dfrac{s_0\pi^{s_0/2}}{\Gamma(\frac{s_0}{2}+1)} \right)^{-1}\cdot \dfrac{\Gamma(\frac{s_0-1}{2}) \Gamma(\frac{3}{2})}{2\Gamma(\frac{s_0}{2}+1)} \cdot \dfrac{\pi^{s_0-1/2}}{\Gamma(\frac{s_0-1}{2})}
  =  \dfrac{1}{s_0}.
\end{align*}
And for the third moment,
\begin{align*}
  \Expect|\theta_1|^3  
  = 4\left(\dfrac{s_0\pi^{s_0/2}}{\Gamma(\frac{s_0}{2}+1)} \right)^{-1}\cdot 
\dfrac{2}{s_0^2-1}\cdot \dfrac{\pi^{s_0-1/2}}{\Gamma(\frac{s_0-1}{2})}
  = \dfrac{4}{\sqrt{\pi}s_0(s_0+1)}\dfrac{\Gamma(\frac{s_0+2}{2})}{\Gamma(\frac{s_0+1}{2})}.
\end{align*}
For the fourth moment,
\begin{align*}
  \Expect|\theta_1|^4  
  = 4\left(\dfrac{s_0\pi^{s_0/2} }{\Gamma(\frac{s_0}{2}+1)} \right)^{-1}\cdot 
  \dfrac{3\sqrt{\pi} \Gamma(\frac{s_0-1}{2}) }{4(s_0+2)\Gamma(\frac{s_0}{2}+1)} \cdot \dfrac{\pi^{s_0-1/2}}{\Gamma(\frac{s_0-1}{2})}
  =  \dfrac{3}{s_0(s_0+2)}.
\end{align*}
\endproof

\section{Parallel results under \modelp}
\label{sec:alternative_proof}
In this section, we collect parallel results under \modelp. 
In what follows, Section~\ref{appendix:formulation_modelp} 
formulates the problem, stating the conditions and assumptions.
Section~\ref{appendix:lower_bound_modelp} establishes
the regret lower bound of
$\Omega\big(c \cdot \max\{ M^{-2}2^{-M} \cdot \sqrt{Ts_0} (T/s_0)^{\frac{1}{2(2^M-1)}},
\sqrt{Ts_0}\}\big)$. Section~\ref{appendix:upper_bound_modelp}
presents a matching upper bound of the regret under
the set of (generic) assumptions stated in 
Section~\ref{appendix:formulation_modelp}. In particular, 
we verify the assumptions  in the case of two arms, while 
that of $K>2$ needs more delicate analysis---we leave
that for future work.

\subsection{Problem Fomulation}
\label{appendix:formulation_modelp}
Recall that undep \modelp, we have a set of $K$ parameters
$\{\theta_1^\star,\ldots,\theta_K^\star\}$; when action 
$a\in[K]$ is chosen, a reward $r_{t,a} = x_t^\top \theta^\star_a +\xi_t$
is incurred, where $\{\xi_t\}_{t=0}^{\infty}$ is a 
sequence of  \textbf{iid} zero-mean 1-sub-Gaussian random variables.
We assume $\|\theta_a^\star\|_2 \le 1$ for 
all $a\in[K]$ and the contexts $x_t$ are \textbf{iid} drawn.
The following Assumptions~\ref{assumption:contexts_p}-\ref{assumption:actions_p}
are parallel to Assumptions~\ref{assumption:contexts}-\ref{assumption:actions}
under \modelc.

\begin{assumption}[Sub-Guassianity]\label{assumption:contexts_p}
The marginal distribution of $x_t$ is $1$-sub-Gaussian.
\end{assumption}

\begin{assumption}[Diverse covariate]\label{assumption:rdi_p}
There are (possibly $K$-dependent) positive
constants $\gamma(K)$ and $\rho(K)$, such that
for any $\{\theta_a\}_{ a\in[K]}$,
any unit vector  $v \in \mathbb{R}^d$ 
and any $a\in[K]$, there is 
$\PP\big(v^\top x_tx_t^\top v \cdot 
\Indc\{a^* = a\} \ge \gamma(K)\big) \ge \rho(K)$,
where 
$a^* = \argmax{a\in[K]}~x_t^\top \theta_a$.
\end{assumption}

\begin{assumption}[Sparsity in High-Dimension]\label{assumption:cov_p}
The linear contextual bandits have high-dimensional contexts
$d = \mathbf{Poly}(T)$ 
and sparse parameters: there exists some $\varepsilon > 0$ such that
$\|\theta_a^\star\|_0\le s_0 = O(T^{1-\varepsilon})$ 
for all $a\in[K]$.
\end{assumption}

\begin{assumption}[Not Many Actions]\label{assumption:actions_p}
\revise{The number of actions $K$ satisfies 
$\frac{\log K}{\gamma(K)\rho(K)} = O( d / s_0)$
and $\frac{\log K}{\gamma(K) \rho^3(K)} = O(\sqrt{T^{1-\varepsilon}/s_0})$.
}
\end{assumption}

The following lemma establishes the sufficient
condition for Assumption~\ref{assumption:rdi_p}
for $K=2$.

\begin{lemma}
\label{lemma:diversity_condition_modelp}
 When $K=2$,  suppose both of the 
  following conditions hold
\begin{enumerate}
\item there exists a constant $\Lambda > 0$
such that $\lambda_{\min}\big(\EE[x_tx_t^\top]\big)
\ge \Lambda$;
for any unit vector $v \in \RR^d$ there is 
$v^\top \EE[x_tx_t^\top ]v \ge \Gamma$;
\item there exists a constant $\nu >0$ such that 
the distribution of $x_t$ satisfies 
$p(x_t)\ge \nu \cdot p(-x_t)$.
\end{enumerate}
Then Assumption~\ref{assumption:rdi_p}
holds with $\gamma(K) = \Lambda/2$ and
$\rho(K) = \nu\Lambda^2/128$.
\end{lemma}
\proof{Proof of Lemma~\ref{lemma:diversity_condition_modelp}}
For any unit vector $v\in\RR^d$, and $a \in \{1,2\}$, we have
\begin{align*}
\PP\Big((v^\top x_t)^2\cdot\Indc\{a^* = a\} \ge \frac{\Lambda}{2}\Big)
\ge \nu \cdot 
\PP\Big((v^\top x_t)^2\cdot \Indc\{a^* = 3-a\} \ge \frac{\Lambda}{2}\Big).
\end{align*}
As a result,
\begin{align*}
\PP\Big((v^\top x_t)^2\cdot\Indc\{a^* = a\} \ge \frac{\Lambda}{2}\Big)
\ge & \frac{\nu}{2} \cdot 
\PP\Big((v^\top x_t)^2  \ge \frac{\Lambda}{2}\Big)\\
\ge & \frac{\nu}{2} \cdot 
\PP\Big((v^\top x_t)^2  \ge \frac{1}{2} v^\top \EE[x_tx_t^\top]v\Big)
\ge  \frac{\nu \Lambda^2}{128},
\end{align*}
where the last inequality is due to 
the Paley-Zygmund inequality.
\endproof

\subsection{Regret Lower Bound}
\label{appendix:lower_bound_modelp}
\begin{theorem}\label{thm:lowerbound_modelp}
Under \modelp, consider the setting 
where $K = \log(T/s_0)$ and 
the context $x_t \sim \calN(0,I_d)$ for 
any $t\in[T]$. For any $M \le T$ and for any 
dynamic batch learning algorithm $\mathbf{Alg}$,
we have
\begin{align}
\label{eq:lowerbound_modelp}
\sup_{\{\theta^\star_a\}_{a\in[K]}: \|\theta_a^\star\|_2 \le 1, 
\|\theta_a^\star\|_1 \le s_0} 
\EE_{\{\theta_a^\star\}_{a\in[K]}}
\big[R_T(\mathbf{Alg})\big] \ge 
c\cdot \max\bigg(M^{-2} 2^{-M} \cdot \sqrt{Ts_0} \cdot 
\Big(\frac{T}{s_0}\Big)^{\frac{1}{2(2^M-1)}}
,\sqrt{Ts_0}\bigg)
\end{align}
where $\Expect_{\{\theta^\star_a\}_{a\in[K]}}$
denotes taking expectation w.r.t.~the distribution based 
on the set of  parameters $\{\theta^\star_a\}_{a\in[K]}$,
and $c>0$ is a numerical constant independent of $(T,M,d, s_0)$.
\end{theorem}
The proof of Theorem~\ref{thm:lowerbound_modelp} is similar
to that of Theorem~\ref{thm:dynamiclowerbnd}.
We define for any $m\in[M]$,
\begin{align*}
\Delta_m = \frac{1}{48\cdot M^2\cdot 2^{M}}
\cdot \Big(\frac{T}{s_0}\Big)^{-\frac{1-2^{1-m}}{2(1-2^{-M})}}, 
\qquad
T_m = \bigg \lfloor s_0 \cdot \Big( \frac{T}{s_0}\Big)^{\frac{1-2^{-m}}{1-2^{-M}}} 
\bigg \rfloor.
\end{align*}
We consider $K = 2^M$ arms and construct a prior $Q$ 
for $\{\theta^\star_a\}_{a\in [K]}$
in the following way: draw $\bth_1,\ldots,\bth_M$ 
independently from $\Unif(\mathbb{S}^{s_0-1})$.
Given $a \in [K]$, we can uniquely write $a = 1 + \sum^M_{m=1} a_m \cdot 2^{m-1}$,
where $a_m \in \{0,1\}$. We then let $\tilde{\theta}_a = \sum_{m=1}^M (-1)^{a_m} \cdot  \Delta_m \bth_m$  
and $\theta_a^* $ be a $d$-dimensional vector whose first $s_0$ coordinates
coincide with $\tilde{\theta}_a$ and the remaining zeros.
Moving on, we let $u_t = x_t(S)/\|x_t(S)\|$. 

For notational simplicity, we let $\Theta = (\theta_1,\ldots,\theta_K)$
and correspondingly $\Theta^\star = (\theta_1^\star,\ldots,\theta_K^\star)$.
Then
\begin{align*}
\sup_{\Theta^\star:\|\theta^\star_a\|_2\le 1,
~\|\theta^\star_a\|_0\le s_0,a\in[K]}
\Expect_{\Theta^\star}\big[R_T(\textbf{Alg})\big]
\ge \Expect_Q\Expect_{\Theta}\big[R_T(\textbf{Alg})\big]
= \sum^T_{t=1}\Expect_{Q}\EE_x
\EE_{P^t_{\Theta,x}}\Big[\max_{a\in[K]}x_t^\top \theta_a 
- x_t^\top \theta_{a_t}\Big].
\end{align*}
Given any $m \in [M]$ and any $t \in \{T_{m-1}+1,\ldots,T_m\}$,
define $\calA_m = \{a \in [K] : a_m = 0\}$ and
\begin{align}
  \label{eq:regret_decomp_multi}
   &\Expect_Q \Expect_x \Expect_{P_{\Theta,x}^t} 
  \Big[\max_{a \in [K]} x_t^\top \theta_a - x_t^\top \theta_{a_t} \Big]
  =  \Expect_Q \Expect_x \Expect_{P^t_{\Theta,x}}
  \bigg[\sum_{a \in [K]} \Indc\{a_t = a\} \cdot  
  \Big(\max_{a' \in [K]} x_t^\top \theta_{a'} - x_t^\top \theta_a \Big)\bigg]\nonumber\\
   = & \Expect_Q \Expect_x \Expect_{P_{\Theta,x}^t}
  \bigg[\sum_{a \in \calA_m}  \Indc\{a_t = a\} \cdot  
  \Big(\max_{a' \in [K]} x_t^\top \theta_{a'} - x_t^\top \theta_a \Big)
  +  \Indc\{a_t = a + 2^{m-1}\} \cdot  
  \Big(\max_{a' \in [K]} x_t^\top \theta_{a'} - x_t^\top \theta_{a+2^{m-1}} \Big) \bigg]\nonumber\\
  \ge & \Expect_Q \Expect_x \Expect_{P^t_{\Theta,x}}
  \bigg[\sum_{a \in \calA_m}  \Indc\{a_t = a\} \cdot  
  \Big(\max_{a' \in \{a,a+2^{m-1}\}} x_t^\top \theta_{a'} - x_t^\top \theta_a \Big) + \nonumber\\
  & \qquad\qquad\qquad\qquad \Indc\{a_t = a + 2^{m-1}\} \cdot  
  \Big(\max_{a' \in \{a,a+2^{m-1}\}} x_t^\top \theta_{a'} - x_t^\top \theta_{a+2^{m-1}} \Big) \bigg]\nonumber\\
  \ge & 2 \Delta_m \cdot \Expect_Q \Expect_x \Expect_{P^t_{\Theta,x}}
  \bigg[\sum_{a \in \calA_m}  \Indc\{a_t = a\} \cdot  
    \big(x_t(S)^\top \bth_m\big)_-
  +  \Indc\{a_t = a + 2^{m-1}\} \cdot \big(x_t(S)^\top\bth_m\big)_+ \bigg] \nonumber\\
  = & 2 \Delta_m \cdot \Expect_Q \Expect_x \Expect_{P^t_{\Theta,x}}
  \bigg[ \Indc\{a_t \in \calA_m\} \cdot  
    \big(x_t(S)^\top \bth_m)_-
    +  \Indc\{a_t \in \calA_m^c\} \cdot 
  \big(x_t(S)^\top\bth_m\big)_+ \bigg].
\end{align}
We define two new measures $\Theta$ via
\begin{align*}
  \frac{dQ_{m,t}^+}{dQ}(\Theta) = \frac{(x_t(S)^\top\bth_m)_+}{Z_m(x_t)},
  \quad
  \frac{dQ_{m,t}^-}{dQ}(\Theta) = \frac{(x_t(S)^\top\bth_m)_-}{Z_m(x_t)},
\end{align*}
where $Z_m(x_t) = \Expect_Q[(x_t(S)^\top \bth_m)_+] = \Expect_Q[(x_t(S)^\top\bth_m)_-]$ is the
common normalizing constant. With the new notation, we can write
\begin{align}\label{eq:regret_intermediate}
  \eqref{eq:regret_decomp_multi} = & 2\Delta_m \cdot \Expect_x
  \Big[Z_m(x_t) \cdot 
  \Big(\Expect_{P^t_{\Theta,x} \circ Q_{m,t}^-}\big[\Indc \{a_t\in \calA_m\}\big]
  + \Expect_{P^t_{\Theta,x} \circ Q_{m,t}^+}\big[\Indc \{a_t \in \calA_m^c\}\big]\Big) \Big],
\end{align}
where $P^t_{\Theta,x} \circ Q^+_{m,t}$ (resp. $P^t_{\Theta,x} \circ Q^-_{m,t}$)
is a mixed distribution:  $\Theta$ is drawn from $Q^+_{m,t}$ (resp. $Q^-_{m,t}$)
and observed rewards are then drawn from $P^t_{\Theta,x}$.

\subsubsection{Regret lower bound when a ``bad'' event happens with large probability}\label{subsec:lower1}
As before, the regret is large when a ``bad'' event $B_m$ ($t_{m-1} \le T_{m-1} < T_m \le t_m $) 
is likely to happen under the prior. 
\begin{lemma}\label{lem:bnd}
If there exists $m \in [M]$, such that
\begin{align}
  \sum^{T_m}_{t=T_{m-1}+1} \Expect_x\Big[Z_m(x_t) 
  \cdot \Expect_{P_{\Theta,x}\circ Q^+_{m,t}}\big[\Indc\{A_m\}\big] \Big]
  \ge \dfrac{T_m-T_{m-1}}{8 \cdot M^2\cdot 2^{M}},\label{assumption:asst_on_m}
\end{align} 
then there eixsts a numerical constant $c>0$, independent of $(T,M,d, s_0)$, such that, 
\begin{align*}
  \sup_{\theta^\star:\|\theta^\star\|_2\le 1,~\|\theta^\star\|_0\le s_0} \Expect_{\theta^\star}\big[R_T(\mathbf{Alg})\big]
  \ge \frac{c}{M^2\cdot 2^{M}}\cdot \sqrt{Ts_0}\left(\dfrac{T}{s_0}\right)^{\frac{1}{2(2^M-1)}}.
\end{align*}
\end{lemma}

For any $m\in[M]$
\begin{align}\label{eq:regret_tv}
  \eqref{eq:regret_intermediate} 
  \ge  2\Delta_m \cdot \Expect_x
  \bigg[Z_m(x_t) \cdot 
  \Big(1 - \TV \big(P^t_{\Theta,x} \circ Q_{m,t}^+,
  P^t_{\Theta,x} \circ Q_{m,t}^-\big)\Big) \bigg],
\end{align}
where the inequality is due to $P(A) + Q(A^c) \ge 1 - \TV(P,Q)$.
Above,
\begin{align}\label{eq:tv_dis}
  &1-\TV(P_{\Theta,x}^t \circ  Q_{m,t}^-, P_{\Theta,x}^t \circ Q_{m,t}^+)
  \stackrel{\rm (a)}{\ge}  
  1-\TV(P_{\Theta,x}^{T_m} \circ  Q_{m,t}^-, P_{\Theta,x}^{T_m} \circ Q_{m,t}^+)
  = \int \min \big(dP^{T_m}_{\Theta,x}\circ Q_{m,t}^-,
  dP^{T_m}_{\Theta,x}\circ Q_{m,t}^+\big)\nonumber\\
  \ge &  \int_{B_m} \min \big(dP^{T_m}_{\Theta,x}\circ Q_{m,t}^-,
  dP^{T_m}_{\Theta,x}\circ Q_{m,t}^+\big)
  \stackrel{\rm (b)}{=} \int_{B_m} 
  \min \big(dP^{T_{m-1}}_{\Theta,x}\circ Q_{m,t}^-,
  dP^{T_{m-1}}_{\Theta,x}\circ Q_{m,t}^+\big),
\end{align}
where step (a) is due to the data-processing inequality and 
step (b) follows from the fact that on the event $B_m$,
there is $P_{\Theta,x}^{T_m} = P_{\Theta,x}^{T_{m-1}}$.
Next,
\begin{align*}
  \eqref{eq:tv_dis} = 
  &\frac{1}{2} \int_{B_m} dP^{T_{m-1}}_{\Theta,x} \circ Q_{m,t}^+
  + dP^{T_{m-1}}_{\Theta,x} \circ Q_{m,t}^- 
  - |dP^{T_{m-1}}_{\Theta,x} \circ Q_{m,t}^+ 
  - dP^{T_{m-1}}_{\Theta,x} \circ Q_{m,t}^-|\\
  = & \frac{1}{2}\big(P^{T_{m-1}}_{\Theta,x} \circ Q_{m,t}^+(B_m)
  + P^{T_{m-1}}_{\Theta,x} \circ Q_{m,t}^-(B_m)\big) 
  - \TV\big(dP^{T_{m-1}}_{\Theta,x} \circ Q_{m,t}^+,
  dP^{T_{m-1}}_{\Theta,x} \circ Q_{m,t}^-\big)\\
  \ge & P^{T_{m-1}}_{\Theta,x} \circ Q_{m,t}^+(B_m) 
  - \frac{3}{2} \TV\big(dP^{T_{m-1}}_{\Theta,x} \circ Q_{m,t}^+,
  dP^{T_{m-1}}_{\Theta,x} \circ Q_{m,t}^-\big).
\end{align*}
By Pinsker's inequality, 
\begin{align*}
\TV\big(dP^{T_{m-1}}_{\Theta,x} \circ Q_{m,t}^+,
dP^{T_{m-1}}_{\Theta,x} \circ Q_{m,t}^-\big)
\le \sqrt{\frac{1}{2}D_{\rm KL}
\Big(dP^{T_{m-1}}_{\Theta,x} \circ Q_{m,t}^+~\big\|~
dP^{T_{m-1}}_{\Theta,x} \circ Q_{m,t}^-\Big)}
\end{align*}
Recall that $u_t = x_t(S)/\|x_t(S)\|_2$ and under $Q$, 
\begin{align*}
  \big(\bth_1,\ldots,\bth_m,\ldots,\bth_M\big) 
  \stackrel{\rm d}{=}
  \big(\bth_1,\ldots,\bth_m - 2(u_t^ \top \bth_m) u_t,\ldots,\bth_M\big) 
\end{align*}
Let $\tilde{\Theta}^{(m)} = \{\tilde{\theta}_a^{(m)}\}_{ a\in[K]}$
denote the set of $2^M$ arms 
induced by $(\bth_1,\ldots,\bth_m - 2(u_t^\top\bth_m)u_t,\ldots,\bth_M)$:
\begin{align*}
\tilde{\theta}^{(m)}_a = (-1)^{a_1} \Delta_1 \cdot \bth_1
+ \cdots
+ (-1)^{a_m} \Delta_m \cdot \big(\bth_m - 2(u_t^\top\bth_m)\cdot u_t\big)
+ \cdots
+(-1)^{a_M} \Delta_M \cdot \bth_M.
\end{align*}
Then $\Theta \sim Q_{m,t}^+$ if and only if $\tilde{\Theta}^{(m)} \sim 
Q_{m,t}^-$.
Consequently, 
\begin{align}\label{eq:joint_conv}
   D_{\rm KL}\Big(P_{\Theta,x}^{T_{m-1}} \circ Q_{m,t}^+
  ~\|~ P_{\Theta,x}^{T_{m-1}} \circ Q_{m,t}^-\Big)
  = & D_{\rm KL}\Big(P_{\Theta,x}^{T_{m-1}} \circ Q_{m,t}^+
  ~\|~  P_{\tilde{\Theta}^{(m)},x}^{T_{m-1}} \circ Q_{m,t}^+\Big) \nonumber \\
  \le &  \Expect_{Q_{m,t}^+} \Big[D_{\rm KL}\big(P^{T_{m-1}}_{\Theta,x}
  \| P^{T_{m-1}}_{\tilde{\Theta}^{(m)},x}\big)\Big],
\end{align}
where the last inequality is due to the joint convexity 
of the KL-divergence (see Lemma~\ref{lem:kl_conv}).
By direct computation,
\begin{align*}
  \eqref{eq:joint_conv} = \frac{1}{2}\Expect_{Q_{m,t}^+} 
  \bigg[\sum^{T_{m-1}}_{\tau = 1} \Big(x_{\tau}^\top\theta_{a_{\tau}}
  - x_\tau^\top \tilde{\theta}^{(m)}_{a_{\tau}}\Big)^2\bigg]
  = & 2 \Delta_m^2 \sum^{T_{m-1}}_{\tau = 1} \big(x_{\tau}(S)^\top u_t\big)^2 
  \cdot \Expect_{Q_{m,t}^+}\big[(u_t^\top\bth_m)^2\big]\\
  \le & 4 \frac{\Delta_m^2}{s_0^{3/2}} \frac{\|x_t(S)\|_2}{Z_m(x_t)} 
  \cdot\sum^{T_{m-1}}_{\tau=1}
  \big(x_{\tau}(S)^\top u_t\big)^2.
\end{align*}
The last inequality is because
\begin{align*}
&\Expect_{Q_{m,t}^+}\big[(u_t^\top \bth_m)^2\big]
= \frac{\Expect_{Q}\big[(x_t(S)^\top \bth_m)_+ \cdot (u_t^\top \bth_m)^2\big]}{Z_m(x_t)}
= \frac{1}{2} \cdot \frac{\|x_t(S)\|_2}{Z_m(x_t)} \cdot \Expect_{Q}\big[|u_t^\top\bth_m|^3 \big]\\
= & \frac{1}{2} \cdot \frac{\|x_t(S)\|_2}{Z_m(x_t)} \cdot \Expect_Q\big[|\bth_{m,1}|^3\big]
\le 2 \cdot \frac{\|x_t(S)\|_2}{Z_m(x_t)} \cdot s_0^{-3/2}.
\end{align*}
Using the above,
\begin{align*}
\eqref{eq:regret_intermediate} \ge & 2\Delta_m\cdot
\Bigg(\Expect_x\big[Z_m(x_t) \cdot P_{\Theta,x}^{T_{m-1}}\circ Q_{m,t}^+(A_m)\big] 
- \frac{3}{2}\Expect_x\bigg[Z_m(x_t)
\sqrt{2\frac{\Delta_m^2}{s_0^{3/2}}
\frac{\|x_t(S)\|_2}{Z_m(x_t)} \sum^{T_{m-1}}_{\tau=1}
\big(x_{\tau}(S)^\top u_t\big)^2}\bigg] \Bigg)\\
\stackrel{\rm (a)}{\ge} & 2\Delta_m\cdot
\Bigg(\Expect_x\big[Z_m(x_t) \cdot P_{\Theta,x}^{T_{m-1}}\circ Q_{m,t}^+(A_m)\big] 
- \frac{3}{2}\sqrt{2\Expect_x\bigg[Z_m(x_t)^2 \frac{\Delta_m^2}{s_0^{3/2}}
\frac{\|x_t(S)\|}{Z_m(x_t)}
\sum^{T_{m-1}}_{\tau = 1} \big(x_{\tau}(S)^\top u_t\big)^2\bigg]}
\Bigg)\\
\ge & 2\Delta_m\cdot
\bigg(\Expect_x\big[Z_m(x_t) \cdot P_{\Theta,x}^{T_{m-1}} \circ Q_{m,t}^+(A_m)\big] 
- 3\sqrt{\frac{\Delta_m^2 T_{m-1}}{s_0}}
\bigg)\\
\ge & 2\Delta_m \cdot \bigg(\Expect_x\big[Z_m(x_t) 
\cdot P_{\Theta,x}^{T_{m-1}} \circ Q_{m,t}^+(A_m)\big]
- \frac{1}{16 \cdot M^2 \cdot 2^{M}}\bigg),
\end{align*}
where step (a) is due to Jensen's inequality and the concavity of
$x \mapsto \sqrt{x}$.
So far, we have established for any $m\in [M]$ that
\begin{align*}
  \max_{\Theta^\star} \sum^T_{t=1} 
  \Expect_{\Theta^\star} 
  \Big[\max_{a\in [K]} x_t^\top \theta_a - 
  x_t ^\top \theta_{a_t}\Big] \ge 
  2\Delta_m \cdot\sum^{T_m}_{t = T_{m-1}}
  \bigg(\Expect_x\big[Z_m(x_t) \cdot P_{\theta,x}^{T_{m-1}} \circ Q_{m,t}^+(A_m)\big]
  - \frac{1}{16\cdot M^2 \cdot 2^{M}}\bigg).
\end{align*}
Taking $m$ to be the batch satisfying 
condition~\eqref{assumption:asst_on_m}, we
finish the proof.

\subsubsection{A ``bad'' event happens with large enough probability}\label{subsec:lower2}
It remains to show that with sufficiently high probability~\eqref{assumption:asst_on_m_c} holds: 
\begin{lemma}\label{lem:prob}
There exists some $m\in[M]$, such that: 
\begin{align*}
\sum^{T_m}_{t=T_{m-1}+1}\Expect_x \Big[ Z_m(x_t) 
\cdot \Expect_{P_{\Theta,x} \circ Q^+_{m,t}}
\big[\Indc \{B_m\} \big]\Big]
\ge \dfrac{T_m-T_{m-1}}{8\cdot M^2\cdot 2^{M}}.
\end{align*}
\end{lemma}
For any $m\in[M]$, and any $t\in \{T_{m-1}+1,\ldots,T_m\}$,
we have 
\begin{align}
  \label{eq:common_measure}
  \Expect_x\Big[Z_m(x_t) \cdot P_{\theta,x}^{T_{m-1}} \circ Q_{m,t}^+(B_m)\Big]
  = & \Expect_x\Expect_{Q}
  \Big[\big(x_t(S)^\top \bth_m\big)_+ \cdot 
  P_{\Theta,x}^{T_{m-1}}(B_m)\Big]
\end{align}
Since $B_m = \{t_{m-1} \le T_{m-1} \le T_m \le t_m\}$ 
is determined by $\{x_1,a_1,r_1,\ldots,x_{T_{m-1}},a_{T_{m-1}},r_{T_{m-1}}\}$,
$P_{\Theta,x}^{T_{m-1}}(B_m)$ is independent of $\{x_{\tau}\}_{\tau > T_{m-1}}$. 
Consequently,
\begin{align*}
  \eqref{eq:common_measure} = &\Expect_Q \Expect_x
  \Big[\big(x_T(S)^\top\bth_m \big)_+\cdot P_{\Theta,x}^{T_{m-1}}(B_m)\Big]
  = \Expect_Q\Expect_x \Expect_{P_{\Theta,x}} 
  \Big[\big(x_T(S)^\top \bth_m\big)_+\Indc\{B_m\}\Big]\\ 
  \ge & \Expect_Q \Expect_x\Expect_{P_{\Theta,x}} 
  \Big[\min_{m'\in [M]} \big(x_T(S)^\top\bth_{m'}\big)_+ \Indc\{B_m\}\Big]
  = \Expect_Q\Expect_x\Big[\min_{m'\in [M]} (x_T(S)^\top\bth_{m'})_+\Big] \cdot 
  \Expect_{\tilde{Q}}\Expect_{P_{\Theta,x}}\big[\Indc\{B_m\}\big],
\end{align*}
where the new measure $\tilde{Q}$ is defined via the change of measure:
\begin{align*}
  \frac{d \tilde{Q}}{dQ \times dP_x}(\Theta,x) 
  = \frac{\min_{m' \in [M]}\big(x_T(S)^\top \bth_{m'}\big)_+}
  {\Expect\Big[\min_{m' \in [M]}\big(x_T(S)^\top \bth_{m'}\big)_+\Big]}. 
\end{align*}
By the definition of $u_t$, there is
\begin{align*}
  \Expect_Q \Expect_x \Big[\min_{m'\in[M]} \big(x_t(S)^\top \bth_{m'}\big)_+\Big]
  \ge \Expect_x\Big[\|x_t(S)\|_2 \cdot \Expect_Q 
  \big[\min_{m'\in[M]} u_t^\top \bth_{m'}\big]\Big].
\end{align*}
We can then directly compute:
\begin{align*}
  &\Expect_Q\Big[\min_{m'\in[M]} (u_T^\top \bth_{m'})_+\Big]
  \stackrel{\rm (a)}{=} \Expect_Q\Big[\min_{m'\in[M]} (\bth_{m',1})_+\Big]
  = \int_0^{\infty} \Prob\Big(\min_{m'\in[M]} (\bth_{m',1})_+ > s\Big) {\rm ds}\\
  = & \int_0^{\infty} \Prob \big((\bth_{1,1})_+ > s \big)^M {\rm ds}
  = \frac{1}{2^M}\int_0^{\infty} \Prob\big(|\bth_{1,1}|^2 > s^2 \big)^M {\rm ds}
  \stackrel{\rm (b)}{\ge} \frac{1}{2^M}\int_0^{\frac{1}{2}\mathsf{B}(\frac{1}{2},\frac{s_0-1}{2})}
  \Big(1-\frac{2s}{\mathsf{B}(\frac{1}{2},\frac{s_0-1}{2})}\Big)^M {\rm ds}\\
  \ge & \frac{\mathsf{B}(\frac{1}{2},\frac{s_0-1}{2})}{(M+1)2^{M+1}}
  \ge \frac{1}{(M+1)2^{M+1}\sqrt{s_0}}.
\end{align*}
Above, $\mathsf{B}(\alpha,\beta)$ is the beta function with parameters
$\alpha$ and $\beta$; step (a) is because $\bth_1,\ldots,\bth_M$ 
are mutually independent; step (b) follows from the fact that 
$\bth_{m,1}^2$ follows the beta distribution with parameters $1/2$
and $(s_0-1)/2$.
Taking expectation over $x$, we then have 
\begin{align*}
\Expect_Q \Expect_x \Big[\min_{m'\in[M]}
(x_T(S)^\top \bth_{m'})_+\Big]
\ge \frac{1}{(M+1)2^{M+2}}.
\end{align*}
Furthermore, since the union of $\{B_m\}_{m\in[M]}$ 
is the whole space, by a union bound, we have
$\sum_{m=1}^M\Expect_{\tilde{Q}}
\Expect_{P_{\Theta,x}}[\Indc\{B_m\}] \ge
\Expect_{\tilde Q} \Expect_{P_{\Theta,x}}
[\Indc\{\cup_{m=1}^M B_m\}] = 1$.
Hence there must exist $\bar{m} \in [M]$
such that $\Expect_{\tilde Q}\Expect_{P_{\Theta,x}}(B_{\bar m}) \ge 1/M$
and
\begin{align*}
  \sum^{T_{\bar m}}_{t = T_{\bar{m}-1}}\EE_x
  \Big[Z_{\bar m}(x_t) \cdot \EE_{P_{\Theta,x} 
 \circ Q_{\bar{m},t}^+}[\Indc\{B_{\bar{m}}\}]\Big]1
  \ge \frac{T_{\bar m} - T_{\bar{m}-1}}{M(M+1)2^{M+2}}
  \ge \frac{T_{\bar m} - T_{\bar{m}-1}}{8 \cdot M^2\cdot 2^{M}},
\end{align*}
completing the proof.


\subsubsection{Lower Bound for Fully Online Learning Setting}
\label{sec:lower_bnd_online_modelc}
It suffices now to show that the regret is lower bounded by the second term in~\eqref{eq:lowerbound}.
This is established in the following lemma.

\begin{lemma}\label{lem:online}
When $M = T$, under the setting of two independent 
Guassian contexts, we have (for some numerical 
constant $c$ independent of $T,M,d, s_0$):
\[
\sup_{\Theta^\star:\|\theta^\star_a\|_2\le 1, \|\theta^\star_a\|_0\le s_0,
a\in\{1,2\}}
\Expect_{\theta^\star}\big[R_T(\mathbf{Alg})\big] \ge c\cdot \sqrt{Ts_0}.
\]
\end{lemma}

As in the batched case (and with the same notation), 
we construct a prior $Q$ for $\Theta^\star$: sample
$\bth$ from $\Unif(\mathbb{S}^{s_0 - 1})$; we then  
construct $\theta_1\in \mathbb{R}^d$ such that 
$\theta_1(S) = \Delta\bth$ and $\theta_1(S^c)=0$, 
where $\Delta = \frac{1}{8}\sqrt{\frac{s_0}{T}}$.
Finally we let $\theta_2 = - \theta_1$.
Then:
\begin{align}
&\sup_{\Theta^\star:\|\theta^\star_a\|_2\le 1, \|\theta^\star_a\|_0\le s_0, a\in\{1,2\}} 
\Expect_{\Theta^\star}\big[R_T(\textbf{Alg})\big]
\ge  \Expect_Q\Expect_\Theta \big[R_T(\textbf{Alg})\big] 
= \sum^T_{t=1}\Expect_{Q} 
\Expect_x\Expect_{P^t_{\Theta,x}}\Big[
\max_{a\in \{1,2\}}(x_t^\top\theta_a - x_t^\top\theta_{a_t})
\Big]\nonumber \\
=& 2\Delta \sum^T_{t=1}\Expect_x\bigg[Z(x_t) \cdot 
\Big(\Expect_{P_{\Theta,x}\circ Q^-_t}\big[\Indc(a_t=1)] + 
\Expect_{P_{\Theta,x}\circ Q^+_t}\big[\Indc(a_t=2)] \Big)\bigg],\label{eq:ont1}
\end{align}
where we similarly define two measures via:
$\frac{dQ^-_t}{dQ}(\Theta) = \frac{(x_t(S)^\top\bth)_-}{Z(x_t)},
\frac{dQ^+_t}{dQ} (\Theta)= \frac{(x_t(S)^\top\bth)_+}{Z(x_t)},$
with $Z(x_t) = \frac{1}{2}\Expect_Q \big[|x_t(S)^\top \bth|\big]$ 
6being a common normalizing constant. Note that 
$\bth \stackrel{\rm d}{=} \bth - 2(u_t^\top \bth)u_t$.
Let $\tilde \Theta = \{\tth_1,\tth_2\}$ be the set of vectors
induced by $\bth - 2(u_t^\top \bth) u_t$. Then 
$\Theta \sim Q^-_t$ iff $\tilde \Theta \sim Q^+_t$.
Using this representation, we have 
\begin{align}
\eqref{eq:ont1} \stackrel{(a)}{\ge}& 2\Delta\sum^T_{t=1}\Expect_x 
\Big[Z(x_t) \cdot \big(1-\TV(P^{t-1}_{\Theta,x} \circ Q^-_t,
P_{\Theta,x}^{t-1} \circ Q^+_t ) \big)\Big]\\
\stackrel{(b)}{\ge} & 2\Delta \sum^T_{t=1}\Expect_x\bigg[Z(x_t) 
\cdot \Big(
1 - \sqrt{\frac{1}{2}D_{\rm KL}\big(P^{t-1}_{\Theta,x}\circ Q^-_t ~\|~
P^{t-1}_{\tilde \Theta,x} \circ Q^-_t \big)}
\Big)\bigg]\nonumber\\
\stackrel{(c)}{\ge} & 2\Delta\sum^T_{t=1}\Expect_x\Bigg[ 
Z(x_t) \cdot \bigg(1-\sqrt{\frac{1}{2}\Expect_{Q^-_t} 
\Big[D_{\rm KL}\big(P^{t-1}_{\Theta,x} ~\|~ 
P^{t-1}_{\tilde\Theta,x}\big)\Big]}\bigg) \Bigg] 
\label{eq:ont2}
\end{align}
where step (a) follows from $P(A)+Q(A^c) \ge 1-\TV(P,Q)$; 
step (b) is by Pinsker's inequality; step (c) is because
of the joint convexity of the KL-divergence. The 
KL-divergence is then: 
\begin{align*}
D_{\rm KL} \big(P^{t-1}_{\Theta,x}~\|~P^{t-1}_{\tilde\Theta,x} \big)
= \dfrac{\Delta^2}{2}\sum^{t-1}_{\tau=1}
\Big(2 \big(u_t^\top\bth\big)\cdot\big(u_t^\top x_{\tau}(S)\big)\Big)^2
= 2\Delta^2(u_t^\top\bth)^2 \cdot u_t^\top 
\Big(\sum^{t-1}_{\tau=1}x_{\tau}(S)x_{\tau}(S)^\top \Big)u_t
\end{align*}
Plugging in the expression of the KL-divergence, we have
\begin{align*}
\eqref{eq:ont2} = & 2\Delta\sum^T_{t=1}\Expect_x\Bigg[ Z(x_t)\cdot \bigg(
1-\sqrt{\Delta^2\Expect_{Q^-_t} \big[(u_t^\top \bth)^2\big] \cdot 
u_t^\top\Big(\sum^{t-1}_{\tau=1} x_{\tau}(S)^\top 
x_{\tau}(S)\Big)u_t}\bigg)\Bigg]\\
\stackrel{\rm (a)}{\ge} & 2\Delta \sum^T_{t=1}\Expect_x\bigg[Z(x_t)\cdot\Big(
1-\sqrt{\dfrac{5t\Delta^2}{s_0}}\Big)\bigg]
\stackrel{\rm (b)}{\ge}  \dfrac{T \Delta}{5}
= \dfrac{\sqrt{T s_0}}{40},
\end{align*}
where step (a) is by taking expectation w.r.t. 
$\big\{x_{\tau}\big\}_{\tau\le t-1}$ and 
Lemma~\ref{lemma:uniformmoment},
and step (b)  the choice of $\Delta$.

\subsection{Regret Upper Bound}
\label{appendix:upper_bound_modelp}
Algorithm~\ref{algo.dbgl_p} describes a variant of the LBGL
algorithm under \modelp; Theorem~\ref{thm:upperbnd_p}
establishes a corresponding regret upper bound 
under Assumptions~\ref{assumption:contexts_p}-\ref{assumption:actions_p},
and Corollary~\ref{cor:onlineupperbnd_modelp} 
gives an upper bound for the online learning problem.

\begin{algorithm}[ht]
  \DontPrintSemicolon  
  \SetAlgoLined 
  \BlankLine
  \caption{LASSO Batch Greedy Learning (LBGL) under \modelp}
  \label{algo.dbgl_p}
  \textbf{Input} Time horizon $T$; context dimension $d$; number of batches $M$; sparsity bound $s_0$.\;
  \textbf{Initialize} $b = \Theta\Big( \sqrt{T} \cdot ({T}/{s_0})^{\frac{1}{2(2^M-1)}}\Big)$; $\hat{\theta}_0 = \mathbf{0}\in \mathbb{R}^d$;\;
  \textbf{Static grid} $\calT = \{t_1,\ldots,t_M\}$, with $t_1 = b\sqrt{s_0}$ and $t_m = b\sqrt{t_{m-1}}$ for $t \in\{2,\ldots,M\}$;\;
  \textbf{Partition} each batch into $M$ intervals evenly, i.e., $(t_{m-1},t_{m}] = \cup^M_{j=1}T_m^{(j)}$, for $m\in[M]$.\;
  \For{$m \gets 1$ \KwTo $M$}{
    \For{$t \gets t_{m-1}$ \KwTo $t_m$}{
    (a) Choose $a_t = \argmax{a\in[K]} x_{t}^\top \hat{\theta}_{m-1,a}$ 
    (break ties with lower action index).\;
    (b) Incur reward $r_{t,a_t}$.
  }
  $T^{(m)}\leftarrow \cup^m_{m'=1}T^{(m)}_{m'}$; 
  $\lambda_m \leftarrow 18\cdot \sqrt{\dfrac{\log{T}}{|T^{(m)}|}}$;\;
  Update $\hat{\theta}_{m,a} 
  \leftarrow \argmin{\theta\in\mathbb{R}^d}~\dfrac{1}{2|T^{(m)}|} 
  \sum_{t\in T^{(m)}}(r_{t,a_t} - x_{t}^\top\theta)^2 \cdot \Indc \{a_t = a\} 
  + \lambda_m \|\theta\|_1$.
}
\end{algorithm}

\begin{theorem}
\label{thm:upperbnd_p}
Under \modelp, Assumptions~\ref{assumption:contexts_p}-\ref{assumption:actions_p} 
and $M = O\big(\log{\log{(T/s_0)}}\big)$, we have
\begin{align}
\sup_{\Theta^\star:\|\theta^\star_a\|_2\le 1, \|\theta^\star_a\|_0\le s_0}
\Expect_{\Theta^\star} \big[R_T(\mathbf{Alg})\big]
\le \frac{C\cdot M^{3/2} \sqrt{\log T \log (TK)}}{\gamma(K)\rho(K)} \cdot 
\sqrt{Ts_0}\left(\dfrac{T}{s_0}\right)^{\frac{1}{2(2^M-1)}},\label{eq:main_upperbnd_p}
\end{align}
where \textbf{Alg} is LBGL and $C>0$ is a numerical constant independent of $(T, d, M, K, s_0)$.
\end{theorem}

\begin{corollary}
\label{cor:onlineupperbnd_modelp}
In the fully online learning setting ($M=T$) and under 
Assumptions~\ref{assumption:contexts_p}-\ref{assumption:actions_p}:
\begin{align}
\sup_{\Theta^\star:\|\theta^\star_a\|_2\le 1, \|\theta^\star_a\|_0\le s_0}
\Expect_{\theta^\star}\big[R_T(\mathbf{Alg})\big]
\le \frac{C\sqrt{\big(\log{\log{(T/s_0)}}\big)^3 \cdot \log{T} \cdot \log{(TK)}}}{\gamma(K)\rho(K)} 
\cdot\sqrt{Ts_0},\label{eq:online_upperbnd_p}
\end{align}
where $C>0$ is a numerical constant independent of $(T, d, M, K, s_0)$.
\end{corollary}

\subsubsection{Eigenvalue Conditions}
We define for any $j,m\in[M]$ and $a\in[K]$ the 
empirical covariance matrix: 
$D_{m,j,a} = \sum_{t\in T^{(j)}_m} x_{t}x_{t}^\top \Indc\{a_t = a\}$
and $A_{m,a} = \sum^m_{m'=1} D_{m',m,a}$.
Lemma~\ref{lemma:multi_eigen_condition} shows that 
the restricted eigenvalues are bounded from both 
above and below with high probabilities.

\begin{lemma}
\label{lemma:multi_eigen_condition}
Suppose Assumptions~\ref{assumption:contexts_p}-\ref{assumption:actions_p}
hold. Given a sparsity parameter $s$, for any $j,m \in[M]$ and $a\in[K]$,
with probability at least 
$1 - 2\exp\big(O(s \cdot \log(\frac{d}{\rho(K)\gamma(K)})) - 
\Omega(\rho^2(K)\sqrt{Ts_0}/M)\big)$,
\begin{align*}
\phi_{\max}\Big(s,\frac{D_{m,j,a}}{|\tmj|}\Big)\le \frac{25}{2},
\qquad
\phi_{\min}\Big(s,\frac{D_{m,j,a}}{|\tmj|}\Big) \ge \frac{\rho(K)\cdot \gamma(K)}{4}.
\end{align*}
\end{lemma}

\proof{Proof of Lemma~\ref{lemma:multi_eigen_condition}}
Given a sparsity parameter $s$, let $\calN(\varepsilon)$
denote the $\varepsilon$-net of $\mathbb{S}^{s-1}$.
\paragraph{Upper bound}
Fixing an arbitrary $s$-sparse vector $v\in\bR^d$, 
we let $Y_t = v^\top x_t$. For any $\delta,\mu >0$,
\begin{align}
\label{eq:eigen_upperbnd}
\PP \bigg(\frac{1}{|\tmj|}\sum_{t \in \tmj} Y_t^2 \cdot \Indc\{a_t = a\} 
\ge 4 + \delta \bigg)
\stackrel{\rm (a)}{\le} & \exp \Big(-(4+\delta)\mu \cdot |\tmj|\Big)
\cdot \EE\bigg[\exp\Big(\sum_{t\in\tmj}\mu \cdot Y_t^2 \Indc\{a_t = a\} \Big)\bigg]\nonumber\\
\le & \exp \Big(-(4+\delta)\mu \cdot |\tmj|\Big)
\cdot \EE\bigg[\exp\Big(\sum_{t\in\tmj}\mu \cdot Y_t^2\Big)\bigg]\nonumber\\
\stackrel{\rm (b)}{=} & \exp \big(-\delta \mu \cdot |\tmj|\big)
\cdot \prod_{t\in\tmj}\EE\bigg[\exp \Big(\mu\big(Y_t^2 - 4\big)\Big)\bigg],
\end{align}
where the step (a) is a result of Markov's inequality
and step (b) is due to the independence between $\{x_t\}_{t\in\tmj}$.
Since $x_t$ is $1$-sub-Gaussian, $v^\top x_t$ is $1$-sub-Gaussian.
Hence, $Y_t^2 - \EE[Y_t^2]$ is $(4\sqrt{2},4)$-sub-exponential and $\EE[Y_t^2]\le 4$.
Using this result, we have 
\begin{align*}
\eqref{eq:eigen_upperbnd}
\le \exp\Big(-\delta\mu \cdot |\tmj|\Big) \cdot
\prod_{t\in\tmj}\EE\bigg[\exp\Big(\mu\cdot\big(Y_t^2 - \EE[Y_t^2]\big)\Big)\bigg]
\le \exp\bigg(- \min\Big(\frac{\delta}{8},\frac{\delta^2}{64}\Big) \cdot |\tmj|\bigg)
\end{align*}
Letting $\delta = 8$ and taking a union bound over
all the $d$-dimensional vectors whose support is in
$\calN(\varepsilon)$, we obtain that with probability
at least $1 - \exp \big(s\log d + s \log(1+2/\varepsilon)-|\tmj|\big)$,
$$
\frac{1}{|\tmj|}\sum_{t\in\tmj}(v^\top x_t)^2 
\cdot \Indc\{a_t = a\} <12,
$$
for all $v$ whose support is in $\calN(\varepsilon)$. For
an arbitrary $s$-sparse vector $v$, let $\supp(v)$
denote its support. Without loss of generality, suppose
$|\supp(v)| = s$. By the definition of the $\varepsilon$-net,
we can find $u_0 \in \calN(\varepsilon)$ such that
$\|\supp(v) - u_0\|_2\le \varepsilon$. We then construct 
the $d$-dimensional vector $u$ such that $u(\supp(v)) = u_0$
and $u(\supp(v)^c) = 0$. Then,
\begin{align*}
& \frac{1}{|\tmj|}\sum_{t\in\tmj} (v^\top x_t)^2 \cdot \Indc\{a_t = a\}
- \frac{1}{|\tmj|}\sum_{t\in\tmj} (u^\top x_t)^2 \cdot \Indc\{a_t = a\}\\
= & \frac{1}{|\tmj|}\sum_{t\in\tmj} v^\top x_t x_t^\top (v-u) \cdot \Indc\{a_t = a\}
- \frac{1}{|\tmj|}\sum_{t\in\tmj} u^\top x_t x_t^\top (u-v) \cdot \Indc\{a_t = a\}\\
\le & 2\varepsilon \cdot \phi_{\max}\Big(s,\frac{D_{m,j,a}}{|\tmj|}\Big). 
\end{align*}
Taking the supremum over all $s$-sparse vectors $v$ and
rearranging the above, we conclude that with probability
at least $1 - \exp(s \log d + s \log(1 + 2/\varepsilon) - |\tmj|)$,
\begin{align*}
\phi_{\max}\Big(s,\frac{D_{m,j,a}}{|\tmj|}\Big)
\le \frac{12}{1 - 2\varepsilon}.
\end{align*}

\paragraph{Lower bound}
We again fix a $s$-sparse vector $v\in \bR^d$
and let $Y_t = v^\top x_t$.
For any $a\in[K]$,
\begin{align}\label{eq:eigen_lowerbnd}
&\PP\bigg(\frac{1}{|\tmj|}\sum_{t\in\tmj} Y_t^2 
\cdot\Indc \{a_t = a\} \le \frac{\rho(K)\gamma(K)}{2}\bigg)
= \PP\bigg(\frac{1}{\tmj}\sum_{t\in\tmj} 
\frac{Y_t^2}{\gamma(K)}\cdot \Indc\{a_t = a\}
\le \frac{\rho(K)}{2}\bigg) \nonumber\\
\le &\PP\bigg(\frac{1}{\tmj}\sum_{t\in\tmj} 
\Indc\big\{Y_t^2 \ge \gamma(K), a_t = a\big\}
\le \frac{\rho(K)}{2}\bigg) \nonumber\\
= & \PP\bigg(\frac{1}{\tmj}\sum_{t\in\tmj} 
\Indc\big\{Y_t^2 \ge \gamma(K), a_t = a\big\}
- \PP\big(Y_t^2 \ge \gamma(K) , a_t = a\big)
\le \frac{\rho(K)}{2} - 
\PP\big(Y_t^2 \ge \gamma(K),a_t = a\big)\bigg) \nonumber\\
\le &\PP\bigg(\frac{1}{\tmj}\sum_{t\in\tmj} 
\Indc\big\{Y_t^2 \ge \gamma(K), a_t = a\big\}
- \PP\big(Y_t^2 \ge \gamma(K) , a_t = a\big)
\le -\frac{\rho(K)}{2}\bigg), 
\end{align}
where the last inequality is due to Assumption~\ref{assumption:rdi}.
Applying Hoeffding's inequality, we obtain that
\begin{align*}
\eqref{eq:eigen_lowerbnd} \le 
\exp \Big(- \frac{|\tmj| \cdot \rho^2(K)}{2}\Big).
\end{align*}
Taking a union bound over all the $d$-dimensional
vectors whose support is in $\calN(\varepsilon)$, 
we have with probability at least 
$1 - \exp(s \log d + s \log(1 + 2/\varepsilon) - 
|\tmj| \cdot \rho^2(K)/2)$,
\begin{align*}
  \frac{1}{|\tmj|} \sum_{t\in\tmj} (v^\top x_t)^2
  \cdot \Indc\{a_t = a\} \ge \frac{\rho(K)\gamma(K)}{2},
\end{align*}
for all $v$ whose support is in $\calN(\varepsilon)$.
Conditional on the above event, for an arbitrary
$s$-sparse $d$-dimensional vector 
$v$, by the definition of an $\varepsilon$-net,
we can find $u_0 \in \calN(\varepsilon)$ such 
that $\|u_0 - \supp(v)\|_2 \le \varepsilon$.
Let $u\in \mathbb{R}^d$ such that $u(\supp(v)) = u_0$
and $u(\supp(v)^c)=0$. We then have
\begin{align*}
\frac{1}{|\tmj|}\sum_{t\in \tmj} (v^\top x_t)^2 
\cdot \Indc\{a_t = a\}
\ge & \frac{1}{|\tmj|} \sum_{t\in\tmj}\Big( (u^\top x_t)^2 
+ 2(v-u)^\top x_t x_t^\top u  \Big) \cdot\Indc\{a_t = a\}\\
\ge & \frac{\rho(K)\gamma(K)}{2} - 2\varepsilon 
\phi_{\max}\Big(s,\frac{D_{m,j,a}}{|\tmj|}\Big).
\end{align*}
Finally, taking $\varepsilon = \min(\frac{1}{50},
\frac{\rho(K)\gamma(K)}{100})$, we have 
with probability at least 
$1 - 2 \exp\big(O(s\log\frac{d}{\rho(K)\gamma(K)}) - 
\Omega(\rho^2(K)\sqrt{T s_0}/M)\big)$,
\begin{align*}
  \phi_{\max}\Big(s,\frac{D_{m,j,a}}{|\tmj|}\Big) \le \frac{25}{2},
  \qquad
  \phi_{\min}\Big(s,\frac{D_{m,j,a}}{|\tmj|}\Big) 
  \ge \frac{\rho(K)\cdot \gamma(K)}{4}.
\end{align*}
\endproof

\subsubsection{Lasso Estimation Error}

With well-behaved restricted eigenvalues, Lemma~\ref{lemma:multi_lasso_err}
leverages standard Lasso results to prove an
estimation error bound for $\|\hat{\theta}_{m,a}-\theta^\star_a\|_2$.

\begin{lemma}
\label{lemma:multi_lasso_err}
Suppose Assumptions~\ref{assumption:contexts_p}-\ref{assumption:actions_p}
hold. Given any $a\in[K]$ and $m \ge 2$, with probability at 
least $1 - 2M \exp(O(\frac{s_0}{\rho(K)\gamma(K)}) \cdot 
\log (\frac{d}{\rho(K)\gamma(K)}) - \Omega(\sqrt{Ts_0}/M)) - 2T^{-2} -
2\exp(\log d - \Omega(\sqrt{Ts_0}/M))$, 
\begin{align*}
\big\|\theta_a - \hat{\theta}_{m,a}\big\|_2 
\le \frac{2048}{\rho(K)\gamma(K)}\cdot 
\sqrt{\frac{Ms_0\log T}{t_m}}.
\end{align*}
\end{lemma}

By the definition of $\hth_{m,a}$, 
\begin{align*}
\frac{1}{2|\tm|}\sum_{t\in\tm} \big(r_{t,a_t} - x_t^\top \hth_{m,a}\big)^2
\cdot \Indc\{a_t = a\} + \lambda_m \|\hth_{m,a}\|_1
\le \frac{1}{2|\tm|}\sum_{t\in\tm} \big(r_{t,a_t} - x_t^\top \theta_{a}\big)^2
\cdot \Indc\{a_t = a \} + \lambda_m \|\theta_{a}\|_1.
\end{align*}
Rearranging yields
\begin{align*}
\frac{1}{2|\tm|}\sum_{t\in\tm}(x_t^\top \theta_a - x_t^\top \hth_{m,a})^2
\cdot \Indc\{a_t = a\} + \lambda_m \|\hth_{m,a}\|_1 
\le \frac{1}{|\tm|}\sum_{t\in\tm}(x_t^\top \hth_{m,a} - x_t^\top\theta_a)
\cdot \varepsilon_t \cdot \Indc\{a_t = a\}
  + \lambda_m \|\theta_a\|_1.
\end{align*}
Due to the construction of $\tm$, conditional on 
$\{x_t,a_t\}_{t\in\tm}$, $\{\varepsilon_t\}_{t\in\tm}$
are mutually independent.
\begin{align}\label{eq:lasso_intermediate}
\frac{1}{|\tm|}\sum_{t\in \tm} x_t^\top
(\hth_{m,a} - \theta_a) \cdot \varepsilon_t 
\cdot \Indc\{a_t = a\} 
= & \frac{1}{|\tm|} 
\sum_{j\in[d]}(\hth_{m,a,j} - \theta_{a,j})
\sum_{t \in \tm}
x_{t,j}   \varepsilon_t \Indc\{a_t = a\}\nonumber\\
\le & \frac{1}{|\tm|}\sum_{j\in[d]}
|\hth_{m,a,j} - \theta_{a,j}| \cdot
\bigg|\sum_{t\in\tm} x_{t,j}\varepsilon_t \Indc\{a_t = a\}\bigg|
\end{align}
For a given $j\in[d]$ and $\delta,\delta_1>0$,
\begin{align*}
& \PP\Bigg(\frac{1}{|\tm|}\bigg|\sum_{t\in\tm} x_{t,j} \Indc\{a_t = a\}
\varepsilon_t \bigg| \ge \delta \Bigg)
=\EE\Bigg[\PP\bigg(\frac{1}{|\tm|}\Big|\sum_{t\in\tm} x_{t,j} \Indc\{a_t = a\}
\varepsilon_t \Big| \ge \delta \,\Big|\, \{x_{t,j}\}_{t\in\tm}\bigg)\Bigg]\\
\stackrel{(a)}{\le} & 2 \EE\bigg[\exp\Big(-\frac{|\tm|^2 \delta^2}
{2 \sum_{t \in \tm} x_{t,j}^2\Indc\{a_t = a\} }\Big)\bigg]\\
\stackrel{(b)}{\le} & 2 \exp\bigg(-\frac{|\tm| \delta^2}
{2 (4+\delta_1) } \bigg) 
+2 \PP \bigg(\frac{1}{|\tm|}\sum_{t\in\tm} x_{t,j}^2
\ge 4+\delta_1 \bigg),
\end{align*}
where step (a) uses Hoeffding's inequality and 
step (b) applies a union bound.
By the assumption, $x_t$ is $1$-sub-Gaussian, and 
hence $x_{t,j}$ is also $1$-sub-Gaussian;
$x_{t,j}^2 - \EE[x_{t,j}^2]$ is 
$(4\sqrt{2},4)$-sub-exponential and 
$\EE[x_{t,j}^2] \le 4$. Consequently,
\begin{align*}
\PP\Big(\frac{1}{|\tm|} \sum_{t\in\tm} x_{t,j}^2 \ge 4 + \delta_1\Big)
\le \PP\Big(\frac{1}{|\tm|} \sum_{t\in\tm} x_{t,j}^2 - \EE[x_{t,j}^2] \ge \delta_1 \Big)
\le \exp\bigg(-  \min\Big(\frac{\delta_1^2}{64}, \frac{\delta_1}{8}\Big)\cdot |\tm| \bigg).
\end{align*}
Letting $\delta = 9\sqrt{\log T / |\tm|}$ and $\delta_1 = 8$
and taking a union bound over $j\in[d]$,
we have with probability at least $1 - 2T^{-2} - 2\exp(\log d -|\tm|)$,
\begin{align}\label{eq:control_event}
  \frac{1}{|\tm|} \cdot \Big|\sum_{t\in\tm} x_{t,j} \varepsilon_t \Indc\{a_t = a\}\Big|
  \le 9\sqrt{\frac{\log T}{|\tm|}} = \frac{\lambda_m}{2},\qquad \mbox{for all }j\in[d].
\end{align}
On the event~\eqref{eq:control_event},
$\eqref{eq:lasso_intermediate} \le \frac{\lambda_m}{2}\|\hth_{m,a} - \theta_a\|_1$.
Consequently,
\begin{align*}
  \frac{1}{2|\tm|}\sum_{t\in\tm} (x_t^\top\theta_a - x_t^\top \hth_{m,a})^2
  \cdot \Indc\{a_t = a\} + \frac{\lambda_m}{2}\|\hth_{m,a} - \theta_a\|_1
  \le \lambda_m \cdot 
  \big( \|\hth_{m,a} - \theta_a\|_1 + \|\theta_a\|_1 - \|\hth_{m,a}\|_1\big).
\end{align*}
Denote $S_a = \supp(\theta_a)$. The above inequality yields
\begin{align*}
  &\frac{1}{2} \|\hth_{m,a} - \theta_a\|_1
  \le \|\hth_{m,a}(S_a) - \theta_a(S_a) \|_1
  + \|\theta_a(S_a)\|_1 - \|\hth_{m,a}(S_a)\|_1\\
  \Rightarrow 
  & \|\hth_{m,a}(S_a^c) - \theta_a(S_a^c)\|_1 
  \le 3\|\hth_{m,a}(S_a) - \theta_{a}(S_a) \|_1.
\end{align*}
Define $B_{m,a} = \sum_{t\in\tm} x_tx_t^\top \cdot \Indc\{a_t = t\}$.
For any unit vector $v\in \bR^d$,
\begin{align*}
  v^\top \frac{B_{m,a}}{\tm} v = 
  \sum_{j \le m} \frac{|\tmj|}{|\tm|} \cdot v^\top \frac{D_{j,m,a}}{|\tmj|}v.
\end{align*}
Combining the above and Lemma~\ref{lemma:multi_eigen_condition}, we have with
probability at least $1 - 2M\cdot \exp\big(O(s\log \frac{d}{\rho(K)\gamma(K)}) - 
\Omega(\rho^2(K) \cdot \sqrt{Ts_0}/M)\big)$,
\begin{align*}
  \phi_{\max}\Big(s, \frac{B_{m,a}}{|\tm|}\Big) \le \frac{25}{2},
  \qquad
  \phi_{\min}\Big(s,\frac{B_{m,a}}{|\tm|}\Big) \ge \frac{\rho(K)\cdot \gamma(K)}{4}.
\end{align*}

We now let $r = \frac{1800s_0}{\rho(K)\gamma(K)}$
and $s = s_0 + r$. With probability at least
$1 - 2M\exp(O(\frac{s_0}{\rho(K)\gamma(K)}\log \frac{d}{\rho(K)\gamma(K)})
-\Omega(\rho^2(K) \cdot \sqrt{Ts_0}/M))$,
\begin{align*}
  \frac{9s_0\phi_{\max}\Big(r, \frac{B_{m,a}}{|\tm|}\Big)}
  {r\phi_{\min}\Big(s_0+r, \frac{B_{m,a}}{|\tm|}\Big)}
  \le \frac{1}{4}
\end{align*}
and hence
\begin{align*}
  \kappa = \sqrt{\phi_{\min}\Big(s_0+r,\frac{B_{m,a}}{|\tm|}\Big)}
  \cdot \Bigg(1 - 3\sqrt{\frac{s_0\phi_{\max}\Big(r,\frac{B_{m,a}}{|\tm|}\Big)}
  {r\phi_{\max}\Big(s_0+r,\frac{B_{m,a}}{|\tm|}\Big)}}\Bigg)
  \ge \frac{\sqrt{\rho(K)\gamma(K)}}{4}.
\end{align*}
We now make use of Lemma~\ref{lemma:reconditions}:
\begin{align}\label{eq:aug1}
  \frac{1}{2|\tm|} \sum_{t\in\tm}
  (x_t^\top \theta_a - x_t^\top\hth_{m,a})^2
  \cdot \Indc\{a_t = a\}
  \ge \frac{\rho(K)\gamma(K)}{32} \|\hth_{m,a}(\tilde{S}_a) 
  - \theta_a(\tilde{S}_a)\|_2^2.
\end{align}
On the other hand,
\begin{align}\label{eq:aug2}
& \frac{1}{2|\tm|}\sum_{t\in\tm}
(x_t^\top\theta_A - x_t^\top\hth_{m,a})^2
\cdot \Indc\{a_t = a\}
\le  2\lambda_m\cdot \big\|\theta_a(S_a) - \hth_{m,a}(S_a)\big\|_1 \nonumber \\
\le & 2\lambda_m \sqrt{s_0} \cdot \big\|\theta_a(S_a) - \hth_{m,a}(S_a)\big\|_2
\le 2\lambda_m \sqrt{s_0} \cdot 
\big\|\theta_a(\tilde{S}_a) - \hth_{m,a}(\tilde{S}_a)\big\|_2.
\end{align}
Combining~\eqref{eq:aug1} and~\eqref{eq:aug2}, we
have
\begin{align*}
  \big\|\hth_{m,a}(\tilde{S}_a) - \theta_a(\tilde{S}_a)\big\|_2 \le 
  \frac{64\lambda_m\sqrt{s_0}}{\rho(K)\gamma(K)}.
\end{align*}
Observe that the $k$-th largest coordinate of 
$|\hth_{m,a}(S_a^c) - \theta_a(S_a^c)|$
is bounded by $\|\hth_{m,a}(S_a^c) - \theta_a(S_a^c) \|_1/k$. Then
\begin{align*}
  & \big\|\hth_{m,a}(\tilde{S}_a^c) - \theta_a(\tilde{S}_a^c)\big\|_2^2
  \le \big\|\hth_{m,a}(S_a^c) - \theta_a(S_a^c) \big\|_1^2
  \sum_{\ell = r+1}^{d-s_0}\frac{1}{\ell^2}
  \le \frac{1}{r} \cdot
  \big\|\hth_{m,a}(S_a^c) - \theta_a(S_a^c)\big\|_1^2\\
  \le & \frac{9}{r} \cdot \big\|\hth_{m,a}(S_a) - \theta_a(S_a)\big\|_1^2
  \le  \frac{9s_0}{r} \cdot \big\|\hth_{m,a}(S_a) - \theta_a(S_a)\big\|_2^2
  \le \frac{9s_0}{r} \cdot 
  \big\|\hth_{m,a}(\tilde{S}_a) - \theta_a(\tilde{S}_a) \big\|_2^2
\end{align*}
Combining everything above, we have with probability at least 
$1 - 2M \cdot \exp\big(O(\frac{s_0}{\rho(K)\gamma(K)} \log \frac{d}{\rho(K)\gamma(K)})-
\Omega(\rho^2(K) \cdot \sqrt{Ts_0}/M)\big) - 2T^{-2} - 
2\exp\big(\log d - \Omega(\sqrt{Ts_0}/M)\big)$,
\begin{align*}
  \big\|\hth_{m,a} - \theta_a \big\|_2 \le \sqrt{1 + \frac{9s_0}{r}} 
  \cdot
  \big\|\hth_{m,a}(\tilde{S}_a) - \theta_a(\tilde{S}_a)\big\|_2
  \le \sqrt{1+\frac{9s_0}{r}} \cdot \frac{64\lambda_m\sqrt{s_0}}
  {\rho(K)\gamma(K)}
  \le \frac{4608}{\rho(K)\gamma(K)}\sqrt{\frac{M s_0 \log T}{t_m}}.
\end{align*}

\subsubsection{Regret Analysis}
Given $m\in[M]$ and $t\in \{t_{m-1}+1,\ldots,t_m\}$, the instantantous
regret can be bounded as
\begin{align*}
& \max_{a\in[K]}~x_t^\top \theta_a - x_t^\top \theta_{a_t}
= \max_{a\in[K]}~x_t^\top \theta_a - x_t^\top \theta_{a_t}
- x_t^\top\hat{\theta}_{m-1,a_t} + x_t^\top \hat{\theta}_{m-1,a_t}\\
\stackrel{\rm (a)}{\le} & \max_{a\in[K]}~x_t^\top \theta_a - x_t^\top \theta_{a_t}
- x_t^\top\hat{\theta}_{m-1, a} + x_t^\top \hat{\theta}_{m-1, a_t} 
\le  2\cdot \max_{a \in [K]}~\big|x_t^\top(\theta_a - \hat{\theta}_{m-1,a})\big|,
\end{align*}
where step (a) is due to the definition of $a_t$.
Conditional on the previous batches, $x_t^\top (\theta_a - \hat{\theta}_{m-1,a})$
is $\|\theta_a - \hat{\theta}_{m-1,a}\|_2^2$-sub-Gaussian for a given $a \in [K]$.
Letting $\calH_t = \{x_1,a_1,r_1,\ldots,x_t,a_t,r_t\}$, we have
\begin{align*}
&\PP\bigg(2\cdot \max_{a\in[K]}~ 
\big|x_t^\top(\theta_a - \hat{\theta}_{m-1,a})\big|
\ge 6 \sqrt{\log(TK)} \cdot \max_{a\in[K]}~\big\|\theta_a - \hat{\theta}_{m-1,a}\big\|_2
\,\Big|\, \calH_{t_{m-1}}\bigg)\\
\le & \sum_{a\in[K]} \PP\bigg(\big|x_t^\top(\theta_a - \hat{\theta}_{m-1,a})\big| 
\ge 3\sqrt{\log (TK)}\cdot \max_{a\in[K]}~ \big\|\theta_a - \hat{\theta}_{m-1,a}\big\|_2
\,\Big |\, \calH_{t_{m-1}}\bigg)\\
\le & \sum_{a\in[K]} \PP\bigg(\big|x_t^\top(\theta_a - \hat{\theta}_{m-1,a})\big| 
\ge 3\sqrt{\log (TK)}\cdot  \big\|\theta_a - \hat{\theta}_{m-1,a}\big\|_2
\,\Big |\, \calH_{t_{m-1}}\bigg) \le \frac{1}{T^4},
\end{align*}
where the last inequality is due to the Chernoff bound.
Applying Lemma~\ref{lemma:multi_lasso_err} and a union bound
over $t\in\{t_{m-1}+1,\ldots,t_m\}$ and $a\in[K]$,
for $m \ge 2$, with probability
at least $1 - T^{-3} - 2MK\exp(O(\frac{s_0}{\rho(K)\gamma(K)}
\cdot \log (\frac{d}{\rho(K)\gamma(K)})) -\Omega(\sqrt{Ts_0}/M))
-2KT^{-2} - 2K\exp(\log d - \Omega(\sqrt{Ts_0}/M))$,
we  bound the regret incurred in the $m$-th batch as
\begin{align*}
  \sum^{t_m}_{t = t_{m-1}+1} 2\cdot\max_{a\in[K]}~\big|x_t^\top (\theta_a - \hth_{m-1,a})\big| 
  \le & 12288 \cdot \frac{\sqrt{M\log T \log (TK)}}{\rho(K)\gamma(K)}
  \cdot \sqrt{\frac{s_0}{t_{m-1}}}\cdot t_m\\
  \le & c_1 \cdot\frac{\sqrt{M\log T \log (TK)}}{\rho(K)\gamma(K)}
  \cdot \sqrt{Ts_0}\cdot \Big(\frac{T}{s_0}\Big)^{\frac{1}{2(2^M-1)}},
\end{align*}
where the last inequality follows from the choice
of the grids and $c_1 > 0$ is a numerical constant. 
Next, for the first batch, 
\begin{align*}
  \sum^{t_1}_{t=1} \max_{a\in[K]}~x_t^\top\theta_a 
  - x_t ^\top \theta_{a_t} \le 
  2 \sum_{t=1}^{t_1} \max_{a\in [K]}~\big|x_t^\top \theta_a\big|.
\end{align*}
Applying a maximal sub-Gaussian inequality and 
taking a union bound over $t\in[t_1]$, we have 
with probability at least $1 - T^{-2}$
\begin{align*}
  \sum^{t_1}_{t=1} \max_{a\in[K]}~x_t^\top\theta_a - x_t^\top \theta_{a_t}
  \le 6\sqrt{\log (TK)} \cdot t_1 \le c_2 \sqrt{\log (TK)}
  \cdot \sqrt{Ts_0}\cdot \Big(\frac{T}{s_0}\Big)^{\frac{1}{2(2^M-1)}},
\end{align*}
where $c_2 > 0$ is a numerical constant.
Combining everything above, we have with probability at least
$1 - (1+M+2MK)\cdot T^{-2} - 2MK\exp(\log d -\Omega(\sqrt{Ts_0}/M))
- 2M^2K\exp(O(\frac{s_0}{\rho(K)\gamma(K)} \cdot \log(\frac{d}{\rho(K)\gamma(K)}))
- \Omega(\sqrt{Ts_0}/M))$,
\begin{align*}
  R_T(\alg) \le c_3 \cdot\frac{\sqrt{M^3\log T \log (TK)}}{\rho(K)\gamma(K)}
  \cdot \sqrt{Ts_0}\cdot \Big(\frac{T}{s_0}\Big)^{\frac{1}{2(2^M-1)}},
\end{align*}
where $c_3 > 0$ is a numerical constant independent of
$(T,d,M,K,s_0)$.

\end{document}